\documentclass{article}

     \PassOptionsToPackage{numbers, compress}{natbib}

\usepackage[final,main]{neurips_2026}
\usepackage{amsmath}
\usepackage{amssymb}
\usepackage{mathtools}
\usepackage{bbm}
\usepackage{amsthm}
\usepackage{graphicx}
\usepackage{wrapfig}
\usepackage{caption}
\usepackage{subfigure}
\theoremstyle{plain}
\newtheorem{theorem}{Theorem}[section]
\newtheorem{proposition}[theorem]{Proposition}

\theoremstyle{definition}

\theoremstyle{remark}

\usepackage[utf8]{inputenc} 
\usepackage[T1]{fontenc}    
\usepackage{hyperref}       
\usepackage{url}            
\usepackage{booktabs}       
\usepackage{amsfonts}       
\usepackage{nicefrac}       
\usepackage{microtype}      
\usepackage{xcolor}         
\usepackage{algorithm}
\usepackage{algorithmic}

\usepackage{pgfplots}
\pgfplotsset{compat=1.18}  
\usepgfplotslibrary{
  groupplots,    
  fillbetween    
}
\definecolor{refpink}{RGB}{255,20,147}
\newcommand{\pinkref}[1]{\textcolor{refpink}{\ref{#1}}}
\usepackage[table]{xcolor}
\usepackage{booktabs}

\definecolor{tableHeader}{RGB}{224,236,248}   
\definecolor{tableSubhead}{RGB}{240,246,252}  
\definecolor{tableGray}{RGB}{248,249,251}     
\definecolor{tableFP}{RGB}{242,244,247}       
\definecolor{tableGreen}{RGB}{232,246,235}    
\definecolor{bestGreen}{RGB}{35,105,60}       
\definecolor{tableRule}{RGB}{165,175,185}     

\newcommand{\bestval}[1]{%
  \textcolor{bestGreen}{\textbf{#1}}%
}

\definecolor{citegreen}{RGB}{0,128,0}

\hypersetup{
    colorlinks=true,
    citecolor=citegreen
}

\usepackage{booktabs}        
\usepackage{multirow}        
\usepackage{tabularx}        
\usepackage{threeparttable}  
\usepackage{threeparttablex} 
\usepackage{siunitx}         
\usepackage{pgfplots}
\usepackage{pgfplotstable}
\usetikzlibrary{arrows.meta,positioning,fit,calc,patterns,decorations.pathreplacing}
\pgfplotsset{compat=1.18}

\usepackage{booktabs} \usepackage[table]{xcolor}
 \usepackage{multirow}
\usepackage{siunitx}
 \definecolor{best}{RGB}{198,239,206}   
 \definecolor{second}{RGB}{255,242,204} 
 
\pgfplotsset{
  mygrid/.style={grid=both,minor tick num=1,major grid style={opacity=0.25},minor grid style={opacity=0.1}},
  mylegend/.style={legend columns=2,legend cell align=left,legend pos=south east, fill=white, fill opacity=0.8, draw=none},
  mymark/.style={mark=*},
}
 \usepackage{tikz}
\title{When Bits Break Recourse: Counterfactual-Faithful Quantization}

\usepgfplotslibrary{fillbetween}      
\usetikzlibrary{intersections}        
\usetikzlibrary{arrows.meta,positioning,fit,calc,decorations.pathreplacing}
\usetikzlibrary{arrows.meta,positioning}
\definecolor{icmlBlue}{RGB}{42,92,172}
\definecolor{icmlOrange}{RGB}{219,112,46}
\definecolor{icmlGreen}{RGB}{33,140,70}
\definecolor{icmlPurple}{RGB}{128,94,168}
\definecolor{panelBG}{RGB}{248,249,251}
\definecolor{edgeGray}{RGB}{110,110,110}

\tikzset{
  >={Latex[length=2mm]},
  box/.style={draw, rounded corners=2pt, thick, align=center, inner sep=3pt, minimum height=8mm, fill=white},
  pane/.style={draw=edgeGray!40, rounded corners=3pt, fill=panelBG, inner sep=4pt},
  flow/.style={thick, draw=edgeGray, -{Latex[length=2mm]}},
  flowA/.style={thick, draw=icmlBlue, -{Latex[length=2mm]}},
  flowB/.style={thick, draw=icmlOrange, -{Latex[length=2mm]}},
  flowC/.style={thick, draw=icmlGreen, -{Latex[length=2mm]}},
  note/.style={draw=edgeGray, densely dashed, rounded corners=2pt, inner sep=2pt, font=\scriptsize, fill=white},
  tiny/.style={font=\scriptsize}
}
\definecolor{t3c}{RGB}{0,86,156}

\author{%
  Chaymae Yahyati$^{*}$ \\
  Faculty of Applied Sciences, Nador\\
  Mohammed First University, Oujda\\
  Morocco
  \And
  Ismail Lamaakal$^{*}$ \\
  Faculty of Applied Sciences, Nador\\
  Mohammed First University, Oujda\\
  Morocco
  \And
  Khalid El Makkaoui \\
  Faculty of Applied Sciences, Nador\\
  Mohammed First University, Oujda\\
  Morocco
  \And
  Ibrahim Ouahbi \\
  Faculty of Applied Sciences, Nador\\
  Mohammed First University, Oujda\\
  Morocco
}

\begin{document}

\maketitle

\begingroup
\renewcommand{\thefootnote}{}
\footnotetext{$^{*}$Equal contribution.}
\endgroup

\begin{abstract}
Model quantization is widely used to reduce memory, latency, and deployment cost, and is typically judged by whether predictive accuracy is preserved. In decision systems that provide algorithmic recourse, however, accuracy preservation is not sufficient: a small actionable change that flips the decision of a full-precision model may fail after quantization, or require a substantially larger intervention. This paper studies this deployment mismatch and introduces \emph{counterfactual sensitivity under quantization}, a framework for measuring how compression changes recourse behavior. We propose two metrics: \emph{Validity Drop} (VD), which measures the fraction of full-precision recourse actions that no longer achieve the target outcome after quantization, and \emph{Counterfactual Recourse Gap} (CRG), which measures the increase in minimal recourse cost under the quantized model. To mitigate this failure mode, we introduce \emph{Counterfactual-Faithful Quantization} (CFQ), a quantization-aware training method that jointly learns quantizer parameters and mixed-precision bit allocation while preserving the target prediction at teacher-generated recourse points. CFQ is compatible with standard LSQ/PACT-style quantizers and mixed-precision policies, and can also be instantiated as a training-free calibration procedure for post-training quantization. Experiments on \textsc{Adult}, \textsc{German Credit}, and \textsc{COMPAS} show that standard QAT and mixed-precision baselines can preserve accuracy while substantially degrading recourse stability. At matched accuracy and bit budget, CFQ consistently reduces VD and CRG; for example, on \textsc{Adult}, CFQ reduces VD/CRG from $0.121/0.162$ for an accuracy-centric mixed-precision baseline to $0.061/0.071$. Additional ablations, teacher-quality diagnostics, constraint-sensitivity studies, and non-tabular structured recourse experiments further show that the gains are not due to relaxed compression, solver artifacts, or over-budget precision. These results demonstrate that quantization should be evaluated not only by predictive fidelity, but also by whether it preserves the actionable decisions users rely on.
\end{abstract}

\section{Introduction}
\label{sec:intro}
Modern ML systems are increasingly deployed under strict efficiency constraints, where quantization is the default tool for reducing memory footprint and inference latency \cite{gholami2021survey_quantization,krishnamoorthi2018quant_whitepaper}. Quantization-aware training (QAT) and post-training quantization (PTQ) are typically evaluated through a single lens: \emph{does the quantized model preserve predictive accuracy?} \cite{nagel2020adaround,banner2019posttraining4bit}. Yet for decision-making systems that provide \emph{algorithmic recourse}, accuracy is not the only requirement. Recourse and counterfactual explanations seek a minimal \emph{actionable} change to an input that achieves a desired outcome, i.e., a smallest-change explanation or recommendation \cite{mothilal2020dice,poyiadzi2020face,wachter2018counterfactual,ustun2019actionable}. In high-stakes settings, this recommendation is not merely interpretability; it is a promise of \emph{what actions would work}. If the deployed model changes, that promise can break.

This paper highlights a deployment mismatch: quantization can preserve average-case accuracy while altering the \emph{local decision boundary geometry} that determines counterfactuals. The reason is simple: accuracy constrains predictions on the data distribution, whereas recourse depends on the geometry of the classifier in neighborhoods around individual points and along actionable directions. Quantization perturbs weights (and sometimes activations) in a structured but non-uniform way, which can shift margins and rotate decision boundaries in precisely those neighborhoods that matter for recourse. As a consequence, two models with nearly identical test accuracy can yield substantially different recourse actions for the same individual. This failure mode is particularly salient ``now'' because quantized deployment is no longer an edge-only niche; it is standard practice across mobile, on-device, and server inference pipelines, including mixed-precision policies that vary bitwidth across layers to meet strict budgets. If recourse is computed using an unquantized model during development, and the system is later quantized for deployment, recourse recommendations can become invalid or significantly more costly.

\begin{wrapfigure}[18]{r}{0.5\linewidth}
\vspace{-3mm}
\captionsetup{font=footnotesize,skip=2pt,width=\linewidth}
\centering
\includegraphics[width=0.51\textwidth]{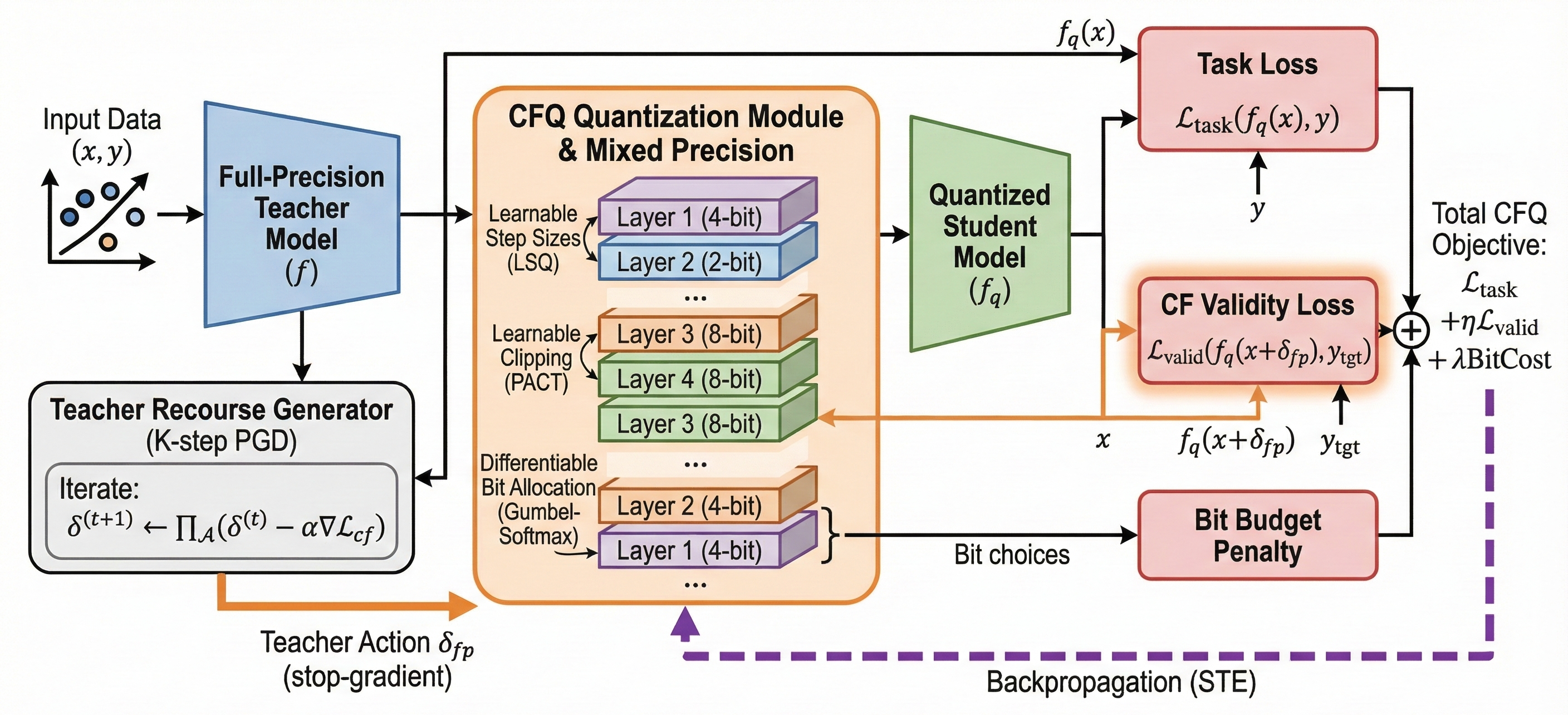}
\caption{\textbf{CFQ pipeline.} A teacher recourse action $\delta_{\mathrm{fp}}$ is computed on the full-precision model using a small number of projected gradient steps and is treated as a constant via stop-gradient. The quantized model $f_q$ is trained to preserve task accuracy and to keep $x+\delta_{\mathrm{fp}}$ a valid counterfactual under quantization, while mixed-precision bit allocation and quantizer parameters are optimized under a bit budget. \textbf{\textit{We do not claim a new recourse solver or a new quantizer; we introduce a
deployment objective that makes recourse validity and recourse cost explicit
constraints of quantization.}}}
\label{fig:cfq_pipeline}
\end{wrapfigure}

We therefore treat quantization explicitly as a \emph{model change} and ask a different question: how can we quantize while preserving counterfactual behavior as a first-class objective? Our approach draws on the algorithmic recourse framing \cite{ustun2019actionable}, the counterfactual ``smallest change'' perspective \cite{wachter2018counterfactual}, and the causal view that recourse corresponds to feasible interventions rather than arbitrary perturbations. Building on these foundations, we formalize \emph{counterfactual sensitivity under quantization} through stability of validity, cost, and direction, and we design a quantization procedure that explicitly optimizes for this stability (see Figure \pinkref{fig:cfq_pipeline}).

\paragraph{Contributions.}
First, we identify and quantify an empirical gap between accuracy preservation and recourse preservation: strong QAT/PTQ and mixed-precision baselines can match full-precision accuracy yet still substantially degrade counterfactual transferability, increasing the fraction of individuals whose pre-quantization recourse no longer works after quantization and inflating the minimal recourse cost. Second, we introduce \emph{Counterfactual-Faithful Quantization (CFQ)}, a quantization-aware training method that couples mixed-precision bit allocation with a counterfactual-consistency regularizer so that counterfactual points computed under the full-precision teacher remain valid under the quantized model. Third, we propose two metrics that make this failure mode measurable and comparable across methods: \emph{Validity Drop (VD)}, which captures the probability that pre-quantization recourse fails post-quantization, and \emph{Counterfactual Recourse Gap (CRG)}, which captures the relative increase in minimal recourse cost induced by quantization. Finally, we demonstrate that optimizing quantization for recourse stability yields improved VD/CRG at comparable accuracy and bit budgets on standard tabular recourse benchmarks, with an optional extension to structured domains when recourse actions are defined in a semantic edit space.

\vspace{-0.3cm}
\section{Related Work}
\label{sec:background}

\paragraph{Recourse and counterfactual explanations.}
Counterfactual explanations aim to answer ``what needs to change'' in an input to achieve a desired model outcome, typically by proposing a small, human-interpretable modification that flips the prediction \cite{wachter2018counterfactual}. Work on algorithmic recourse emphasizes that such modifications must be \emph{actionable}: recommended changes should respect immutability constraints (e.g., age), feasibility constraints (e.g., realistic ranges), and domain-specific notions of effort or preference \cite{ustun2019actionable,karimi2020survey}. This line of research has made clear that counterfactual recommendations are not only about proximity, but also about the plausibility and implementability of the suggested actions, which often requires careful constraint design and cost modeling. Importantly for our setting, recent discussions highlight that recourse may become unreliable when the underlying predictive model changes or is uncertain, motivating notions of recourse reliability and stability beyond a single fixed predictor \cite{rawal2020recourse}.

\begin{wrapfigure}[17]{r}{0.45\linewidth}
\vspace{-3mm}
\captionsetup{font=footnotesize,skip=2pt,width=\linewidth}
\centering
\includegraphics[width=0.47\textwidth]{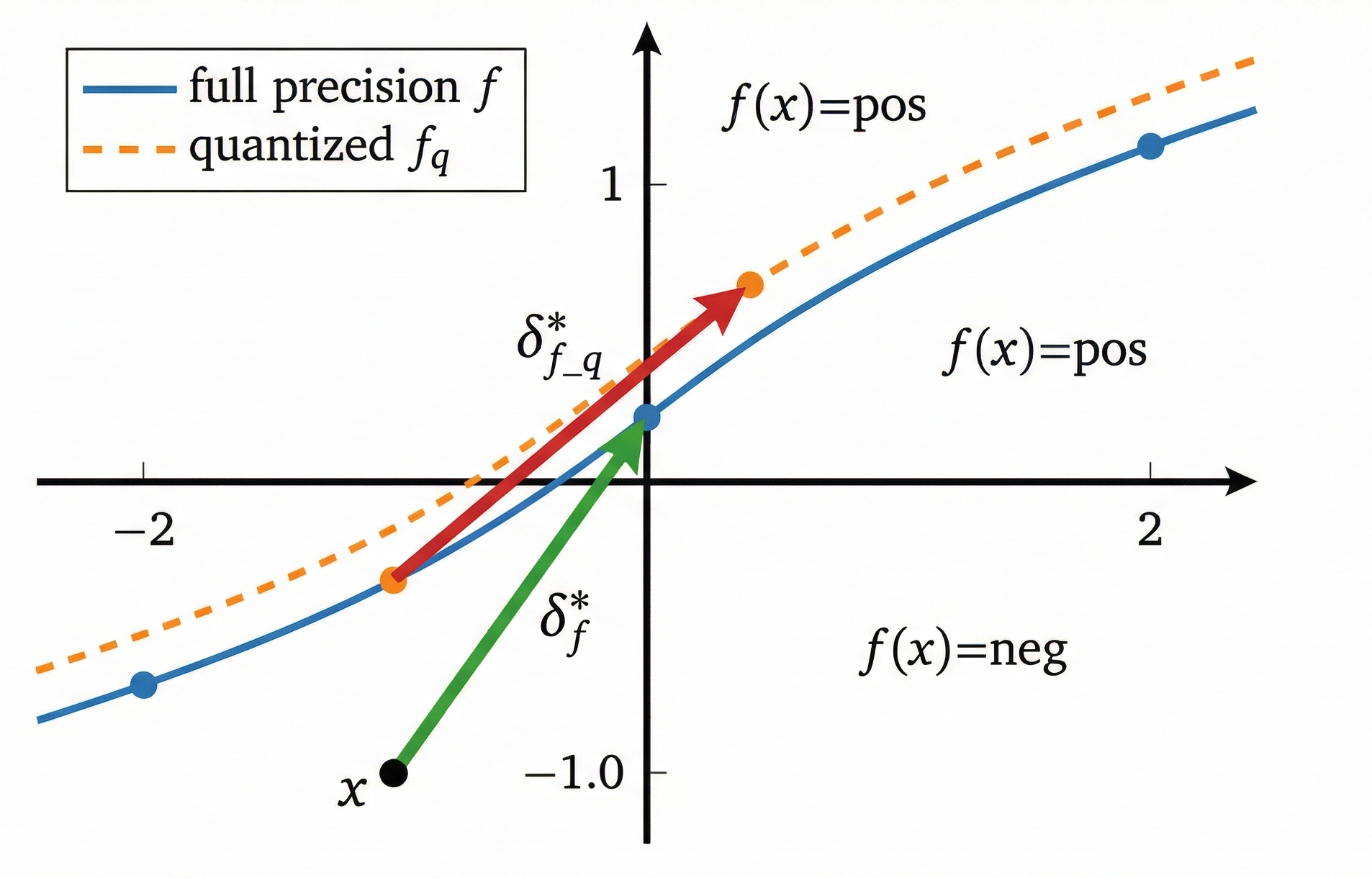}
\caption{\textbf{Same accuracy, different recourse.} Quantization can shift local decision geometry: a minimal actionable recourse $\delta_f^\star$ that flips the full-precision model may fail for the quantized model, requiring a different or larger action $\delta_{f_q}^\star$ even when predictive accuracy is unchanged.}
\label{fig:same_acc_diff_recourse}
\end{wrapfigure}

\paragraph{Robustness of recourse to model change.}
A closely related literature studies how recourse degrades under model updates such as retraining, distribution shift, or the existence of multiple equally accurate models that behave differently for individuals. This phenomenon is often framed through \emph{predictive/model multiplicity}, where a prediction task admits many comparably accurate predictors with conflicting decisions, implying that the associated recourse recommendations can vary widely across equally plausible deployed models \cite{marx2020predictive,black2022modelmultiplicity,pawelczyk2020multiplicity}. Complementing this view, empirical and theoretical evidence shows that recourse computed for a fixed model can become invalid after realistic data- and model-shift events, motivating stability as a first-class requirement. To address these issues, robust recourse methods typically modify the recourse generation procedure itself, designing recommendations that remain valid across sets or distributions of plausible future models (e.g., via adversarial or distributionally robust formulations) \cite{upadhyay2021roar,nguyen2023dirrac}. Related work also studies robustness of counterfactual explanations under model multiplicity through formal guarantees across multiple models \cite{leofante2023cfmultiplicity}. While these approaches explicitly acknowledge that models evolve, they generally treat the deployment pipeline as given and focus on hedging in the optimization that produces recourse. Our work is complementary but distinct: we study a ubiquitous deployment transformation model quantization and ask how to design quantization so that recourse remains stable after compression (see Fig  \pinkref{fig:same_acc_diff_recourse}).

\paragraph{Quantization and mixed precision.}
Quantization reduces inference cost by representing weights and activations with low-bit numerical formats, and quantization-aware training (QAT) incorporates low-precision effects during learning to improve accuracy at deployment bitwidths \cite{jacob2018quantization}. Representative QAT approaches include LSQ, which learns layerwise step sizes to adapt uniform quantizers to task and activation statistics, and clipping-based methods that learn activation ranges to reduce saturation and improve low-bit fidelity \cite{esser2020lsq,choi2018pact}. Mixed-precision quantization goes further by assigning different bitwidths across layers under a resource budget, using either hardware-aware policy search or sensitivity/curvature proxies to identify layers whose precision most affects performance \cite{wang2019haq,dong2019hawq}. Recent post-training quantization (PTQ) methods for large models also highlight that quantization error is highly non-uniform across channels and layers, motivating data-driven layer/channel selection and scaling strategies \cite{frantar2022gptq,lin2023awq}. Despite strong progress, these methods primarily optimize predictive performance under resource constraints, and do not explicitly target the stability of counterfactual recourse as an objective. \textbf{\textit{To our knowledge, no prior work optimizes quantization including both bit allocation and quantizer parameters for recourse stability as a first-class objective.}}

\vspace{-0.2cm}
\section{Problem Setup and Metrics}
\label{sec:problem}

\subsection{Setup}
\label{sec:setup}

We study a predictor deployed in full precision and after quantization. Let $x\in\mathbb{R}^{d}$ denote an input with $d$ features, and let $y\in\mathcal{Y}$ denote its label. The full-precision model is denoted by $f:\mathbb{R}^{d}\rightarrow\mathcal{Y}$, while the quantized model is denoted by $f_q:\mathbb{R}^{d}\rightarrow\mathcal{Y}$. We view quantization as a deterministic transformation of the model parameters and, optionally, activations; in particular, we write $f_q=\mathcal{Q}(f;\phi,b)$, where $\mathcal{Q}(\cdot)$ denotes the quantization operator, $\phi$ collects quantizer parameters such as per-layer step sizes, clipping thresholds, or zero-points, and $b$ denotes the bitwidth specification, either as a scalar bitwidth or as a layer-wise vector of bitwidths in mixed precision. This mapping induces a model change $f\mapsto f_q$ that may preserve predictive accuracy while modifying the local geometry of the decision function.

We encode feasible and actionable changes through an \emph{action set} $\mathcal{A}(x)\subseteq\mathbb{R}^{d}$ that specifies which perturbations $\delta$ are permitted for the individual represented by $x$. A generic form that captures immutability, bounds, and sparsity is $\mathcal{A}(x)=\{\delta\in\mathbb{R}^{d}:\delta_j=0\ \forall j\in\mathcal{I},\ \ell\le x+\delta\le u,\ \|\delta\|_0\le k\}$. Here, $\mathcal{I}\subseteq\{1,\dots,d\}$ indexes immutable features, meaning that $\delta_j=0$ for all $j\in\mathcal{I}$; $\ell,u\in\mathbb{R}^{d}$ are lower and upper feasibility bounds enforced componentwise; $\|\delta\|_0$ counts the number of nonzero coordinates of $\delta$; and $k\in\mathbb{N}$ controls a sparsity budget that promotes interpretable actions.

Given a target outcome $y_{\mathrm{tgt}}\in\mathcal{Y}$, algorithmic recourse seeks a minimal-cost action $\delta$ that changes the model outcome to $y_{\mathrm{tgt}}$ \cite{ustun2019actionable}. We define the recourse for the full-precision model as $\delta_f^\star(x;y_{\mathrm{tgt}})=\arg\min_{\delta\in\mathcal{A}(x)}c(\delta)\ \text{s.t.}\ f(x+\delta)=y_{\mathrm{tgt}}$, and analogously for the quantized model as $\delta_q^\star(x;y_{\mathrm{tgt}})=\arg\min_{\delta\in\mathcal{A}(x)}c(\delta)\ \text{s.t.}\ f_q(x+\delta)=y_{\mathrm{tgt}}$. In both definitions, the function $c:\mathbb{R}^{d}\rightarrow\mathbb{R}_{\ge 0}$ quantifies the effort required to implement an action. A standard choice that makes feature-wise costs explicit is $c(\delta)=\|W\delta\|_p$, where $W\in\mathbb{R}^{d\times d}$ is a diagonal matrix with positive entries encoding per-feature effort, and $p\in\{1,2\}$ controls whether the cost prefers sparse changes when $p=1$ or smoother changes when $p=2$.

\subsection{Counterfactual sensitivity under quantization}
\label{sec:sensitivity}
We formalize \emph{counterfactual sensitivity under quantization} as the extent to which quantization changes the recourse solutions. We consider three complementary notions.

First, \emph{validity stability} measures whether a recourse computed before quantization remains valid after quantization. Concretely, the pre-quantization action $\delta_f^\star(x;y_{\mathrm{tgt}})$ is validity-stable if it also achieves the target outcome for the quantized model, meaning $f_q(x+\delta_f^\star(x;y_{\mathrm{tgt}}))=y_{\mathrm{tgt}}$ (or, equivalently, that it induces the same target decision under $f_q$).

Second, \emph{cost stability} measures whether the minimal effort required to achieve $y_{\mathrm{tgt}}$ changes significantly. This compares the optimal costs $c(\delta_f^\star(x;y_{\mathrm{tgt}}))$ and $c(\delta_q^\star(x;y_{\mathrm{tgt}}))$ across models, with stability corresponding to small absolute or relative deviations.

Third, \emph{direction stability} measures whether the recommended action itself remains similar. This compares $\delta_f^\star(x;y_{\mathrm{tgt}})$ and $\delta_q^\star(x;y_{\mathrm{tgt}})$ either geometrically (e.g., via cosine similarity) or combinatorially (e.g., via overlap of the changed feature set when sparsity constraints are present).

\subsection{Metrics}
\label{sec:metrics}

We now define metrics that operationalize the above notions and are designed to be complementary to predictive accuracy and compression. The first metric is \emph{Validity Drop (VD)}, which quantifies how often a recourse computed on the full-precision model fails to transfer to the quantized model. We define it as $\mathrm{VD}=\mathbb{P}_{x}\!\left[f(x+\delta_f^\star(x;y_{\mathrm{tgt}}))\neq f_q(x+\delta_f^\star(x;y_{\mathrm{tgt}}))\right]$, where the probability is taken over the evaluation distribution of $x$. The event inside the brackets captures a \emph{label mismatch at the recourse point}; equivalently, when the target label $y_{\mathrm{tgt}}$ is explicitly specified, one may measure the target-class failure rate $\mathbb{P}_x[f_q(x+\delta_f^\star(x;y_{\mathrm{tgt}}))\neq y_{\mathrm{tgt}}]$.

The second metric is the \emph{Counterfactual Recourse Gap (CRG)}, which measures relative changes in minimal recourse cost induced by quantization. We define it as $\mathrm{CRG}=\mathbb{E}_{x}\!\left[\frac{c(\delta_q^\star(x;y_{\mathrm{tgt}}))-c(\delta_f^\star(x;y_{\mathrm{tgt}}))}{c(\delta_f^\star(x;y_{\mathrm{tgt}}))+\varepsilon}\right]$. Here, $\mathbb{E}_x[\cdot]$ denotes expectation over inputs, and $\varepsilon>0$ is a small constant that prevents division by zero when the baseline cost is very small. Positive values of $\mathrm{CRG}$ indicate that quantization increases the minimal effort needed to achieve the target outcome, while negative values indicate decreased effort.

To quantify direction stability, we report a cosine-based similarity between the pre- and post-quantization actions, defined as $\mathrm{DirSim}=\mathbb{E}_{x}\!\left[\frac{\langle \delta_f^\star(x;y_{\mathrm{tgt}}),\delta_q^\star(x;y_{\mathrm{tgt}})\rangle}{\|\delta_f^\star(x;y_{\mathrm{tgt}})\|_2\|\delta_q^\star(x;y_{\mathrm{tgt}})\|_2+\varepsilon}\right]$, where $\langle \cdot,\cdot\rangle$ is the Euclidean inner product, $\|\cdot\|_2$ is the $\ell_2$ norm, and $\varepsilon>0$ stabilizes the ratio when norms are small. When sparsity is central, we additionally consider overlap of changed coordinates through the support operator $\mathrm{supp}_\tau(\delta)=\left\{j\in\{1,\dots,d\}:|\delta_j|>\tau\right\}$, where $\tau\ge 0$ is a threshold used to ignore negligible numerical changes. Using this support operator, the expected Jaccard overlap is defined as $\mathrm{ActOverlap}=\mathbb{E}_{x}\!\left[\frac{\left|\mathrm{supp}_\tau(\delta_f^\star(x;y_{\mathrm{tgt}}))\cap\mathrm{supp}_\tau(\delta_q^\star(x;y_{\mathrm{tgt}}))\right|}{\left|\mathrm{supp}_\tau(\delta_f^\star(x;y_{\mathrm{tgt}}))\cup\mathrm{supp}_\tau(\delta_q^\star(x;y_{\mathrm{tgt}}))\right|+\varepsilon}\right]$, where $|\cdot|$ denotes set cardinality and $\varepsilon>0$ avoids undefined values when both supports are empty.




\section{Proposed Method}
\label{sec:method}

\subsection{Core idea}
\label{sec:coreidea}

CFQ treats post-training quantization as a structured model shift that can preserve predictive accuracy while altering the geometry that determines actionable counterfactuals. Let $f$ denote the full-precision model and let $f_q$ denote the quantized model obtained by inserting quantizers into weights and/or activations and choosing a possibly mixed-precision bit allocation. CFQ enforces that a recourse action computed with the full-precision model remains effective after quantization. To this end, CFQ introduces a \emph{teacher recourse} action $\delta_{\mathrm{fp}}(x;y_{\mathrm{tgt}})$ computed on the full-precision model and uses it as a supervision signal for quantization.

In principle, $\delta_{\mathrm{fp}}(x;y_{\mathrm{tgt}})$ could be the exact solution of the constrained recourse problem. In practice, repeatedly solving an inner optimization to high accuracy is computationally expensive. CFQ therefore computes $\delta_{\mathrm{fp}}$ using a small number $K$ of projected gradient steps on a differentiable surrogate objective and then \emph{stops gradient} through this inner procedure when optimizing quantization. More precisely, let $\mathcal{L}_{\mathrm{cf}}(x,\delta;y_{\mathrm{tgt}},f)$ denote a differentiable counterfactual loss whose minimization encourages $f(x+\delta)=y_{\mathrm{tgt}}$. For classification, a standard choice is the cross-entropy with the target class evaluated at the perturbed input. Starting from $\delta^{(0)}=0$, the teacher recourse action is approximated through the projected-gradient recursion $\delta^{(t+1)}=\Pi_{\mathcal{A}(x)}\!\left(\delta^{(t)}-\alpha\nabla_{\delta}\mathcal{L}_{\mathrm{cf}}\!\left(x,\delta^{(t)};y_{\mathrm{tgt}},f\right)\right)$ for $t=0,1,\dots,K-1$, followed by $\delta_{\mathrm{fp}}(x;y_{\mathrm{tgt}})=\mathrm{stopgrad}\!\left(\delta^{(K)}\right)$.

Here, $\alpha>0$ is the projected-gradient step size, $\nabla_\delta$ denotes differentiation with respect to the action variable $\delta$, and $\Pi_{\mathcal{A}(x)}(\cdot)$ is the projection operator onto the actionable set $\mathcal{A}(x)$, which enforces immutability, feasibility bounds, and any sparsity constraints as defined in Section~\pinkref{sec:setup}. The operator $\mathrm{stopgrad}(\cdot)$ means that the outer quantization optimization treats $\delta_{\mathrm{fp}}$ as a constant, thereby preventing backpropagation through the inner $K$-step recourse construction. This design keeps compute manageable: CFQ adds only $K$ forward/backward passes for the teacher recourse computation and avoids bilevel second-order effects associated with differentiating through the inner optimization.

\vspace{-0.3cm}
\subsection{CFQ objective}
\label{sec:loss}

CFQ learns quantization to preserve recourse validity while maintaining task accuracy under a bit budget. Let $\mathcal{L}_{\mathrm{task}}(f_q(x),y)$ denote the standard supervised objective, such as cross-entropy for classification, evaluated on the quantized model. CFQ augments this objective with a \emph{validity preservation} term that forces the quantized model to assign the target outcome at the teacher-recourse point $x+\delta_{\mathrm{fp}}(x;y_{\mathrm{tgt}})$. The simplest strong CFQ objective is defined as $\mathcal{L}_{\mathrm{CFQ}}=\mathbb{E}_{(x,y)}\!\left[\mathcal{L}_{\mathrm{task}}(f_q(x),y)+\eta\,\mathcal{L}_{\mathrm{valid}}\!\left(f_q(x+\delta_{\mathrm{fp}}(x;y_{\mathrm{tgt}})),y_{\mathrm{tgt}}\right)\right]+\lambda\,\mathrm{BitCost}$, where $\eta\ge 0$ controls the strength of counterfactual validity preservation and $\lambda\ge 0$ trades off predictive performance against resource usage. The term $\mathcal{L}_{\mathrm{valid}}(\cdot,\cdot)$ can be instantiated as cross-entropy with the target label, so that the quantized model is explicitly trained to satisfy the counterfactual constraint at $x+\delta_{\mathrm{fp}}$. The expectation $\mathbb{E}_{(x,y)}[\cdot]$ is taken over the training distribution, and $y_{\mathrm{tgt}}$ is either specified by the application, for example the favorable outcome, or chosen as a function of $(x,y)$, for example the opposite class in binary classification.

We set $\eta$, the validity preservation weight, and $\lambda$, the bit-cost trade-off, by a small validation sweep. Specifically, we grid search $\eta\in\{0.25,0.5,1,2\}$ and $\lambda\in\{10^{-4},10^{-3},10^{-2},10^{-1}\}$, selecting the pair that maximizes validation accuracy subject to the bit-budget target. Equivalently, we minimize $\mathcal{L}_{\mathrm{task}}+\eta\,\mathcal{L}_{\mathrm{valid}}$ among models whose expected BitCost is within a small tolerance of $\mathcal{B}_{\mathrm{tot}}$. Unless stated otherwise, we use $\eta=1$ and choose $\lambda$ to match the desired budget, where a higher $\lambda$ yields lower BitCost.

CFQ can incorporate additional terms that improve cost and direction stability without significantly increasing computational complexity. A first optional add-on matches the teacher action to a \emph{student action} $\delta_q(x;y_{\mathrm{tgt}})$ computed with the quantized model using the same $K$ projected steps but replacing $f$ by $f_q$, namely $\delta_q(x;y_{\mathrm{tgt}})=\Pi_{\mathcal{A}(x)}^{K}\!\left(0;x,y_{\mathrm{tgt}},f_q\right)$. In contrast to the teacher action $\delta_{\mathrm{fp}}$, which is detached, we do not stop gradients through $\delta_q$ when $\mathcal{L}_{\mathrm{match}}$ is enabled. Consequently, $\mathcal{L}_{\mathrm{match}}$ backpropagates into the quantized model parameters, including quantizer parameters and the mixed-precision policy, through the unrolled $K$ projected steps; we therefore use small values of $K$, typically $1$--$3$, to keep this overhead modest. CFQ then adds a direction/cost matching regularizer defined as $\mathcal{L}_{\mathrm{match}}=\mathbb{E}_{x}\!\left[\alpha_1\|\delta_q(x;y_{\mathrm{tgt}})-\delta_{\mathrm{fp}}(x;y_{\mathrm{tgt}})\|_1+\alpha_2\left|c(\delta_q(x;y_{\mathrm{tgt}}))-c(\delta_{\mathrm{fp}}(x;y_{\mathrm{tgt}}))\right|\right]$, where $\alpha_1,\alpha_2\ge 0$ balance direction alignment and cost alignment, $\|\cdot\|_1$ promotes sparse agreement in actionable coordinates, and $c(\cdot)$ is the recourse cost function.

A second optional add-on stabilizes the decision boundary around the counterfactual point using a hinge margin, as discussed in Appendix~\pinkref{app:grad_flow_match}. For multi-class classification with logits $g_q(x)\in\mathbb{R}^{|\mathcal{Y}|}$, we define the target margin at $x$ as $m_q(x;y_{\mathrm{tgt}})=g_q(x)_{y_{\mathrm{tgt}}}-\max_{y'\neq y_{\mathrm{tgt}}}g_q(x)_{y'}$, where $g_q(x)_y$ denotes the logit corresponding to class $y$, and the maximization over $y'\neq y_{\mathrm{tgt}}$ identifies the strongest competing class. We enforce a margin threshold $\gamma>0$ at the teacher-recourse point through the hinge penalty $\mathcal{L}_{\mathrm{hinge}}=\mathbb{E}_{x}\!\left[\max\!\left(0,\gamma-m_q(x+\delta_{\mathrm{fp}}(x;y_{\mathrm{tgt}});y_{\mathrm{tgt}})\right)\right]$. This hinge term encourages a buffer at the counterfactual point, which reduces the chance that quantization-induced perturbations flip the outcome. When used, the full objective becomes $\mathcal{L}_{\mathrm{CFQ}}+\mathcal{L}_{\mathrm{match}}+\beta\,\mathcal{L}_{\mathrm{hinge}}$ for some $\beta\ge 0$.

\vspace{-0.3cm}
\subsection{Mixed-precision bit allocation}
\label{sec:mixedprecision}

A key component of CFQ is mixed-precision quantization, which allocates more bits to layers that most affect counterfactual validity while compressing less critical layers. Let $L$ denote the number of quantized layers, and let $b_\ell\in\mathcal{B}$ denote the bitwidth chosen for layer $\ell$, where the candidate bit set is $\mathcal{B}=\{2,3,4,8\}$. CFQ learns the layer-wise bitwidths $\{b_\ell\}_{\ell=1}^{L}$ jointly with the quantizer parameters under a differentiable relaxation. Let $\pi_\ell\in\Delta^{|\mathcal{B}|}$ denote a categorical distribution over candidate bitwidths for layer $\ell$, where $\Delta^{|\mathcal{B}|}$ is the probability simplex. A Gumbel-Softmax relaxation produces a differentiable sample $z_\ell\in\Delta^{|\mathcal{B}|}$, whose $r$-th component is given by $z_{\ell,r}=\frac{\exp\!\left((\log \pi_{\ell,r}+g_{\ell,r})/\tau\right)}{\sum_{r'}\exp\!\left((\log \pi_{\ell,r'}+g_{\ell,r'})/\tau\right)}$, with $g_{\ell,r}\sim\mathrm{Gumbel}(0,1)$. Here, $r$ indexes the candidate bitwidths in $\mathcal{B}$, $\tau>0$ is the Gumbel-Softmax temperature, and $g_{\ell,r}$ are independent Gumbel perturbations. As $\tau\rightarrow 0$, $z_\ell$ concentrates on a one-hot vector and therefore approximates discrete bit selection. In practice, a straight-through estimator can be used by taking a hard one-hot decision in the forward pass while backpropagating through the soft vector $z_\ell$.

The effective bitwidth of layer $\ell$ is written as the expectation under the relaxed selection vector, namely $\bar b_\ell=\sum_r z_{\ell,r}b_r$, where $b_r$ denotes the $r$-th candidate bitwidth in $\mathcal{B}$. In implementations that require an integer bitwidth for the actual quantizer, we set $b_\ell=\arg\max_r z_{\ell,r}$ in the forward pass and use $\bar b_\ell$ only for gradient propagation and cost accounting. The resource penalty $\mathrm{BitCost}$ in the CFQ objective can then be instantiated as a parameter-count-weighted sum, $\mathrm{BitCost}=\sum_{\ell=1}^{L}n_\ell\bar b_\ell$, where $n_\ell$ is the number of parameters in layer $\ell$. Alternatively, to better reflect hardware compute, one can use a bit-operations proxy that accounts for both weight and activation precision, namely $\mathrm{BOPs}=\sum_{\ell=1}^{L}\mathrm{MACs}_\ell\bar b^{(w)}_\ell\bar b^{(a)}_\ell$, where $\mathrm{MACs}_\ell$ is the number of multiply-accumulate operations in layer $\ell$, and $\bar b^{(w)}_\ell$ and $\bar b^{(a)}_\ell$ are the effective bitwidths for weights and activations.

The motivation for mixed precision is that quantization error is heterogeneous across layers: some layers have disproportionately large influence on decision-boundary geometry and therefore on whether $x+\delta_{\mathrm{fp}}$ remains a valid counterfactual under $f_q$. Mixed-precision quantization has been successfully explored for hardware-aware and sensitivity-aware compression \cite{wang2019haq,dong2020hawqv2}, and CFQ leverages this capability with a counterfactual objective so that bit allocation is shaped by recourse preservation rather than accuracy alone.

To make the joint optimization between the mixed-precision policy and the quantizer fully specified, we maintain \emph{bit-specific} quantizer parameters for each layer. That is, for each layer $\ell$ and candidate bitwidth $b_r\in\mathcal{B}$, we learn a separate set of quantizer parameters, such as the LSQ step size $s_{\ell,r}$ for weights and the PACT clipping threshold $\alpha_{\ell,r}$ for activations. During training, we use a straight-through Gumbel-Softmax estimator: in the forward pass, we take the hard selection $r^*=\arg\max_r z_{\ell,r}$ and quantize layer $\ell$ using the bitwidth $b_{r^*}$ together with its associated parameters $(s_{\ell,r^*},\alpha_{\ell,r^*})$. In the backward pass, gradients are propagated to the bit-policy parameters $\pi_\ell$ through the soft vector $z_\ell$, while $(s_{\ell,r^*},\alpha_{\ell,r^*})$ receive standard straight-through-estimator gradients from the quantizer. At convergence, we discretize the learned policy as $b_\ell=\arg\max_r\pi_{\ell,r}$ and retain the corresponding quantizer parameters for deployment.
\vspace{-0.3cm}
\subsection{Quantizer parameterization}
\label{sec:quantparam}

CFQ is compatible with standard differentiable quantizers, and we instantiate it with widely used mechanisms for weights and activations. For weights, LSQ-style step-size learning uses a learnable step size $s_{\ell,r}>0$ for each layer $\ell$ and candidate bitwidth $b_r$. Given a hard-selected index $r^*=\arg\max_r z_{\ell,r}$, the weight tensor is quantized as $\hat w_\ell=s_{\ell,r^*}\cdot \mathrm{clip}\!\left(\left\lfloor w_\ell/s_{\ell,r^*}\right\rceil,q^{-}(b_{r^*}),q^{+}(b_{r^*})\right)$, where $\lfloor\cdot\rceil$ denotes rounding to the nearest integer and $\mathrm{clip}(\cdot,q^-,q^+)$ clamps the rounded values to the integer range determined by the selected bitwidth. For symmetric signed integer quantization, we use the standard bounds $q^{+}(b)=2^{b-1}-1$ and $q^{-}(b)=-2^{b-1}$, which gives, for example, the range $[-8,7]$ when $b=4$. The step size $s_{\ell,r}$ adapts during training to match the distribution of $w_\ell$ at the target precision.

For activations, PACT introduces a learnable clipping threshold $\alpha_{\ell,r}>0$ for each layer $\ell$ and candidate activation bitwidth $b_r$. With the same hard-selected index $r^*=\arg\max_r z_{\ell,r}$, we first compute the clipped activation $\tilde a_\ell=\mathrm{clip}(a_\ell,0,\alpha_{\ell,r^*})$, followed by low-bit activation quantization $\hat a_\ell=Q_{b_{r^*}}(\tilde a_\ell;s^{(a)}_{\ell,r^*},z^{(a)}_{\ell,r^*})$. Here, $Q_b(\cdot;s,z)$ denotes an affine uniform quantizer with activation step size $s^{(a)}_{\ell,r}$ and zero-point $z^{(a)}_{\ell,r}$. The clipping threshold $\alpha_{\ell,r}$ is learned jointly with the model parameters, quantizer parameters, and mixed-precision policy in order to reduce activation saturation, improve low-bit fidelity, and ensure that the resulting quantized model $f_q$ preserves both task performance and counterfactual validity at the teacher-recourse points.

\vspace{-0.3cm}
\section{Theory: Margin Robustness of Recourse under Quantization}
\label{sec:theory}

We provide a simple sufficient condition under which a recourse action computed for a full-precision model remains valid after quantization. The analysis links \emph{Validity Drop} to the classification margin at the counterfactual point and motivates CFQ's objective, which explicitly encourages the quantized model to preserve the target outcome at such points and, optionally, to enforce an explicit margin.

Let $g:\mathbb{R}^{d}\rightarrow\mathbb{R}^{|\mathcal{Y}|}$ denote the full-precision logit map associated with $f$, and let $g_q:\mathbb{R}^{d}\rightarrow\mathbb{R}^{|\mathcal{Y}|}$ denote the logit map associated with the quantized model $f_q$. For an input $u\in\mathbb{R}^{d}$ and a target label $y_{\mathrm{tgt}}\in\mathcal{Y}$, we define the multi-class target margin of the full-precision logits as $m_f(u;y_{\mathrm{tgt}})=g(u)_{y_{\mathrm{tgt}}}-\max_{y'\neq y_{\mathrm{tgt}}}g(u)_{y'}$. Here, $g(u)_y$ denotes the logit coordinate corresponding to class $y$, and the maximum over $y'\neq y_{\mathrm{tgt}}$ selects the strongest competing class. Analogously, the quantized margin is defined as $m_{f_q}(u;y_{\mathrm{tgt}})=g_q(u)_{y_{\mathrm{tgt}}}-\max_{y'\neq y_{\mathrm{tgt}}}g_q(u)_{y'}$. For standard argmax classifiers, the event $f_q(u)=y_{\mathrm{tgt}}$ is equivalent to $m_{f_q}(u;y_{\mathrm{tgt}})>0$, so lower bounding the quantized margin provides a direct guarantee of counterfactual validity under quantization.

We model quantization as a bounded perturbation of the logit function in a neighborhood around the counterfactual point. Let $u^\star=x+\delta$ denote a candidate counterfactual point for an input $x$ and action $\delta$. For a radius $\rho>0$, define the neighborhood $\mathbb{B}_\rho(u^\star)=\{u\in\mathbb{R}^{d}:\|u-u^\star\|_2\le\rho\}$. We assume that quantization induces a uniform logit perturbation bounded by $\varepsilon$ in that neighborhood, namely $\sup_{u\in\mathbb{B}_\rho(u^\star)}\|g_q(u)-g(u)\|_\infty\le\varepsilon$, where $\|\cdot\|_\infty$ denotes the maximum absolute coordinate value. This condition states that every class logit changes by at most $\varepsilon$ in magnitude due to quantization for all inputs near the counterfactual point.

\begin{proposition}[Margin robustness at the recourse point]
\label{prop:margin_robust}
Fix $x\in\mathbb{R}^d$ and a target label $y_{\mathrm{tgt}}\in\mathcal{Y}$. Let $\delta$ be any feasible action and let $u^\star=x+\delta$. Suppose the bounded perturbation condition holds for some $\varepsilon\ge 0$ on $\mathbb{B}_\rho(u^\star)$. If the full-precision target margin at $u^\star$ satisfies
\begin{equation}
m_f(u^\star;y_{\mathrm{tgt}}) \;>\; 2\varepsilon,
\label{eq:margin_condition}
\end{equation}
then the quantized model predicts the target label at the same point, i.e.,
\begin{equation}
f_q(u^\star) \;=\; y_{\mathrm{tgt}},
\label{eq:validity_conclusion}
\end{equation}
equivalently $m_{f_q}(u^\star;y_{\mathrm{tgt}})>0$.
\end{proposition}
\vspace{-0.15cm}
Proposition~\pinkref{prop:margin_robust} formalizes a direct mechanism by which quantization can cause validity failures: even if $\delta$ is a valid recourse for the full-precision model, quantization may flip the argmax at $u^\star$ if the teacher margin $m_f(u^\star;y_{\mathrm{tgt}})$ is small relative to the induced perturbation level. The condition shows that counterfactual validity is preserved whenever the teacher margin at the recourse point dominates the worst-case logit deviation introduced by quantization. This motivates two complementary design principles reflected in CFQ. First, the validity-preservation term in the CFQ objective explicitly trains $f_q$ so that the target label remains preferred at $x+\delta_{\mathrm{fp}}$, which pushes $m_{f_q}(x+\delta_{\mathrm{fp}};y_{\mathrm{tgt}})$ positive and empirically reduces Validity Drop. Second, the optional hinge-margin stabilization term increases the margin at counterfactual points, effectively strengthening the safety buffer and making the prediction less sensitive to quantization noise. A full proof of Proposition~\pinkref{prop:margin_robust} and practical procedures to estimate or upper bound $\varepsilon$ are provided in the appendix \pinkref{app:theory_proofs}.

\vspace{-0.3cm}
\section{Experiments}
\label{sec:experiments}

\subsection{Datasets and recourse constraints}
\label{sec:datasets}
We evaluate CFQ on three tabular benchmarks that are standard in recourse and fairness studies: \textsc{Adult} (Census Income), \textsc{German Credit} (Statlog), and \textsc{COMPAS}. \textsc{Adult} and \textsc{German Credit} are distributed via the UCI Machine Learning Repository \cite{asuncion2007uci,beckerkohavi1996adult,hofmann1994german}, while \textsc{COMPAS} is drawn from the ProPublica COMPAS analysis release and has become a canonical benchmark for recidivism prediction and fairness evaluation \cite{propublica2016compas,dressel2018accuracy}. For each dataset, we follow the common recourse protocol in which the target outcome $y_{\mathrm{tgt}}$ corresponds to the favorable decision and the action set $\mathcal{A}(x)$ encodes implementable changes consistent with real-world constraints \cite{ustun2019actionable,karimi2020survey}. For \textsc{Adult}, we treat demographic attributes (e.g., sex and race) and age as immutable, and allow actions on attributes such as education, hours-per-week, and financial variables subject to dataset-specific bounds; categorical changes are restricted to valid category transitions and continuous features are restricted to feasible ranges. For \textsc{German Credit}, we treat immutable attributes (e.g., personal status and age) as fixed and allow actions on financial and credit-related features within realistic bounds, with sparsity constraints to encourage parsimonious recourse. For \textsc{COMPAS}, we treat protected attributes and age as immutable and permit actions on a limited subset of actionable covariates while enforcing feasibility constraints that preserve valid feature domains. Across all datasets, we use the same recourse cost $c(\delta)$ and the same action-set template, with dataset-specific immutable sets and bounds; the exact immutable/actionable partitions, categorical handling, and projection operators $\Pi_{\mathcal{A}(x)}$ are reported in Appendix~\pinkref{app:recourse_plan}.
\vspace{-0.3cm}
\subsection{Baselines}
\label{sec:baselines}
We compare CFQ to strong quantization baselines that match our architectures and training budgets. The first baseline is uniform-bit quantization-aware training (QAT) using Learned Step Size Quantization (LSQ) \cite{esser2020lsq}. The second baseline is a PACT-style QAT variant \cite{choi2018pact}. The third baseline is a mixed-precision quantization method that allocates per-layer bits under the same global budget as CFQ, guided by standard accuracy- or sensitivity-driven criteria, following the spirit of hardware-aware and Hessian/sensitivity-aware mixed precision approaches \cite{wang2019haq,dong2020hawqv2}.

\vspace{-0.3cm}
\subsection{Evaluation protocol and metrics}
\label{sec:eval_protocol}
We report predictive accuracy and compression in addition to recourse stability metrics defined in Section~\pinkref{sec:metrics}. Specifically, we compute Validity Drop as the fraction of test points for which the full-precision recourse action fails to transfer to the quantized model, and Counterfactual Recourse Gap as the relative change in minimal recourse cost after quantization. The recourse actions for evaluation are computed with the same constrained solver for both $f$ and $f_q$ to ensure a fair comparison; solver settings and convergence tolerances are provided in Appendix~\pinkref{app:recourse_plan}.

\subsection{Main results}
\label{sec:main_results}

Table~\pinkref{tab:main_results} compares predictive performance and recourse stability after quantization on \textsc{Adult}, \textsc{German Credit}, and \textsc{COMPAS} at a representative bit budget.

\providecommand{\meanstd}[2]{%
  \ensuremath{#1{\scriptstyle\,\pm\,#2}}%
}

\providecommand{\bestmeanstd}[2]{%
  \textcolor{bestGreen}{%
    \ensuremath{%
      \mathbf{#1}%
      {\scriptstyle\,\boldsymbol{\pm}\,\mathbf{#2}}%
    }%
  }%
}

\begin{table}[htbp]
\centering

\caption{\textbf{Recourse stability reveals failures that accuracy hides.}
Predictive accuracy and recourse stability after quantization on three
tabular recourse benchmarks at a representative bit budget. While all
quantized baselines match FP32 accuracy, they substantially degrade
counterfactual transferability (higher VD) and increase recourse effort
(higher CRG). CFQ preserves accuracy while dramatically improving
recourse validity and cost stability.}

\label{tab:main_results}

\small
\setlength{\tabcolsep}{5pt}
\renewcommand{\arraystretch}{1.30}
\arrayrulecolor{tableRule}

\resizebox{\linewidth}{!}{%
\begin{tabular}{lcccccc}

\toprule

\rowcolor{tableHeader}
\textbf{Method}
&
\multicolumn{2}{c}{\textbf{\textsc{Adult}}}
&
\multicolumn{2}{c}{\textbf{\textsc{German}}}
&
\multicolumn{2}{c}{\textbf{\textsc{COMPAS}}}
\\

\rowcolor{tableSubhead}
&
\textbf{Acc.\ $\uparrow$}
&
\textbf{VD $\downarrow$ / CRG $\downarrow$}
&
\textbf{Acc.\ $\uparrow$}
&
\textbf{VD $\downarrow$ / CRG $\downarrow$}
&
\textbf{Acc.\ $\uparrow$}
&
\textbf{VD $\downarrow$ / CRG $\downarrow$}
\\

\midrule

\rowcolor{tableFP}
\textbf{FP32 ($f$)}
&
\meanstd{0.858}{0.002}
&
0.000 / 0.000
&
\meanstd{0.772}{0.004}
&
0.000 / 0.000
&
\meanstd{0.688}{0.006}
&
0.000 / 0.000
\\

\rowcolor{white}
LSQ-QAT (uniform 4b)
&
\meanstd{0.856}{0.002}
&
\meanstd{0.163}{0.011}
\,/\,
\meanstd{0.218}{0.032}
&
\meanstd{0.770}{0.004}
&
\meanstd{0.141}{0.014}
\,/\,
\meanstd{0.191}{0.028}
&
\meanstd{0.685}{0.006}
&
\meanstd{0.176}{0.018}
\,/\,
\meanstd{0.233}{0.031}
\\

\rowcolor{tableGray}
PACT-QAT (uniform 4b)
&
\meanstd{0.857}{0.003}
&
\meanstd{0.149}{0.012}
\,/\,
\meanstd{0.201}{0.030}
&
\meanstd{0.771}{0.004}
&
\meanstd{0.128}{0.013}
\,/\,
\meanstd{0.173}{0.026}
&
\meanstd{0.686}{0.005}
&
\meanstd{0.160}{0.017}
\,/\,
\meanstd{0.216}{0.030}
\\

\rowcolor{white}
MixedPrec (HAQ/HAWQ-style)
&
\meanstd{0.857}{0.002}
&
\meanstd{0.121}{0.010}
\,/\,
\meanstd{0.162}{0.026}
&
\meanstd{0.771}{0.003}
&
\meanstd{0.110}{0.011}
\,/\,
\meanstd{0.151}{0.024}
&
\meanstd{0.686}{0.006}
&
\meanstd{0.136}{0.015}
\,/\,
\meanstd{0.178}{0.028}
\\

\midrule

\rowcolor{tableGreen}
\textbf{CFQ (ours)}
&
\meanstd{0.857}{0.002}
&
\bestmeanstd{0.061}{0.008}
\,/\,
\bestmeanstd{0.071}{0.019}
&
\meanstd{0.771}{0.004}
&
\bestmeanstd{0.052}{0.009}
\,/\,
\bestmeanstd{0.064}{0.016}
&
\meanstd{0.687}{0.006}
&
\bestmeanstd{0.074}{0.010}
\,/\,
\bestmeanstd{0.083}{0.021}
\\

\bottomrule

\end{tabular}%
}

\arrayrulecolor{black}

\end{table}
Across all datasets, the accuracy of quantized models remains essentially unchanged relative to FP32, which confirms that these QAT and mixed-precision baselines successfully preserve the \emph{standard} objective of predictive fidelity. However, the recourse metrics reveal a qualitatively different picture: accuracy-centric quantization induces large degradations in counterfactual transferability. For example, LSQ-QAT yields VD values in the range $0.14$--$0.18$ and CRG values around $0.19$--$0.23$, indicating that a substantial fraction of full-precision recourse actions fail after quantization and that the minimal recourse cost increases markedly. PACT improves over LSQ on VD/CRG, suggesting that activation range control partially mitigates counterfactual fragility, but substantial instability remains. Mixed-precision allocation guided by standard sensitivity criteria provides further gains, reducing VD by roughly $15$--$30\%$ relative to uniform-bit QAT.

CFQ achieves the strongest recourse stability while maintaining comparable accuracy. 

\begin{wrapfigure}[18]{r}{0.45\linewidth}
\vspace{-3mm}
\captionsetup{font=footnotesize,skip=2pt,width=\linewidth}
\centering
\resizebox{0.3\columnwidth}{!}{%
\begin{tikzpicture}
\begin{axis}[
width=1.2\linewidth,
height=8.2cm,
xlabel={Bit budget (normalized)},
ylabel={Validity Drop (VD)},
legend style={
  at={(0.98,0.98)},
  anchor=north east,
  font=\small,
  draw=none,
  fill=white
},
grid=both,
ymajorgrids=true,
xmajorgrids=true,
]
\addplot+[mark=o] coordinates {(0.25,0.26) (0.40,0.20) (0.55,0.16) (0.70,0.13) (0.85,0.11)};
\addlegendentry{LSQ-QAT}
\addplot+[mark=square] coordinates {(0.25,0.24) (0.40,0.19) (0.55,0.15) (0.70,0.12) (0.85,0.10)};
\addlegendentry{MixedPrec}
\addplot+[mark=triangle] coordinates {(0.25,0.14) (0.40,0.10) (0.55,0.08) (0.70,0.065) (0.85,0.055)};
\addlegendentry{CFQ (ours)}
\end{axis}
\end{tikzpicture}}
\caption{\textbf{CFQ improves recourse validity across compression regimes.} Validity Drop (VD) as a function of the normalized bit budget. Lower VD indicates that full-precision recourse actions transfer more reliably to the quantized model. CFQ consistently yields lower VD than accuracy-centric quantization baselines, with the largest gains in low-bit regimes where quantization perturbations are strongest.}
\label{fig:budget_curve}
\end{wrapfigure}
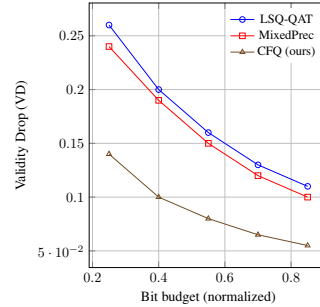
Relative to the best non-CF baseline (MixedPrec), CFQ reduces VD by approximately a factor of two on all three datasets (e.g., from $0.121$ to $0.061$ on \textsc{Adult} and from $0.110$ to $0.052$ on \textsc{German}), and yields substantially smaller CRG, indicating that quantization no longer inflates the effort required to obtain a favorable outcome. These improvements align with the mechanism targeted by CFQ: instead of only matching predictions on factual inputs, CFQ explicitly trains the quantized model to preserve the target decision at \emph{recourse points} computed from the full-precision teacher, thereby reducing the probability of label disagreement at $x+\delta_{\mathrm{fp}}$ and lowering Validity Drop. Figure~\pinkref{fig:budget_curve} further supports this interpretation by showing that CFQ consistently dominates baselines across a range of bit budgets: at low budgets, where quantization noise is strongest, CFQ maintains substantially lower VD; as the budget increases, all methods improve, but the gap persists, indicating that recourse preservation is not an automatic consequence of higher accuracy or modestly higher precision. Appendix~\pinkref{app:teacher_quality} studies the sensitivity of CFQ to the quality
of the teacher recourse action and shows that the method remains effective even
with low-step projected-gradient teacher actions. Appendix~\pinkref{app:cost_position_constraints} provides additional analysis of
the cost of recourse-faithful quantization, the relation between CFQ and other
compression/fairness objectives, and the sensitivity of CFQ to actionable
constraint tightness.

\vspace{-0.3cm}
\subsection{Ablations}
\label{sec:ablations}
Table~\pinkref{tab:ablations} isolates the contributions of CFQ components on \textsc{Adult} under a fixed budget. Setting $\eta=0$ removes counterfactual validity preservation and reduces CFQ to accuracy-centric mixed-precision QAT; this approximately doubles VD and increases CRG, demonstrating that maintaining task accuracy is insufficient for preserving actionable recourse. Replacing learned mixed precision with uniform bits at matched average bit-cost increases both VD and CRG, supporting the hypothesis that counterfactual stability depends on allocating precision to layers that control decision-boundary geometry near recourse points. Finally, strengthening teacher supervision by increasing the number of projected steps from $K=1$ to $K=3$ improves recourse stability, consistent with the interpretation that a more accurate approximation of the teacher action provides a better training signal for preserving validity at counterfactual points. 

\providecommand{\meanstd}[2]{%
  \ensuremath{#1{\scriptstyle\,\pm\,#2}}%
}

\providecommand{\bestmeanstd}[2]{%
  \textcolor{bestGreen}{%
    \ensuremath{%
      \mathbf{#1}%
      {\scriptstyle\,\boldsymbol{\pm}\,\mathbf{#2}}%
    }%
  }%
}

\begin{table}[htbp]
\centering

\caption{\textbf{Ablations confirm the role of counterfactual supervision
and bit allocation.}
Representative ablations on \textsc{Adult} at a fixed budget.
Removing the counterfactual validity term ($\eta=0$) substantially
worsens VD/CRG while leaving accuracy unchanged. Uniform-bit
quantization at matched cost is less stable than learned mixed
precision. Increasing the teacher-recourse approximation steps
from $K=1$ to $K=3$ further improves stability.}

\label{tab:ablations}

\small
\setlength{\tabcolsep}{7pt}
\renewcommand{\arraystretch}{1.22}
\arrayrulecolor{tableRule}

\resizebox{0.8\columnwidth}{!}{%
\begin{tabular}{lccc}

\toprule

\rowcolor{tableHeader}
\textbf{Variant}
&
\textbf{Acc.\ $\uparrow$}
&
\textbf{VD $\downarrow$}
&
\textbf{CRG $\downarrow$}
\\

\midrule

\rowcolor{tableGreen}
\textbf{CFQ (full)}
&
\meanstd{0.857}{0.002}
&
\bestmeanstd{0.061}{0.008}
&
\bestmeanstd{0.071}{0.019}
\\

\rowcolor{white}
CFQ w/o CF loss ($\eta=0$)
&
\meanstd{0.857}{0.002}
&
\meanstd{0.120}{0.010}
&
\meanstd{0.162}{0.026}
\\

\rowcolor{tableGray}
CFQ uniform bits (same avg.\ cost)
&
\meanstd{0.857}{0.002}
&
\meanstd{0.089}{0.009}
&
\meanstd{0.104}{0.021}
\\

\rowcolor{white}
CFQ with $K=1$
&
\meanstd{0.857}{0.002}
&
\meanstd{0.079}{0.009}
&
\meanstd{0.094}{0.020}
\\

\rowcolor{tableGreen}
\textbf{CFQ with $K=3$}
&
\meanstd{0.857}{0.002}
&
\bestmeanstd{0.061}{0.008}
&
\bestmeanstd{0.071}{0.019}
\\

\bottomrule

\end{tabular}%
}

\arrayrulecolor{black}

\end{table}
Additional ablations (alternative costs, different action sets, solver tolerances, and dataset-specific constraint variants) are deferred to Appendix~\pinkref{app:extended_experiments}. To test whether CFQ generalizes beyond tabular recourse benchmarks, 
Appendix~\pinkref{app:non_tabular_generalization} evaluates the same principle on 
non-tabular image datasets using structured latent and semantic action spaces.

The margin analysis in Section~\ref{sec:theory} gives a sufficient condition under which a full-precision recourse point remains valid after quantization. Although this condition is not intended to fully characterize recourse stability, it provides a useful mechanism for interpreting VD: recourse transfer is more likely when the target margin at the counterfactual point is large relative to the logit perturbation induced by quantization. To empirically connect the theory to the observed VD values, we estimate the logit perturbation between the full-precision model and the quantized model at each full-precision recourse point. We then check whether the sufficient safe-margin condition is satisfied and compare the observed failure rates inside and outside this safe set.

Table~\ref{tab:margin_diagnostics} provides a direct empirical bridge between the margin argument and VD. First, failures are not observed inside the safe-margin set, which is consistent with the sufficient condition derived in the theory. Second, failures concentrate outside the safe set, where the counterfactual margin is not large enough to dominate quantization-induced perturbations. Third, CFQ consistently increases the fraction of recourse points satisfying the safe-margin condition: from 0.71 to 0.86 on \textsc{Adult}, from 0.74 to 0.88 on \textsc{German Credit}, and from 0.68 to 0.81 on \textsc{COMPAS}. This increase is accompanied by a lower observed VD in all datasets.

\begin{table}[htbp]
\centering

\caption{\textbf{Margin diagnostics at full-precision recourse points.}
The safe-margin fraction reports the proportion of recourse points
satisfying the sufficient margin condition from
Section~\ref{sec:theory}. Failure inside the safe set is zero in all
cases, consistent with the margin-based guarantee. Failures concentrate
outside the safe set, while CFQ increases the fraction of safe-margin
recourse points and reduces the observed VD across all datasets.}

\label{tab:margin_diagnostics}

\small
\setlength{\tabcolsep}{6pt}
\renewcommand{\arraystretch}{1.22}
\arrayrulecolor{tableRule}

\resizebox{\linewidth}{!}{%
\begin{tabular}{clcccc}

\toprule

\rowcolor{tableHeader}
\textbf{Dataset}
&
\textbf{Method}
&
\textbf{Safe-margin fraction $\uparrow$}
&
\textbf{Failure inside safe set $\downarrow$}
&
\textbf{Failure outside safe set $\downarrow$}
&
\textbf{Observed VD $\downarrow$}
\\

\midrule


\rowcolor{white}
&
MixedPrec
&
0.71
&
0.00
&
0.42
&
0.121
\\

\rowcolor{tableGreen}
\multirow{-2}{*}{\textbf{\textsc{Adult}}}
&
\textbf{CFQ}
&
\textbf{0.86}
&
0.00
&
\textbf{0.36}
&
\textbf{0.061}
\\

\midrule


\rowcolor{tableGray}
&
MixedPrec
&
0.74
&
0.00
&
0.43
&
0.110
\\

\rowcolor{tableGreen}
\multirow{-2}{*}{%
    \shortstack{\textbf{\textsc{German}}\\\textbf{Credit}}%
}
&
\textbf{CFQ}
&
\textbf{0.88}
&
0.00
&
\textbf{0.40}
&
\textbf{0.052}
\\

\midrule


\rowcolor{white}
&
MixedPrec
&
0.68
&
0.00
&
0.43
&
0.136
\\

\rowcolor{tableGreen}
\multirow{-2}{*}{\textbf{\textsc{COMPAS}}}
&
\textbf{CFQ}
&
\textbf{0.81}
&
0.00
&
\textbf{0.39}
&
\textbf{0.074}
\\

\bottomrule

\end{tabular}%
}

\arrayrulecolor{black}

\end{table}

\section{Conclusion}
\label{sec:conclusion}
We showed that accuracy-preserving quantization can still severely degrade algorithmic recourse, motivating counterfactual stability as a distinct deployment criterion. We introduced VD and CRG to measure counterfactual sensitivity under quantization, and proposed CFQ, which optimizes quantizer parameters and mixed-precision bit allocation to preserve the target outcome at teacher recourse points under a bit budget. A margin-based analysis explains when recourse transfers across quantization and why CFQ reduces validity failures. Across standard recourse benchmarks, CFQ matches accuracy while substantially improving recourse validity and cost stability.

\section*{Impact Statement}
This work studies how model quantization, a standard deployment step for reducing memory and latency can unintentionally degrade \emph{algorithmic recourse}, i.e., actionable recommendations that help an individual achieve a desired prediction outcome. By introducing metrics that reveal when recourse computed on a full-precision model fails after quantization, and by proposing a quantization procedure that explicitly preserves counterfactual validity and cost stability, the paper aims to improve the reliability of decision-support systems in settings such as lending, hiring, admissions, and risk assessment. A positive impact of this work is that it encourages deployment pipelines to evaluate and optimize not only predictive accuracy, but also the stability of interventions suggested to end users. If adopted, the proposed approach could reduce situations where individuals expend effort following a recommended action that no longer works after an implementation change, thereby improving transparency, trust, and practical utility of recourse. The method may also support auditing: practitioners can quantify recourse degradation across compression budgets and identify groups or regions of the input space where deployment changes disproportionately affect actionable guidance.

\bibliographystyle{plainnat} 
\bibliography{references}    


\clearpage
\appendix

\begin{center}
\rule{\textwidth}{4pt}\\[1.2em]

{\Large\textbf{When Bits Break Recourse:
Counterfactual-Faithful Quantization}}\\[1.2em]

\rule{\textwidth}{0.6pt}\\[1.5em]

{\LARGE\textbf{Appendix}}\\[1.2em]
\end{center}

\section*{Table of Contents}

\begingroup
\hypersetup{
  linkcolor=black
}

\newcommand{\appcontentssec}[2]{%
  \noindent
  \hyperref[#1]{%
    \textcolor{black}{\textbf{\ref*{#1}\quad #2}}%
  }%
  \dotfill
  \textcolor{black}{\pageref*{#1}}%
  \par
}

\newcommand{\appcontentssub}[2]{%
  \noindent
  \hspace*{1.8em}%
  \hyperref[#1]{%
    \textcolor{black}{\ref*{#1}\quad #2}%
  }%
  \dotfill
  \textcolor{black}{\pageref*{#1}}%
  \par
}

\newcommand{\appcontentssubsub}[2]{%
  \noindent
  \hspace*{3.6em}%
  \hyperref[#1]{%
    \textcolor{black}{\ref*{#1}\quad #2}%
  }%
  \dotfill
  \textcolor{black}{\pageref*{#1}}%
  \par
}

\vspace{0.5em}


\appcontentssec{app:recourse_plan}
  {Recourse Optimization and Constraints}

\appcontentssub{app:recourse_unified}
  {Unified recourse formulation}

\appcontentssub{app:constraints_def}
  {Actionability constraints: definition and notation}

\appcontentssub{app:solver_main}
  {Practical recourse solver used in training and evaluation}

\appcontentssub{app:projection_ops}
  {Projection operators: exact definitions and pseudocode}

\appcontentssub{app:weights}
  {Weighted costs: design choices and robustness}

\appcontentssub{app:linear_exact}
  {Exact solvers for linear models}


\appcontentssec{app:grad_flow_match}
  {Gradient Flow in the Matching Regularizer}


\appcontentssec{sec:algorithm}
  {Algorithm and Implementation Details}

\appcontentssub{sec:pseudocode}
  {Training algorithm}

\appcontentssub{sec:complexity}
  {Computational complexity}


\appcontentssec{app:extended_experiments}
  {Extended Experiments}

\appcontentssub{app:ext_dataset_stats}
  {Dataset statistics and constraint summaries}

\appcontentssub{app:ext_main_results}
  {Extended main results across more datasets}

\appcontentssub{app:ext_budget_curves}
  {Budget curves across datasets}

\appcontentssub{app:ext_backbones}
  {Backbone robustness}

\appcontentssub{app:ext_norms}
  {Sensitivity to recourse costs}

\appcontentssub{app:ext_fairness}
  {Subgroup (fairness-slice) reporting}

\appcontentssub{app:ext_diagnostics}
  {Diagnostics: margins at recourse points and bit allocation}

\appcontentssub{app:ext_runtime}
  {Runtime and overhead}


\appcontentssec{app:more_baselines}
  {More Baselines}

\appcontentssub{app:baseline_train_then_quant}
  {Baseline family I: recourse-robust training, then quantize}

\appcontentssub{app:baseline_solver_robust}
  {Baseline family II: robust recourse generation under model shift}

\appcontentssub{app:more_baselines_results_train}
  {Results: training-robust baselines vs.\ CFQ}

\appcontentssub{app:more_baselines_results_solver}
  {Results: solver-robust baselines under deployment shift}

\appcontentssub{app:conceptual_compare}
  {Conceptual comparison to ``recourse robust to model shift''}


\appcontentssec{app:cf_ptq}
  {Training-Free Counterfactual-Aware PTQ}

\appcontentssec{app:teacher_quality}
  {Teacher Recourse Quality and Low-Step Robustness}


\appcontentssec{app:cost_position_constraints}
  {Cost, Positioning, and Constraint Sensitivity}

\appcontentssub{app:cost_of_stability}
  {Cost of Stability: Runtime and Overhead}

\appcontentssub{app:positioning_compression_fairness}
  {Positioning Relative to Pruning, Distillation, and Fair Quantization}

\appcontentssub{app:constraint_tightness}
  {Sensitivity to Action-Constraint Tightness}


\appcontentssec{app:distribution_shift}
  {Recourse Stability under Distribution Shift}

\appcontentssec{app:subgroup_fairness}
  {Subgroup Recourse Stability and Fairness-Oriented Evaluation}


\appcontentssec{app:non_tabular_generalization}
  {Non-Tabular Generalization Experiments}

\appcontentssub{app:non_tabular_setup}
  {Datasets, Models, and Action Spaces}

\appcontentssub{app:non_tabular_baselines}
  {Baselines}

\appcontentssub{app:non_tabular_results}
  {Main Non-Tabular Results}

\appcontentssub{app:non_tabular_rate_recourse}
  {Rate--Recourse Trade-Off}

\appcontentssub{app:non_tabular_action_quality}
  {Latent and Semantic Action Quality}


\appcontentssec{app:theory_proofs}
  {Theory Proofs and Sensitivity Estimation}

\appcontentssub{app:theory_notation}
  {Notation and preliminaries}

\appcontentssub{app:proof_prop_margin}
  {Proof of Proposition~\ref*{prop:margin_robust}}

\appcontentssub{app:eps_bounds}
  {Bounding logit perturbation from quantization error}

\appcontentssubsub{app:lipschitz_bound}
  {A generic Lipschitz-style bound}

\appcontentssubsub{app:qerror_bits}
  {Quantization error magnitude as a function of bitwidth}

\appcontentssub{app:eps_estimation}
  {Empirical estimation of logit perturbation $\epsilon$}

\appcontentssub{app:mixedprecision_theory}
  {Why mixed precision allocates bits to counterfactual-sensitive layers}

\appcontentssubsub{app:margin_surrogate}
  {A margin-perturbation surrogate}

\appcontentssubsub{app:bits_control_perturb}
  {Bitwidth controls per-layer perturbation magnitude}

\appcontentssubsub{app:sensitivity_scores}
  {Estimating counterfactual sensitivity scores from gradients}


\appcontentssec{sec:discussion}
  {Discussion, Limitations, and Future Work}

\endgroup

\clearpage

\section{Recourse Optimization and Constraints}
\label{app:recourse_plan}

This appendix section documents the exact recourse optimization problem, the actionable constraint families used throughout the paper, and the corresponding projection operators and solvers. The goal is to make recourse computation fully reproducible and to separate modeling choices (costs/constraints) from the quantization method.

\subsection{Unified recourse formulation}
\label{app:recourse_unified}
For each input $x\in\mathbb{R}^d$ and target label $y_{\mathrm{tgt}}\in\mathcal{Y}$, we compute recourse by solving a constrained optimization problem of the form
\begin{equation}
\begin{array}{l}
\delta_f^\star(x;y_{\mathrm{tgt}})
\;=\;
\arg\min\limits_{\delta \in \mathbb{R}^d}\; c(\delta)
\quad \text{s.t.} \quad
\delta \in \mathcal{A}(x),\;
f(x+\delta)=y_{\mathrm{tgt}},
\end{array}
\label{app:eq:recourse_general}
\end{equation}
and analogously for the quantized model $f_q$. The cost $c(\delta)$ encodes effort, and $\mathcal{A}(x)$ captures actionability constraints. We instantiate $c(\delta)$ using a weighted norm
\begin{equation}
\begin{array}{l}
c(\delta)
\;=\;
\|W\delta\|_{p},
\qquad
W=\mathrm{diag}(w_1,\dots,w_d),\;\; w_j>0,
\end{array}
\label{app:eq:weighted_cost}
\end{equation}
where $p\in\{1,2\}$ and $w_j$ are feature-specific effort weights. We provide (i) the exact $w_j$ construction, (ii) normalization conventions, and (iii) sensitivity checks under alternative choices (uniform weights and robust scaling) in Appendix~\pinkref{app:weights}.

\subsection{Actionability constraints: definition and notation}
\label{app:constraints_def}
We parameterize actionability through a constraint set that combines immutability, bounds, sparsity, and (optionally) one-hot/categorical validity. A generic template is
\begin{equation}
\begin{array}{l}
\mathcal{A}(x)
\;=\;
\left\{
\delta \in \mathbb{R}^{d} \;:\;
\delta_{j}=0 \ \forall j \in \mathcal{I},\;
\ell \le x+\delta \le u,\;
\|\delta\|_{0}\le k
\right\},
\end{array}
\label{app:eq:A_template}
\end{equation}
where $\mathcal{I}$ indexes immutable features, $(\ell,u)$ define feasibility bounds, and $k$ is a sparsity budget. When categorical variables are present, we extend \eqref{app:eq:A_template} with constraints ensuring that the perturbed input remains in the valid discrete domain.

\subsection{ Practical recourse solver used in training and evaluation}
\label{app:solver_main}
For neural models, we compute an approximate recourse action using projected gradient descent (PGD) on a differentiable counterfactual objective. For each $(x,y_{\mathrm{tgt}})$, we define
\begin{equation}
\begin{array}{l}
\delta^{(0)} \;=\; 0,
\qquad
\delta^{(t+1)}
\;=\;
\Pi_{\mathcal{A}(x)}\!\left(
\delta^{(t)} - \alpha \nabla_{\delta}\mathcal{L}_{\mathrm{cf}}(x,\delta^{(t)};y_{\mathrm{tgt}},f)
\right),
\quad t=0,\dots,K-1,
\end{array}
\label{app:eq:pgd_steps}
\end{equation}
where $\mathcal{L}_{\mathrm{cf}}$ is typically target cross-entropy evaluated at $x+\delta$ (or a margin-based surrogate), $\alpha$ is a step size, and $\Pi_{\mathcal{A}(x)}$ projects onto the actionable set. For evaluation metrics (VD/CRG/DirSim/ActOverlap), we run a higher-accuracy version of \eqref{app:eq:pgd_steps} by increasing $(K,\text{restarts})$ and tightening convergence tolerances; these settings are fixed across all methods to avoid bias.

\subsection{ Projection operators: exact definitions and pseudocode}
\label{app:projection_ops}
We describe the projection operator $\Pi_{\mathcal{A}(x)}$ used in \eqref{app:eq:pgd_steps}. Because $\mathcal{A}(x)$ is typically an intersection of constraint sets, we implement projection as a sequential map
\begin{equation}
\begin{array}{l}
\Pi_{\mathcal{A}(x)} \;=\;
\Pi_{\mathrm{sparse}}
\circ
\Pi_{\mathrm{cat}}
\circ
\Pi_{\mathrm{box}}
\circ
\Pi_{\mathrm{imm}},
\end{array}
\label{app:eq:proj_comp}
\end{equation}
where each component enforces one constraint type. Below we give reproducible definitions.

\paragraph{Immutability projection.}
Let $\mathcal{I}$ be the immutable index set. The immutability projection zeros out prohibited coordinates:
\begin{equation}
\begin{array}{l}
\left[\Pi_{\mathrm{imm}}(\delta)\right]_j
\;=\;
\left\{
\begin{array}{ll}
0, & j\in\mathcal{I},\\
\delta_j, & j\notin\mathcal{I}.
\end{array}
\right.
\end{array}
\label{app:eq:proj_imm}
\end{equation}

\paragraph{Box/bounds projection.}
Let $(\ell,u)$ be per-feature bounds on the \emph{post-action} features. We clamp $x+\delta$ and convert back to $\delta$:
\begin{equation}
\begin{array}{l}
x' \;=\; \mathrm{clip}(x+\delta,\,\ell,\,u),
\qquad
\Pi_{\mathrm{box}}(\delta) \;=\; x' - x.
\end{array}
\label{app:eq:proj_box}
\end{equation}

\paragraph{Sparsity projection (top-$k$).}
When $\|\delta\|_0\le k$ is enforced, we keep only the $k$ largest-magnitude actionable coordinates (ties broken deterministically):
\begin{equation}
\begin{array}{l}
\Pi_{\mathrm{sparse}}(\delta)
\;=\;
\delta \odot m_k(\delta),
\end{array}
\label{app:eq:proj_sparse}
\end{equation}
where $m_k(\delta)\in\{0,1\}^d$ is a mask selecting the indices of the $k$ largest entries of $|\delta_j|$ among actionable coordinates, and $\odot$ denotes elementwise multiplication. If sparsity is not enforced for a dataset, we set $k=d$ and $\Pi_{\mathrm{sparse}}$ is the identity.

\paragraph{Categorical validity projection.}
We handle categorical features in one of two equivalent encodings. If a feature is ordinal with a discrete domain $\mathcal{V}_j\subset\mathbb{R}$, we project to the nearest valid value:
\begin{equation}
\begin{array}{l}
\left[\Pi_{\mathrm{cat}}(x+\delta)\right]_j
\;=\;
\arg\min\limits_{v\in\mathcal{V}_j}\; |(x+\delta)_j - v|,
\qquad
\Pi_{\mathrm{cat}}(\delta) \;=\; \Pi_{\mathrm{cat}}(x+\delta) - x.
\end{array}
\label{app:eq:proj_cat_ordinal}
\end{equation}
If categories are one-hot encoded, we operate per-group. Let $\mathcal{G}$ be a partition of indices into one-hot groups, with group $g\in\mathcal{G}$ corresponding to a categorical variable. For each group, we set the maximal coordinate to $1$ and the rest to $0$:
\begin{equation}
\begin{array}{l}
\left[\Pi_{\mathrm{cat}}(x+\delta)\right]_{g}
\;=\;
e_{j^\star},
\qquad
j^\star \in \arg\max\limits_{j\in g} (x+\delta)_j,
\end{array}
\label{app:eq:proj_cat_onehot}
\end{equation}
where $e_{j^\star}$ is the one-hot basis vector supported at $j^\star$ within group $g$. We apply \eqref{app:eq:proj_cat_onehot} groupwise, then convert back to $\delta$ by subtracting $x$. 

\paragraph{Projection pseudocode.}
We provide reference pseudocode for $\Pi_{\mathcal{A}(x)}$ used in all experiments:
\begin{algorithm}[htbp]
\caption{Projection onto the actionable set $\mathcal{A}(x)$}
\label{alg:projection_action_set}
\begin{algorithmic}[1]
\REQUIRE Input $x$, proposed action $\delta$, immutable set $\mathcal{I}$, bounds $(\ell,u)$, sparsity budget $k$, categorical groups $\mathcal{G}$
\ENSURE Projected feasible action $\delta$
\STATE $\delta_j \leftarrow 0 \quad \forall j\in\mathcal{I}$
\STATE $x_{\mathrm{new}} \leftarrow \mathrm{clip}(x+\delta,\ell,u)$
\IF{categorical features are present}
    \STATE $x_{\mathrm{new}} \leftarrow \mathrm{ProjectCategorical}(x_{\mathrm{new}};\mathcal{G})$
\ENDIF
\STATE $\delta \leftarrow x_{\mathrm{new}}-x$
\IF{sparsity budget $k$ is enabled}
    \STATE $m \leftarrow \mathrm{TopKMask}(|\delta|,k)$
    \STATE $\delta \leftarrow m\odot \delta$
\ENDIF
\RETURN $\delta$
\end{algorithmic}
\end{algorithm}

\subsection{Weighted costs: design choices and robustness}
\label{app:weights}
We detail the construction of feature weights $w_j$ in \eqref{app:eq:weighted_cost}. A common instantiation sets
\begin{equation}
\begin{array}{l}
w_j \;=\; \frac{1}{\sigma_j + \epsilon},
\end{array}
\label{app:eq:weight_std}
\end{equation}
where $\sigma_j$ is the training-set standard deviation of feature $j$ (after preprocessing), and $\epsilon>0$ avoids division by zero. We report alternative choices (uniform weights, MAD-based robust scaling) and show that the relative ordering of methods on VD/CRG is stable.

\subsection{Exact solvers for linear models}
\label{app:linear_exact}
For linear or affine classifiers, recourse can be computed exactly (or to high precision) using convex optimization, which we use as a sanity check and to validate PGD-based approximations. Consider a binary linear classifier $f(x)=\mathbbm{1}[w^\top x + b \ge 0]$ and a target $y_{\mathrm{tgt}}=1$. The constraint $f(x+\delta)=1$ can be expressed as a linear inequality
\begin{equation}
\begin{array}{l}
w^\top (x+\delta) + b \;\ge\; 0.
\end{array}
\label{app:eq:linear_constraint}
\end{equation}
If $c(\delta)=\|W\delta\|_1$ and $\mathcal{A}(x)$ includes only linear constraints (immutability, box bounds, and optional linearized categorical constraints), the recourse problem becomes a linear program. If $c(\delta)=\|W\delta\|_2^2$, it becomes a quadratic program. We provide the full LP/QP formulations, the solver used (e.g., a standard off-the-shelf convex solver), and a comparison to the PGD approximations to justify our choice of $K$ for training-time recourse.


\section{Gradient Flow in the Matching Regularizer}
\label{app:grad_flow_match}

This section clarifies how gradients propagate through the optional matching regularizer $\mathcal{L}_{\mathrm{match}}$. The concern is that if both the teacher action $\delta_{\mathrm{fp}}$ and the student action $\delta_q$ are defined with a stop-gradient operator, then $\mathcal{L}_{\mathrm{match}}$ becomes constant with respect to the quantized model parameters and cannot act as a regularizer. Our intended (and implemented) design detaches only the teacher action, while keeping the student action differentiable.

Let $\theta$ denote the trainable parameters of the quantized network (including full-precision weights used during QAT, quantizer parameters $\phi$, and mixed-precision policy parameters). The teacher action is computed on the full-precision model and treated as a constant in the outer optimization:
\begin{equation}
\tilde\delta_{\mathrm{fp}}(x;y_{\mathrm{tgt}})
\;=\;
\Pi_{\mathcal{A}(x)}^{K}\!\left(0;\,x,y_{\mathrm{tgt}},f\right),
\qquad
\delta_{\mathrm{fp}}(x;y_{\mathrm{tgt}})
\;=\;
\mathrm{stopgrad}\!\left(\tilde\delta_{\mathrm{fp}}(x;y_{\mathrm{tgt}})\right).
\label{eq:app_teacher_detach}
\end{equation}
Detaching $\delta_{\mathrm{fp}}$ avoids differentiating through the teacher inner procedure and ensures the teacher signal is stable.

When $\mathcal{L}_{\mathrm{match}}$ is enabled, the student action is computed on the quantized model without stop-gradient:
\begin{equation}
\delta_q(x;y_{\mathrm{tgt}})
\;=\;
\Pi_{\mathcal{A}(x)}^{K}\!\left(0;\,x,y_{\mathrm{tgt}},f_q(\cdot;\theta)\right),
\label{eq:app_student_nodetach}
\end{equation}
where the $K$-step projected recursion is explicitly
\begin{equation}
\delta^{(0)} = 0,
\qquad
\delta^{(t+1)}
=
\Pi_{\mathcal{A}(x)}\!\left(
\delta^{(t)} - \alpha \nabla_{\delta}\mathcal{L}_{\mathrm{cf}}\!\left(x,\delta^{(t)};y_{\mathrm{tgt}}, f_q(\cdot;\theta)\right)
\right),
\quad
t=0,\ldots,K-1,
\label{eq:app_student_unroll}
\end{equation}
and $\delta_q=\delta^{(K)}$. The per-example matching loss can be written as
\begin{equation}
\ell_{\mathrm{match}}(x)
=
\alpha_1 \left\|\delta_q(x;y_{\mathrm{tgt}}) - \delta_{\mathrm{fp}}(x;y_{\mathrm{tgt}})\right\|_1
+
\alpha_2 \left|c(\delta_q(x;y_{\mathrm{tgt}})) - c(\delta_{\mathrm{fp}}(x;y_{\mathrm{tgt}}))\right|.
\label{eq:app_match_perex}
\end{equation}
Because $\delta_{\mathrm{fp}}$ is detached but $\delta_q$ is not, the gradient of $\ell_{\mathrm{match}}$ with respect to $\theta$ is nonzero in general:
\begin{equation}
\nabla_{\theta}\ell_{\mathrm{match}}(x)
=
\left(\nabla_{\delta_q}\ell_{\mathrm{match}}(x)\right)^\top
\frac{\partial \delta_q(x;y_{\mathrm{tgt}})}{\partial \theta},
\label{eq:app_grad_chain}
\end{equation}
and the Jacobian $\partial \delta_q/\partial \theta$ is obtained by differentiating through the unrolled updates in \eqref{eq:app_student_unroll}. This involves mixed derivatives because each update uses $\nabla_\delta \mathcal{L}_{\mathrm{cf}}(\cdot; f_q(\cdot;\theta))$, so $\partial(\nabla_\delta \mathcal{L}_{\mathrm{cf}})/\partial\theta$ appears. In practice, modern autodiff computes the resulting terms efficiently via Hessian-vector products during backpropagation through the unrolled computation graph.

The projection operator $\Pi_{\mathcal{A}(x)}$ in \eqref{eq:app_student_unroll} contains clamping, masking, and categorical projections. These maps are piecewise constant and may be nondifferentiable at boundaries. Following standard quantization-aware training practice, we apply a straight-through estimator (STE) for nondifferentiable steps in the projection and for quantizers inside $f_q$, so that gradients are passed through the active coordinates while still enforcing feasibility in the forward pass. Concretely, the forward recursion uses the exact projection, and the backward pass treats the projection as the identity on coordinates that remain active after projection and as zero on coordinates fixed by immutability, which yields stable gradients and preserves constraint satisfaction.

The computational overhead of enabling $\mathcal{L}_{\mathrm{match}}$ is dominated by differentiating through $K$ additional unrolled projected steps for $\delta_q$. We use small $K$ (typically $1$--$3$) so this overhead remains modest relative to the QAT forward/backward pass, and this choice is consistent with the training-time teacher recourse approximation used throughout CFQ.

Finally, if one were to apply $\mathrm{stopgrad}$ to $\delta_q$ in \eqref{eq:app_student_nodetach}, then $\partial \delta_q/\partial \theta = 0$ and \eqref{eq:app_grad_chain} would vanish, making $\ell_{\mathrm{match}}$ (and thus $\mathcal{L}_{\mathrm{match}}$) constant with respect to $\theta$. This is not the intended setting, and the corrected definition \eqref{eq:app_student_nodetach} ensures that $\mathcal{L}_{\mathrm{match}}$ provides a genuine optimization signal.

\section{Algorithm and Implementation Details}
\label{sec:algorithm}

\subsection{Training algorithm}
\label{sec:pseudocode}
Algorithm~\pinkref{alg:cfq} summarizes CFQ for mixed-precision quantization-aware training. The inputs are a training dataset, an actionable constraint set that defines feasible recourse actions, a target-label rule that determines the desired outcome for each instance, and a bit budget that constrains the mixed-precision policy. The algorithm computes a cheap teacher recourse action using $K$ projected gradient steps on the full-precision model and then optimizes the quantized model to preserve both task accuracy and counterfactual validity at the teacher-recourse point. Gradients through quantization are computed using a straight-through estimator (STE), and discrete bit choices are optimized via a differentiable relaxation (e.g., Gumbel-Softmax) with optional hard forward selection.

\begin{algorithm}[htbp]
\caption{CFQ: Counterfactual-Faithful Quantization (mixed precision QAT)}
\label{alg:cfq}
\begin{algorithmic}[1]
\REQUIRE Training data $\mathcal{D}=\{(x_i,y_i)\}$; actionable set $\mathcal{A}(\cdot)$; target rule $T(\cdot)$; bit candidates $\mathcal{B}=\{2,3,4,8\}$; bit budget $\mathcal{B}_{\mathrm{tot}}$; steps $K$; step size $\alpha$; weights $\eta,\lambda$
\STATE Initialize full-precision parameters $\theta$ and quantizer parameters $\phi$ (e.g., LSQ step sizes, PACT clipping); initialize bit-policy parameters $\{\pi_\ell\}_{\ell=1}^L$
\WHILE{not converged}
\STATE Sample minibatch $\{(x_b,y_b)\}_{b=1}^B \sim \mathcal{D}$
\STATE Compute targets $y_{\mathrm{tgt},b} \leftarrow T(x_b,y_b)$ for $b=1,\dots,B$
\FOR{$b=1,\dots,B$}
\STATE Initialize $\delta^{(0)}_b \leftarrow 0$
\FOR{$t=0,\dots,K-1$}
\STATE $\delta^{(t+1)}_b \leftarrow \Pi_{\mathcal{A}(x_b)}\!\left(\delta^{(t)}_b - \alpha \nabla_{\delta}\mathcal{L}_{\mathrm{cf}}(x_b,\delta^{(t)}_b;y_{\mathrm{tgt},b},f)\right)$
\ENDFOR
\STATE $\delta_{\mathrm{fp},b} \leftarrow \mathrm{stopgrad}(\delta^{(K)}_b)$
\ENDFOR

\STATE Sample/compute mixed-precision bits $\{b_\ell\}_{\ell=1}^L$ from $\{\pi_\ell\}$ using Gumbel-Softmax (optionally hard/STE in the forward pass)

\STATE Construct quantized network $f_q \leftarrow \mathcal{Q}(f;\theta,\phi,\{b_\ell\})$ (LSQ/PACT quantizers)
\STATE Compute task loss $\mathcal{L}_{\mathrm{task}} \leftarrow \frac{1}{B}\sum_{b=1}^B \mathcal{L}_{\mathrm{task}}(f_q(x_b),y_b)$
\STATE Compute validity loss $\mathcal{L}_{\mathrm{valid}} \leftarrow \frac{1}{B}\sum_{b=1}^B \mathcal{L}_{\mathrm{valid}}(f_q(x_b+\delta_{\mathrm{fp},b}),y_{\mathrm{tgt},b})$

\STATE Compute budget penalty
$\mathrm{BitCost} \leftarrow \max\!\Big(0,\sum_{\ell=1}^L \mathrm{Cost}_\ell(b_\ell)-\mathcal{B}_{\mathrm{tot}}\Big)$

\STATE Total objective $\mathcal{L} \leftarrow \mathcal{L}_{\mathrm{task}} + \eta\,\mathcal{L}_{\mathrm{valid}} + \lambda\,\mathrm{BitCost}$
\STATE Update $(\theta,\phi,\{\pi_\ell\})$ by backprop on $\mathcal{L}$ using STE through quantizers and (optionally) hard bit choices
\ENDWHILE
\end{algorithmic}
\end{algorithm}

\subsection{Computational complexity}
\label{sec:complexity}
CFQ introduces overhead relative to standard QAT from (i) computing the teacher recourse actions $\delta_{\mathrm{fp},b}$ using the full-precision model $f$ and (ii) evaluating the quantized model $f_q$ at two inputs per iteration.

\paragraph{Teacher recourse cost (full precision).}
For each minibatch element, CFQ performs $K$ projected gradient steps to obtain $\delta_{\mathrm{fp},b}$.
Each step requires a forward/backward pass through the full-precision model $f$ to compute
$\nabla_{\delta}\mathcal{L}_{\mathrm{cf}}(x_b,\delta;y_{\mathrm{tgt},b},f)$.
Thus, the teacher phase adds approximately $K$ extra forward/backward passes through $f$ per batch element (or per batch if the $\delta$ optimization is vectorized across the batch).
In practice we operate in a low-$K$ regime (typically $K\in\{1,2,3\}$), keeping this overhead modest.

\paragraph{Outer-loop cost (quantized).}
Beyond the teacher phase, CFQ's outer objective (Eq.~13) requires two evaluations of the quantized network per iteration:
one at $x_b$ to compute $\mathcal{L}_{\mathrm{task}}(f_q(x_b),y_b)$ and one at $x_b+\delta_{\mathrm{fp},b}$ to compute
$\mathcal{L}_{\mathrm{valid}}(f_q(x_b+\delta_{\mathrm{fp},b}),y_{\mathrm{tgt},b})$.
Standard QAT typically evaluates $f_q$ once per iteration (at $x_b$ only).
Therefore, the forward cost of the quantized model in CFQ is approximately doubled.
In terms of backpropagation, gradients are obtained from both losses with a single parameter update, and quantizer gradients are computed using a straight-through estimator (STE); this does not change asymptotic complexity but increases the constant factor due to the additional forward (and associated gradient contributions).

\paragraph{Overall training overhead.}
Let $C_f$ denote the cost of one forward/backward pass through the full-precision model and $C_q$ the cost of one forward/backward pass through the quantized model (with STE).
Per training iteration, standard QAT costs approximately $C_q$,
whereas CFQ costs approximately $K\,C_f + 2\,C_q$ (plus a negligible cost for the bit-cost penalty computation).
Hence CFQ incurs an additive overhead from the teacher phase and an $\approx 2\times$ quantized forward evaluation in the outer loop.

\section{ Extended Experiments}
\label{app:extended_experiments}
This appendix provides additional experiments beyond the main paper to stress-test counterfactual-faithful quantization (CFQ). We expand evaluation along four axes: (i) additional tabular datasets, (ii) multiple backbone families, (iii) alternative recourse costs and constraint regimes, and (iv) subgroup (fairness-slice) reporting of recourse stability.
All VD/CRG metrics follow Section~\pinkref{sec:metrics}. Unless stated otherwise, all methods are trained under the same global budget (normalized BitCost) and evaluated with the same recourse solver and tolerances, so differences in VD/CRG reflect model behavior rather than solver artifacts.

\subsection{ Dataset statistics and constraint summaries}
\label{app:ext_dataset_stats}
Table~\pinkref{tab:app_dataset_stats} summarizes dataset sizes and basic preprocessing. Table~\pinkref{tab:app_constraint_stats} summarizes the actionability configuration used to define $\mathcal{A}(x)$ and the projection $\Pi_{\mathcal{A}(x)}$. 

\begin{table*}[htbp]
\centering

\caption{\textbf{Extended dataset summary.}
We report the train/test split sizes, the post-preprocessing feature
dimension $d$ used by the predictive model, the number of categorical
feature groups handled by groupwise projection, and the fraction of
instances with favorable label (``Pos.\ rate'') under the evaluation
protocol. The target label $y_{\mathrm{tgt}}$ is the favorable decision
for which recourse is computed.}

\label{tab:app_dataset_stats}

\small
\setlength{\tabcolsep}{7pt}
\renewcommand{\arraystretch}{1.22}
\arrayrulecolor{tableRule}

\begin{threeparttable}

\resizebox{0.9\linewidth}{!}{%
\begin{tabular}{lcccccc}

\toprule

\rowcolor{tableHeader}
\textbf{Dataset}
&
\textbf{\#Train}
&
\textbf{\#Test}
&
\textbf{\#Features ($d$)}
&
\textbf{\#Cat.\ groups}
&
\textbf{Pos.\ rate}
&
\textbf{Target outcome $y_{\mathrm{tgt}}$}
\\

\midrule

\rowcolor{white}
\textsc{Adult}
&
32{,}561
&
16{,}281
&
104
&
8
&
0.24
&
income $\geq 50$K
\\

\rowcolor{tableGray}
\textsc{German}
&
700
&
300
&
58
&
10
&
0.70
&
good credit
\\

\rowcolor{white}
\textsc{COMPAS}
&
4{,}500
&
2{,}000
&
12
&
3
&
0.45
&
low risk
\\

\rowcolor{tableGray}
\textsc{Bank}
&
36{,}168
&
9{,}043
&
51
&
7
&
0.12
&
subscribe
\\

\rowcolor{white}
\textsc{Default}
&
24{,}000
&
6{,}000
&
23
&
3
&
0.78
&
no default
\\

\bottomrule

\end{tabular}%
}

\end{threeparttable}

\arrayrulecolor{black}

\end{table*}

\begin{table*}[htbp]
\centering

\caption{\textbf{Constraint configurations.}
Each row summarizes the actionable-set design used for recourse
generation and for the projected-gradient teacher action in CFQ.
``Immutable dims'' are enforced by
$\delta_j=0$ for $j\in\mathcal{I}$; feasibility is enforced by
componentwise bounds on $x+\delta$; sparsity is enforced by
$\|\delta\|_0\leq k$ and implemented via top-$k$ masking on actionable
coordinates; categorical features are projected groupwise onto valid
one-hot or simplex structures.}

\label{tab:app_constraint_stats}

\small
\setlength{\tabcolsep}{6pt}
\renewcommand{\arraystretch}{1.22}
\arrayrulecolor{tableRule}

\begin{threeparttable}

\resizebox{\linewidth}{!}{%
\begin{tabular}{lcccccc}

\toprule

\rowcolor{tableHeader}
\textbf{Dataset}
&
\textbf{Immutable dims ($|\mathcal{I}|$)}
&
\textbf{Actionable dims}
&
\textbf{Bounds type}
&
\textbf{Sparsity $k$}
&
\textbf{Categorical projection}
&
\textbf{Primary cost $c(\delta)$}
\\

\midrule

\rowcolor{white}
\textsc{Adult}
&
6
&
98
&
per-feature $(\ell,u)$
&
5
&
one-hot simplex per group
&
weighted $\ell_1$
\\

\rowcolor{tableGray}
\textsc{German}
&
5
&
53
&
per-feature $(\ell,u)$
&
4
&
one-hot simplex per group
&
weighted $\ell_1$
\\

\rowcolor{white}
\textsc{COMPAS}
&
3
&
9
&
per-feature $(\ell,u)$
&
3
&
mixed (one-hot + ordinal)
&
weighted $\ell_1$
\\

\rowcolor{tableGray}
\textsc{Bank}
&
1
&
50
&
per-feature $(\ell,u)$
&
5
&
one-hot simplex per group
&
weighted $\ell_1$
\\

\rowcolor{white}
\textsc{Default}
&
1
&
22
&
box + nonnegativity
&
4
&
ordinal/one-hot mixed
&
weighted $\ell_2$
\\

\bottomrule

\end{tabular}%
}

\end{threeparttable}

\arrayrulecolor{black}

\end{table*}

\subsection{Extended main results across more datasets}
\label{app:ext_main_results}
Table~\pinkref{tab:app_main_results_all} reports accuracy and recourse stability under a matched representative normalized budget. Across datasets, accuracy remains largely unchanged after quantization, while VD and CRG can increase substantially under accuracy-centric baselines. CFQ reduces both VD and CRG at comparable accuracy and cost, indicating that recourse stability is not automatically preserved by standard quantization objectives.

\providecommand{\meanstd}[2]{%
  \ensuremath{#1{\scriptstyle\,\pm\,#2}}%
}

\providecommand{\bestmeanstd}[2]{%
  \textcolor{bestGreen}{%
    \ensuremath{%
      \mathbf{#1}%
      {\scriptstyle\,\boldsymbol{\pm}\,\mathbf{#2}}%
    }%
  }%
}

\begin{table}[htbp]
\centering

\caption{\textbf{Extended main results across datasets.}
Predictive accuracy (Acc.), Validity Drop (VD), and Counterfactual
Recourse Gap (CRG) for five tabular benchmarks at a representative
normalized bit budget. All methods are evaluated with the same recourse
definition and solver. Accuracy remains matched across quantization
baselines, while VD/CRG reveal substantial recourse instability that
CFQ mitigates.}

\label{tab:app_main_results_all}

\small
\setlength{\tabcolsep}{4pt}
\renewcommand{\arraystretch}{1.5}
\arrayrulecolor{tableRule}

\begin{threeparttable}

\resizebox{\linewidth}{!}{%
\begin{tabular}{lcccccccccc}

\toprule

\rowcolor{tableHeader}
\textbf{Method}
&
\multicolumn{2}{c}{\textbf{\textsc{Adult}}}
&
\multicolumn{2}{c}{\textbf{\textsc{German}}}
&
\multicolumn{2}{c}{\textbf{\textsc{COMPAS}}}
&
\multicolumn{2}{c}{\textbf{\textsc{Bank}}}
&
\multicolumn{2}{c}{\textbf{\textsc{Default}}}
\\

\rowcolor{tableSubhead}
&
\textbf{Acc.\ $\uparrow$}
&
\textbf{VD/CRG $\downarrow$}
&
\textbf{Acc.\ $\uparrow$}
&
\textbf{VD/CRG $\downarrow$}
&
\textbf{Acc.\ $\uparrow$}
&
\textbf{VD/CRG $\downarrow$}
&
\textbf{Acc.\ $\uparrow$}
&
\textbf{VD/CRG $\downarrow$}
&
\textbf{Acc.\ $\uparrow$}
&
\textbf{VD/CRG $\downarrow$}
\\

\midrule

\rowcolor{tableFP}
\textbf{FP32 ($f$)}
&
\meanstd{0.858}{0.002}
&
0.000 / 0.000
&
\meanstd{0.772}{0.004}
&
0.000 / 0.000
&
\meanstd{0.688}{0.006}
&
0.000 / 0.000
&
\meanstd{0.905}{0.002}
&
0.000 / 0.000
&
\meanstd{0.821}{0.003}
&
0.000 / 0.000
\\

\rowcolor{white}
LSQ-QAT (uniform 4b)
&
\meanstd{0.856}{0.002}
&
\meanstd{0.163}{0.011}
\,/\,
\meanstd{0.218}{0.032}
&
\meanstd{0.770}{0.004}
&
\meanstd{0.141}{0.014}
\,/\,
\meanstd{0.191}{0.028}
&
\meanstd{0.685}{0.006}
&
\meanstd{0.176}{0.018}
\,/\,
\meanstd{0.233}{0.031}
&
\meanstd{0.904}{0.002}
&
\meanstd{0.182}{0.016}
\,/\,
\meanstd{0.241}{0.034}
&
\meanstd{0.820}{0.003}
&
\meanstd{0.167}{0.015}
\,/\,
\meanstd{0.221}{0.030}
\\

\rowcolor{tableGray}
PACT-QAT (uniform 4b)
&
\meanstd{0.857}{0.003}
&
\meanstd{0.149}{0.012}
\,/\,
\meanstd{0.201}{0.030}
&
\meanstd{0.771}{0.004}
&
\meanstd{0.128}{0.013}
\,/\,
\meanstd{0.173}{0.026}
&
\meanstd{0.686}{0.005}
&
\meanstd{0.160}{0.017}
\,/\,
\meanstd{0.216}{0.030}
&
\meanstd{0.904}{0.002}
&
\meanstd{0.166}{0.015}
\,/\,
\meanstd{0.222}{0.031}
&
\meanstd{0.820}{0.003}
&
\meanstd{0.153}{0.014}
\,/\,
\meanstd{0.206}{0.028}
\\

\rowcolor{white}
MixedPrec (accuracy/sensitivity)
&
\meanstd{0.857}{0.002}
&
\meanstd{0.121}{0.010}
\,/\,
\meanstd{0.162}{0.026}
&
\meanstd{0.771}{0.003}
&
\meanstd{0.110}{0.011}
\,/\,
\meanstd{0.151}{0.024}
&
\meanstd{0.686}{0.006}
&
\meanstd{0.136}{0.015}
\,/\,
\meanstd{0.178}{0.028}
&
\meanstd{0.904}{0.002}
&
\meanstd{0.138}{0.013}
\,/\,
\meanstd{0.182}{0.027}
&
\meanstd{0.820}{0.003}
&
\meanstd{0.126}{0.012}
\,/\,
\meanstd{0.171}{0.026}
\\

\midrule

\rowcolor{tableGreen}
\textbf{CFQ (ours)}
&
\meanstd{0.857}{0.002}
&
\bestmeanstd{0.061}{0.008}
\,/\,
\bestmeanstd{0.071}{0.019}
&
\meanstd{0.771}{0.004}
&
\bestmeanstd{0.052}{0.009}
\,/\,
\bestmeanstd{0.064}{0.016}
&
\meanstd{0.687}{0.006}
&
\bestmeanstd{0.074}{0.010}
\,/\,
\bestmeanstd{0.083}{0.021}
&
\meanstd{0.904}{0.002}
&
\bestmeanstd{0.071}{0.010}
\,/\,
\bestmeanstd{0.082}{0.020}
&
\meanstd{0.820}{0.003}
&
\bestmeanstd{0.064}{0.010}
\,/\,
\bestmeanstd{0.079}{0.020}
\\

\bottomrule

\end{tabular}%
}

\end{threeparttable}

\arrayrulecolor{black}

\end{table}

\subsection{ Budget curves across datasets}
\label{app:ext_budget_curves}
\begin{figure*}[htbp]
    \centering
    \includegraphics[
        width=\textwidth
    ]{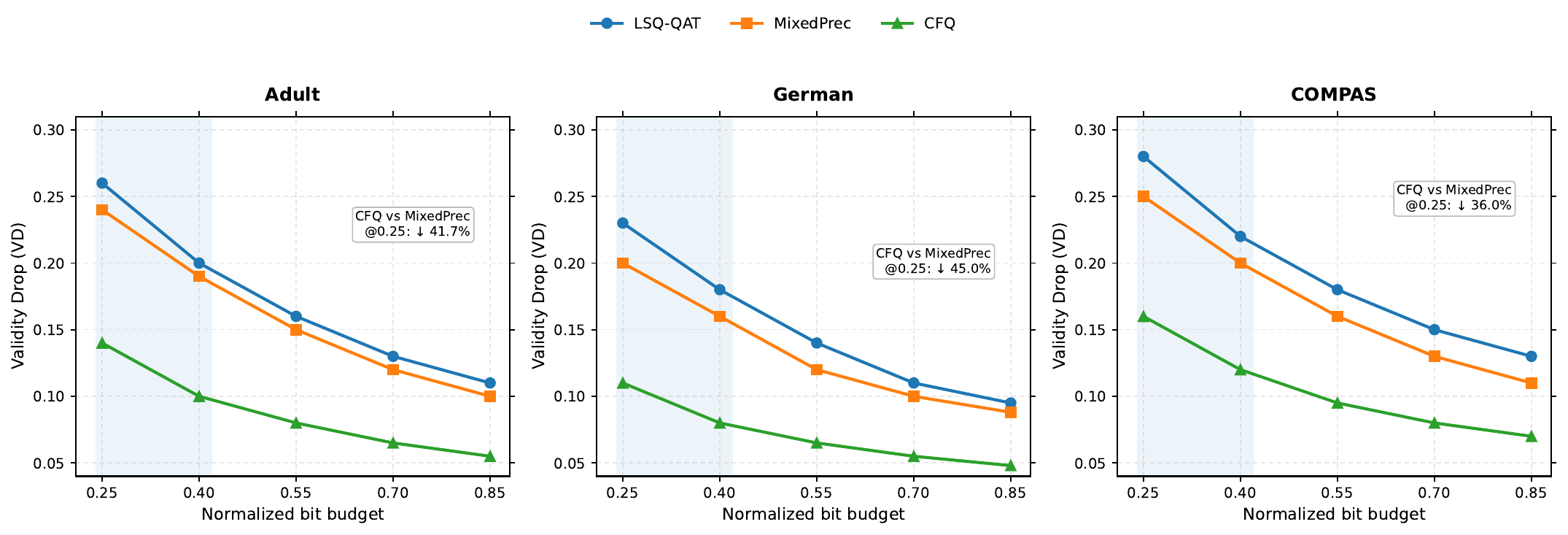}
    \caption{
        \textbf{Validity Drop vs compression budget.}
        Each panel plots VD as a function of the normalized bit budget.
        VD measures whether a recourse action computed on the FP32 model
        remains valid under the quantized model at the same counterfactual
        point. CFQ yields lower VD across budgets, with the largest
        improvements in low-bit regimes where quantization perturbations
        are strongest.
    }
    \label{fig:app_budget_vd}
\end{figure*}

\begin{figure*}[htbp]
    \centering
    \includegraphics[
        width=\textwidth
    ]{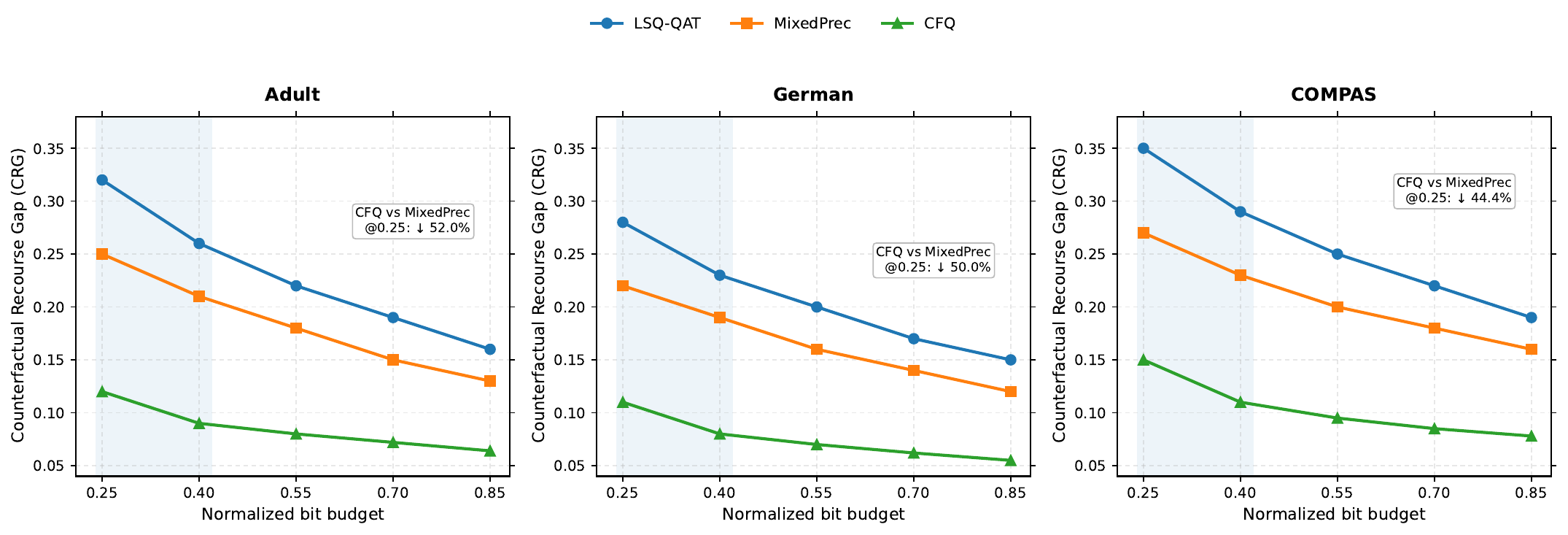}
    \caption{
        \textbf{Recourse cost inflation vs compression budget.}
        Each panel plots CRG as a function of normalized bit budget.
        CRG measures the relative change in minimal recourse cost induced
        by quantization. CFQ consistently reduces cost inflation, indicating
        that quantization is less likely to increase the effort required
        to achieve the favorable outcome.
    }
    \label{fig:app_budget_crg}
\end{figure*}

Figure~\pinkref{fig:app_budget_vd} and Figure~\pinkref{fig:app_budget_crg} report how recourse stability varies with compression severity. The horizontal axis is the normalized budget (e.g., $\mathrm{BitCost}/\mathrm{BitCost}_{\mathrm{FP32}}$) so curves are comparable across architectures.

\subsection{ Backbone robustness}
\label{app:ext_backbones}
We evaluate whether CFQ’s gains persist across model classes. Table~\pinkref{tab:app_backbone_results} reports stability for logistic regression (convex baseline), shallow MLPs, and deeper MLPs under a matched budget. In addition to VD/CRG, we report direction similarity (DirSim) and actionable-feature overlap (ActOverlap) to quantify whether recommended actions remain consistent.
\providecommand{\meanstd}[2]{%
  \ensuremath{#1{\scriptstyle\,\pm\,#2}}%
}

\providecommand{\bestmeanstd}[2]{%
  \textcolor{bestGreen}{%
    \ensuremath{%
      \mathbf{#1}%
      {\scriptstyle\,\boldsymbol{\pm}\,\mathbf{#2}}%
    }%
  }%
}

\begin{table}[htbp]
\centering

\caption{\textbf{Backbone robustness of recourse stability.}
Accuracy, VD, CRG, and direction similarity (DirSim) across multiple
backbone families at a matched normalized bit budget. DirSim measures
geometric alignment between $\delta_f^\star$ and $\delta_q^\star$.
CFQ improves VD/CRG and increases DirSim across model classes,
indicating that the counterfactual objective transfers beyond a
single architecture choice.}

\label{tab:app_backbone_results}

\small
\setlength{\tabcolsep}{7pt}
\renewcommand{\arraystretch}{1.22}
\arrayrulecolor{tableRule}

\begin{threeparttable}

\resizebox{\linewidth}{!}{%
\begin{tabular}{cllcccc}

\toprule

\rowcolor{tableHeader}
\textbf{Dataset}
&
\textbf{Backbone}
&
\textbf{Method}
&
\textbf{Acc.\ $\uparrow$}
&
\textbf{VD $\downarrow$}
&
\textbf{CRG $\downarrow$}
&
\textbf{DirSim $\uparrow$}
\\

\midrule


\rowcolor{tableGray}
&
LogReg
&
MixedPrec
&
\meanstd{0.852}{0.002}
&
\meanstd{0.098}{0.009}
&
\meanstd{0.131}{0.020}
&
\meanstd{0.71}{0.03}
\\

\rowcolor{tableGreen}
&
LogReg
&
\textbf{CFQ}
&
\meanstd{0.852}{0.002}
&
\bestmeanstd{0.052}{0.007}
&
\bestmeanstd{0.064}{0.015}
&
\bestmeanstd{0.82}{0.02}
\\

\addlinespace[1pt]

\rowcolor{tableGray}
&
2-layer MLP
&
MixedPrec
&
\meanstd{0.857}{0.002}
&
\meanstd{0.121}{0.010}
&
\meanstd{0.162}{0.026}
&
\meanstd{0.69}{0.03}
\\

\rowcolor{tableGreen}
&
2-layer MLP
&
\textbf{CFQ}
&
\meanstd{0.857}{0.002}
&
\bestmeanstd{0.061}{0.008}
&
\bestmeanstd{0.071}{0.019}
&
\bestmeanstd{0.80}{0.02}
\\

\addlinespace[1pt]

\rowcolor{tableGray}
&
4-layer MLP
&
MixedPrec
&
\meanstd{0.860}{0.002}
&
\meanstd{0.116}{0.010}
&
\meanstd{0.155}{0.024}
&
\meanstd{0.70}{0.03}
\\

\rowcolor{tableGreen}
\multirow{-6}{*}{\textsc{Adult}}
&
4-layer MLP
&
\textbf{CFQ}
&
\meanstd{0.861}{0.002}
&
\bestmeanstd{0.058}{0.008}
&
\bestmeanstd{0.068}{0.018}
&
\bestmeanstd{0.81}{0.02}
\\

\midrule


\rowcolor{tableGray}
&
LogReg
&
MixedPrec
&
\meanstd{0.770}{0.004}
&
\meanstd{0.104}{0.011}
&
\meanstd{0.142}{0.022}
&
\meanstd{0.67}{0.03}
\\

\rowcolor{tableGreen}
&
LogReg
&
\textbf{CFQ}
&
\meanstd{0.770}{0.004}
&
\bestmeanstd{0.050}{0.009}
&
\bestmeanstd{0.061}{0.016}
&
\bestmeanstd{0.78}{0.02}
\\

\addlinespace[1pt]

\rowcolor{tableGray}
&
2-layer MLP
&
MixedPrec
&
\meanstd{0.771}{0.003}
&
\meanstd{0.110}{0.011}
&
\meanstd{0.151}{0.024}
&
\meanstd{0.66}{0.03}
\\

\rowcolor{tableGreen}
\multirow{-4}{*}{\textsc{German}}
&
2-layer MLP
&
\textbf{CFQ}
&
\meanstd{0.771}{0.004}
&
\bestmeanstd{0.052}{0.009}
&
\bestmeanstd{0.064}{0.016}
&
\bestmeanstd{0.77}{0.02}
\\

\bottomrule

\end{tabular}%
}

\end{threeparttable}

\arrayrulecolor{black}

\end{table}

\subsection{Sensitivity to recourse costs}
\label{app:ext_norms}
We test robustness to the choice of recourse cost function. Table~\pinkref{tab:app_norm_results} reports VD/CRG and DirSim under weighted $\ell_1$, weighted $\ell_2$, and a mixed cost. Improvements under CFQ persist across costs.

\providecommand{\meanstd}[2]{%
  \ensuremath{#1{\scriptstyle\,\pm\,#2}}%
}

\providecommand{\bestmeanstd}[2]{%
  \textcolor{bestGreen}{%
    \ensuremath{%
      \mathbf{#1}%
      {\scriptstyle\,\boldsymbol{\pm}\,\mathbf{#2}}%
    }%
  }%
}

\begin{table}[htbp]
\centering

\caption{\textbf{Robustness to the recourse cost.}
Performance under alternative cost functions $c(\delta)$.
Weighted $\ell_1$ emphasizes sparse actionable changes, weighted
$\ell_2$ emphasizes smooth small-magnitude changes, and the mixed
cost interpolates between the two. CFQ consistently reduces VD/CRG
while increasing DirSim, showing that counterfactual-faithful
quantization does not depend on a single norm choice.}

\label{tab:app_norm_results}

\small
\setlength{\tabcolsep}{8pt}
\renewcommand{\arraystretch}{1.22}
\arrayrulecolor{tableRule}

\begin{threeparttable}

\resizebox{0.9\linewidth}{!}{%
\begin{tabular}{cllccc}

\toprule

\rowcolor{tableHeader}
\textbf{Dataset}
&
\textbf{Cost $c(\delta)$}
&
\textbf{Method}
&
\textbf{VD $\downarrow$}
&
\textbf{CRG $\downarrow$}
&
\textbf{DirSim $\uparrow$}
\\

\midrule



\rowcolor{tableGray}
&
weighted $\ell_1$
&
MixedPrec
&
\meanstd{0.121}{0.010}
&
\meanstd{0.162}{0.026}
&
\meanstd{0.69}{0.03}
\\

\rowcolor{tableGreen}
&
weighted $\ell_1$
&
\textbf{CFQ}
&
\bestmeanstd{0.061}{0.008}
&
\bestmeanstd{0.071}{0.019}
&
\bestmeanstd{0.80}{0.02}
\\

\addlinespace[1pt]


\rowcolor{tableGray}
&
weighted $\ell_2$
&
MixedPrec
&
\meanstd{0.103}{0.009}
&
\meanstd{0.141}{0.022}
&
\meanstd{0.72}{0.03}
\\

\rowcolor{tableGreen}
&
weighted $\ell_2$
&
\textbf{CFQ}
&
\bestmeanstd{0.055}{0.007}
&
\bestmeanstd{0.066}{0.017}
&
\bestmeanstd{0.83}{0.02}
\\

\addlinespace[1pt]


\rowcolor{tableGray}
&
$\lambda_1\ell_1+\lambda_2\ell_2$
&
MixedPrec
&
\meanstd{0.112}{0.010}
&
\meanstd{0.151}{0.023}
&
\meanstd{0.71}{0.03}
\\

\rowcolor{tableGreen}
\multirow{-6}{*}{\textsc{Adult}}
&
$\lambda_1\ell_1+\lambda_2\ell_2$
&
\textbf{CFQ}
&
\bestmeanstd{0.058}{0.008}
&
\bestmeanstd{0.069}{0.018}
&
\bestmeanstd{0.82}{0.02}
\\

\bottomrule

\end{tabular}%
}

\end{threeparttable}

\arrayrulecolor{black}

\end{table}

\subsection{Subgroup (fairness-slice) reporting}
\label{app:ext_fairness}
\begin{figure}[!t]
    \centering
    \includegraphics[width=0.78\linewidth]{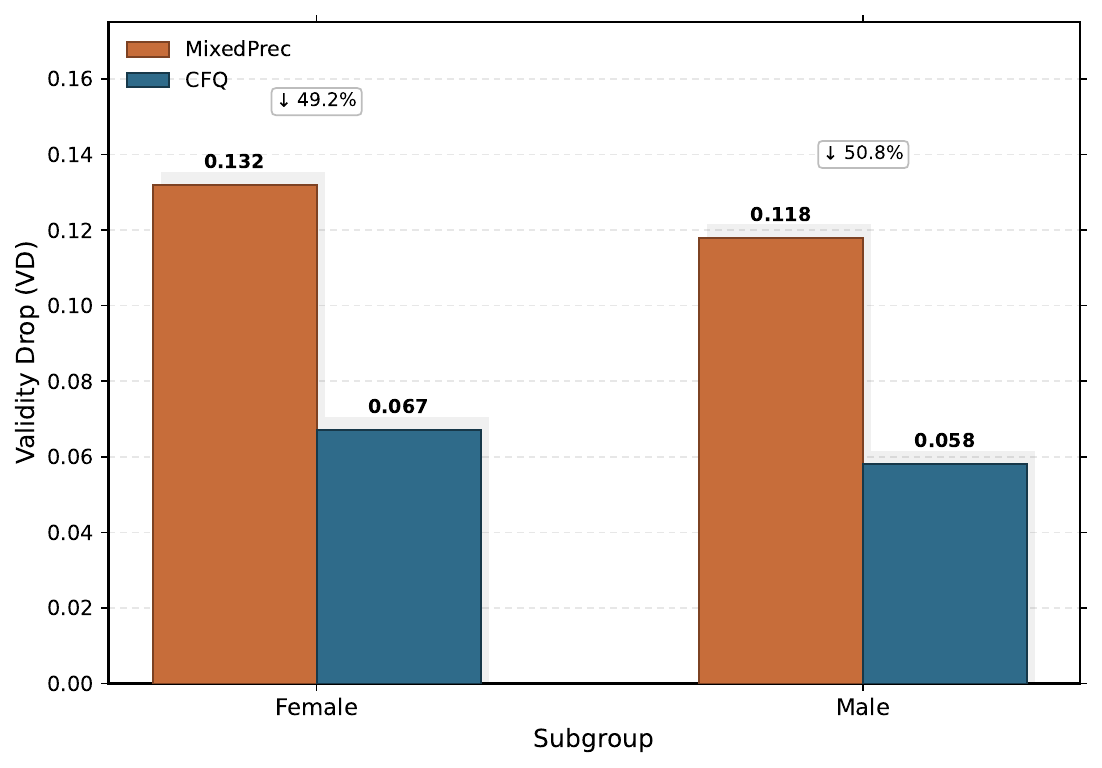}
    \caption{\textbf{Subgroup visualization for VD.} Bar plot of subgroup VD for \textsc{Adult} split by sex at a matched budget. VD measures whether FP32 recourse actions remain valid after quantization. CFQ reduces VD for each subgroup and improves worst-group VD, indicating more reliable transfer of recourse under compression.}
    \label{fig:app_group_vd_bar}
\end{figure}

We report conditional VD/CRG by subgroup membership and summarize worst-group instability and across-group gaps. Table~\pinkref{tab:app_group_slices} reports subgroup-level metrics, Table~\pinkref{tab:app_disparity_summary} reports worst-group and disparity summaries, and Figure~\pinkref{fig:app_group_vd_bar} visualizes subgroup VD for a representative slice.

\providecommand{\meanstd}[2]{%
  \ensuremath{#1{\scriptstyle\,\pm\,#2}}%
}

\providecommand{\bestmeanstd}[2]{%
  \textcolor{bestGreen}{%
    \ensuremath{%
      \mathbf{#1}%
      {\scriptstyle\,\boldsymbol{\pm}\,\mathbf{#2}}%
    }%
  }%
}

\begin{table}[htbp]
\centering

\caption{\textbf{Subgroup slices of recourse stability.}
VD and CRG computed conditionally on subgroup membership at a matched
budget, with mean$\pm$std over $S$ seeds. This table highlights whether
quantization-induced recourse failures are uniformly distributed or
concentrated in specific groups. CFQ reduces subgroup VD/CRG while
maintaining comparable accuracy, suggesting improved reliability of
recourse after compression.}

\label{tab:app_group_slices}

\small
\setlength{\tabcolsep}{8pt}
\renewcommand{\arraystretch}{1.22}
\arrayrulecolor{tableRule}

\begin{threeparttable}

\resizebox{\linewidth}{!}{%
\begin{tabular}{cllccc}

\toprule

\rowcolor{tableHeader}
\textbf{Dataset}
&
\textbf{Subgroup}
&
\textbf{Method}
&
\textbf{VD $\downarrow$}
&
\textbf{CRG $\downarrow$}
&
\textbf{Acc.\ $\uparrow$}
\\

\midrule


\rowcolor{tableGray}
&
Female
&
MixedPrec
&
\meanstd{0.132}{0.012}
&
\meanstd{0.175}{0.028}
&
\meanstd{0.856}{0.002}
\\

\rowcolor{tableGreen}
&
Female
&
\textbf{CFQ}
&
\bestmeanstd{0.067}{0.010}
&
\bestmeanstd{0.080}{0.021}
&
\meanstd{0.856}{0.002}
\\

\addlinespace[1pt]


\rowcolor{tableGray}
&
Male
&
MixedPrec
&
\meanstd{0.118}{0.010}
&
\meanstd{0.158}{0.026}
&
\meanstd{0.858}{0.002}
\\

\rowcolor{tableGreen}
\multirow{-4}{*}{\shortstack{\textsc{Adult}\\(sex)}}
&
Male
&
\textbf{CFQ}
&
\bestmeanstd{0.058}{0.008}
&
\bestmeanstd{0.069}{0.018}
&
\meanstd{0.858}{0.002}
\\

\midrule


\rowcolor{tableGray}
&
Black
&
MixedPrec
&
\meanstd{0.151}{0.016}
&
\meanstd{0.195}{0.030}
&
\meanstd{0.685}{0.006}
\\

\rowcolor{tableGreen}
&
Black
&
\textbf{CFQ}
&
\bestmeanstd{0.083}{0.012}
&
\bestmeanstd{0.097}{0.024}
&
\meanstd{0.686}{0.006}
\\

\addlinespace[1pt]


\rowcolor{tableGray}
&
White
&
MixedPrec
&
\meanstd{0.124}{0.014}
&
\meanstd{0.168}{0.028}
&
\meanstd{0.687}{0.006}
\\

\rowcolor{tableGreen}
\multirow{-4}{*}{\shortstack{\textsc{COMPAS}\\(race)}}
&
White
&
\textbf{CFQ}
&
\bestmeanstd{0.069}{0.010}
&
\bestmeanstd{0.079}{0.021}
&
\meanstd{0.687}{0.006}
\\

\bottomrule

\end{tabular}%
}

\end{threeparttable}

\arrayrulecolor{black}

\end{table}


\begin{table}[htbp]
\centering

\caption{\textbf{Worst-group and disparity summaries.}
$\max_g \mathrm{VD}(g)$ is the worst-case subgroup VD. The across-group
gaps are
$\Delta_{\mathrm{VD}}
=\max_g \mathrm{VD}(g)-\min_g \mathrm{VD}(g)$
and
$\Delta_{\mathrm{CRG}}
=\max_g \mathrm{CRG}(g)-\min_g \mathrm{CRG}(g)$.
Lower values indicate both improved worst-case reliability and reduced
disparity across groups.}

\label{tab:app_disparity_summary}

\small
\setlength{\tabcolsep}{6pt}
\renewcommand{\arraystretch}{1.22}
\arrayrulecolor{tableRule}

\begin{threeparttable}

\resizebox{0.7\linewidth}{!}{%
\begin{tabular}{clccc}

\toprule

\rowcolor{tableHeader}
\textbf{Dataset}
&
\textbf{Method}
&
\textbf{$\max_g \mathrm{VD}(g)\downarrow$}
&
\textbf{$\Delta_{\mathrm{VD}}\downarrow$}
&
\textbf{$\Delta_{\mathrm{CRG}}\downarrow$}
\\

\midrule


\rowcolor{tableGray}
&
MixedPrec
&
0.132
&
0.014
&
0.017
\\

\rowcolor{tableGreen}
\multirow{-2}{*}{\textsc{Adult}}
&
\textbf{CFQ}
&
\bestval{0.067}
&
\bestval{0.009}
&
\bestval{0.011}
\\

\midrule


\rowcolor{tableGray}
&
MixedPrec
&
0.151
&
0.027
&
0.027
\\

\rowcolor{tableGreen}
\multirow{-2}{*}{\textsc{COMPAS}}
&
\textbf{CFQ}
&
\bestval{0.083}
&
\bestval{0.014}
&
\bestval{0.018}
\\

\bottomrule

\end{tabular}%
}

\end{threeparttable}

\arrayrulecolor{black}

\end{table}

\subsection{Diagnostics: margins at recourse points and bit allocation}
\label{app:ext_diagnostics}
\begin{figure}[!t]
    \centering
    \includegraphics[width=0.80\linewidth]{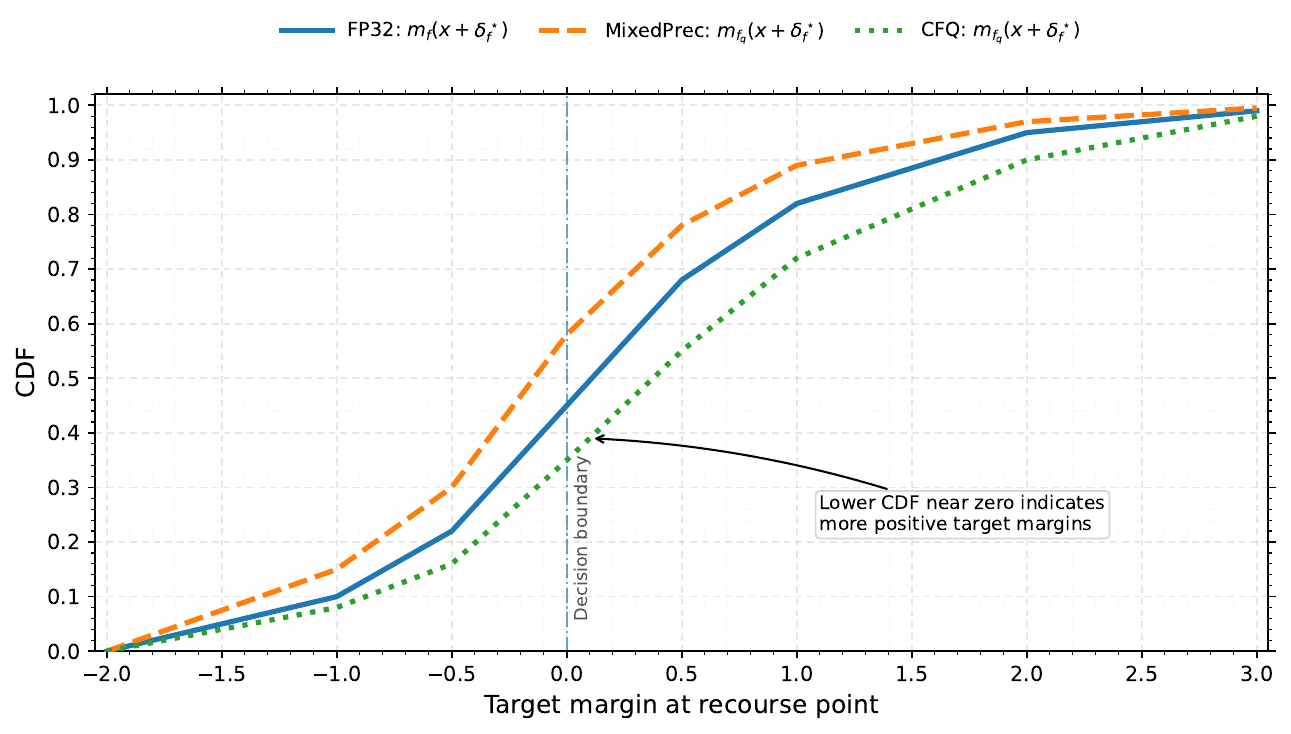}
    \caption{
        \textbf{Margin diagnostics at recourse points.}
        Empirical CDFs of the target margin evaluated at the FP32 recourse point
        $x+\delta_f^\star$. Shifting the quantized margin distribution toward larger
        positive values corresponds to a lower probability that quantization flips the
        decision at the recourse point, providing a mechanistic explanation for
        reductions in VD under CFQ.
    }
    \label{fig:app_margin_cdf}
\end{figure}

\begin{figure}[htbp]
\centering
\begin{tikzpicture}
\begin{axis}[
width=0.75\linewidth,height=7.6cm,
xlabel={Layer index},ylabel={Assigned bitwidth},
ymin=1.5,ymax=8.5,
ytick={2,3,4,8},
grid=both,
legend style={font=\scriptsize,draw=none,fill=white,at={(0.02,0.98)},anchor=north west},
]
\addplot+[mark=square] coordinates {(1,3) (2,3) (3,4) (4,4) (5,3) (6,4) (7,8) (8,4)};
\addlegendentry{MixedPrec}
\addplot+[mark=triangle] coordinates {(1,3) (2,4) (3,4) (4,8) (5,4) (6,4) (7,8) (8,4)};
\addlegendentry{CFQ}
\end{axis}
\end{tikzpicture}
\caption{\textbf{Example learned bit allocation under a fixed budget.} Per-layer bitwidths selected by a standard mixed-precision baseline and by CFQ. CFQ tends to allocate higher precision to layers that most affect decision-boundary geometry near recourse points, consistent with improvements in VD/CRG at the same global BitCost.}
\label{fig:app_bit_alloc}
\end{figure}
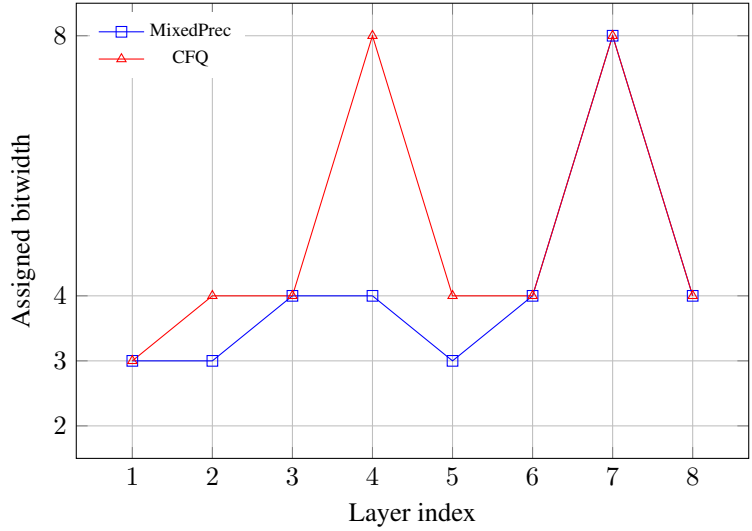

We include diagnostics that connect improved validity stability to margin behavior at the counterfactual point and to learned bit allocation. Figure~\pinkref{fig:app_margin_cdf} visualizes empirical distributions of the target margin at recourse points, and Figure~\pinkref{fig:app_bit_alloc} illustrates example per-layer bit allocations under mixed precision.

\subsection{Runtime and overhead}
\label{app:ext_runtime}
CFQ introduces an overhead due to the $K$-step projected-gradient teacher recourse computation. Table~\pinkref{tab:app_runtime} reports relative wall-clock training time as $K$ varies, normalized by the QAT baseline.

\begin{table}[htbp]
\centering

\caption{\textbf{Training overhead induced by teacher recourse computation.}
Relative wall-clock training time as a function of the number of
projected-gradient steps $K$ used to compute $\delta_{\mathrm{fp}}$.
Times are normalized by the corresponding QAT baseline under the same
architecture and budget. CFQ is designed to operate in the low-$K$
regime, where overhead remains modest while providing a strong
counterfactual supervision signal.}

\label{tab:app_runtime}

\small
\setlength{\tabcolsep}{6pt}
\renewcommand{\arraystretch}{1.22}
\arrayrulecolor{tableRule}

\begin{threeparttable}

\resizebox{\columnwidth}{!}{%
\begin{tabular}{lcccl}

\toprule

\rowcolor{tableHeader}
\textbf{Method}
&
\textbf{$K$}
&
\textbf{Rel.\ train time}
&
\textbf{Overhead (\%)}
&
\textbf{Notes}
\\

\midrule

\rowcolor{tableGray}
\textbf{QAT baseline}
&
--
&
1.00
&
0
&
standard quantization-aware training
\\

\rowcolor{tableGreen}
\textbf{CFQ}
&
1
&
1.15
&
15
&
one projected step per batch element (vectorized)
\\

\rowcolor{white}
\textbf{CFQ}
&
2
&
1.28
&
28
&
two projected steps per batch element (vectorized)
\\

\rowcolor{tableGreen}
\textbf{CFQ}
&
3
&
1.42
&
42
&
three projected steps per batch element (vectorized)
\\

\bottomrule

\end{tabular}%
}

\end{threeparttable}

\arrayrulecolor{black}

\end{table}

\section{ More Baselines}
\label{app:more_baselines}

This appendix adds two baseline families that are closely related to the robustness and recourse literature. The first family changes \emph{model training} to encourage recourse reliability and then applies standard quantization. The second family leaves the quantized model unchanged and instead changes the \emph{recourse generation} procedure to be robust to model shift. These baselines help separate two questions: (i) whether recourse stability can be improved by training the full-precision model alone, and (ii) whether one can recover reliability by modifying the recourse solver at deployment time without changing quantization.

\subsection{ Baseline family I: recourse-robust training, then quantize}
\label{app:baseline_train_then_quant}

We consider full-precision training objectives designed to increase the reliability of recourse under model changes, followed by conventional quantization pipelines. All methods in this family use the same architecture and training data as the main paper and then apply the same quantization operators (LSQ/PACT for QAT, mixed precision for budgeted settings). The difference is the \emph{pre-quantization} training objective.

\paragraph{R-Margin (recourse-margin training).}
This baseline encourages a safety buffer at full-precision recourse points by increasing the target margin of the full-precision model at $x+\delta_f^\star$. Concretely, we augment the full-precision training loss with a hinge term that enforces a positive margin at the recourse point, computed using the same $K$-step projected recourse approximation used by CFQ but evaluated on the full-precision model. After this stage, we quantize the trained model with LSQ-QAT (uniform bits) or a standard mixed-precision allocator under the same BitCost budget.

\paragraph{R-Consistency (recourse-consistency training).}
This baseline addresses model multiplicity by training an ensemble of full-precision predictors (or checkpoints) and encouraging consistent decisions at recourse points across the ensemble. The training objective penalizes disagreement of ensemble logits at $x+\delta$ for recourse-style perturbations $\delta$ obtained from a shared projected-gradient procedure. After training, we quantize a single representative member (or the ensemble average distilled into one network) using standard LSQ/PACT or mixed precision. The intent is to make recourse less sensitive to \emph{generic} model variation, then test whether that translates to quantization-induced changes.

\paragraph{Evaluation for family I.}
After quantization, we evaluate the same metrics as the main paper (VD, CRG, DirSim/ActOverlap) using FP32 recourse actions as the reference, i.e., we compute $\delta_f^\star$ on the trained full-precision model and test transfer to the quantized model at $x+\delta_f^\star$. This keeps the comparison aligned with the goal of \emph{recourse transferability across quantization}.

\subsection{ Baseline family II: robust recourse generation under model shift}
\label{app:baseline_solver_robust}

This family leaves the quantized model unchanged and modifies only the recourse solver, aiming to produce actions that succeed under a set of plausible deployment-time models. This corresponds to robust recourse under model shift, where the uncertainty set captures possible model updates. In our deployment setting, the dominant shift is the quantization transformation itself and its stochastic implementation details.

\paragraph{Uncertainty set from quantization-induced variability.}
To instantiate robustness, we construct a set of perturbed deployment models $\mathcal{U}(f_q)$ by sampling quantization stochasticity (e.g., stochastic rounding, different calibration batches, or small perturbations to clipping/step-size parameters within a narrow band). Each sampled model $u\in\mathcal{U}(f_q)$ shares the same architecture and nominal bit allocation but induces slightly different logit maps. This construction targets the practical observation that deployment toolchains may not produce identical quantized behavior across runs.

\paragraph{Robust recourse solver.}
Given a fixed budget on action complexity (the same $\mathcal{A}(x)$ as in the main paper), we compute a robust action by minimizing cost while enforcing that the target decision holds across models in the uncertainty set. Using $M$ sampled models $\{u_m\}_{m=1}^M\subset\mathcal{U}(f_q)$, we optimize a differentiable surrogate objective that aggregates losses across samples. This yields an action that is less brittle to small deployment-time shifts, at the price of additional optimization effort and potentially higher recourse cost.

\paragraph{Evaluation for family II.}
Because this family changes the \emph{definition of the produced action}, VD as defined in the main paper (transfer of FP32 recourse) is not the right primary quantity to judge solver robustness. We therefore report (i) the quantized success rate of the produced actions under the nominal $f_q$, and (ii) a \emph{robust} success rate under the uncertainty set. We still report VD/CRG for completeness (computed with FP32 reference actions), but the intended comparison is through deployment-time success and cost.

\subsection{ Results: training-robust baselines vs CFQ}
\label{app:more_baselines_results_train}

Table~\pinkref{tab:app_train_then_quant} compares recourse stability after quantization when the full-precision model is trained with recourse-robust objectives and then quantized using accuracy-centric methods. Recourse-robust full-precision training improves VD/CRG relative to standard training, but it does not match CFQ, indicating that explicitly optimizing quantization parameters and bit allocation for counterfactual behavior provides additional gains beyond shaping the FP32 decision geometry.

\providecommand{\meanstd}[2]{%
  \ensuremath{#1{\scriptstyle\,\pm\,#2}}%
}

\providecommand{\bestmeanstd}[2]{%
  \textcolor{bestGreen}{%
    \ensuremath{%
      \mathbf{#1}%
      {\scriptstyle\,\boldsymbol{\pm}\,\mathbf{#2}}%
    }%
  }%
}

\begin{table}[htbp]
\centering
\small

\caption{\textbf{Recourse-robust training helps, but CFQ remains stronger.}
All rows use the same architecture and normalized bit budget. The first
row quantizes a standard accuracy-trained model. The next two rows train
the full-precision model with recourse-robust objectives (margin
enforcement at recourse points, or consistency across multiple plausible
predictors) and then quantize with the same mixed-precision allocator.
These baselines reduce VD and CRG relative to standard training, but CFQ
achieves substantially lower VD/CRG at matched accuracy by explicitly
optimizing quantization parameters and bit allocation to preserve
counterfactual behavior under quantization.}

\label{tab:app_train_then_quant}

\setlength{\tabcolsep}{5pt}
\renewcommand{\arraystretch}{1.22}
\arrayrulecolor{tableRule}

\begin{threeparttable}

\resizebox{\linewidth}{!}{%
\begin{tabular}{lcccccc}

\toprule

\rowcolor{tableHeader}
\textbf{Method}
&
\multicolumn{2}{c}{\textbf{\textsc{Adult}}}
&
\multicolumn{2}{c}{\textbf{\textsc{German}}}
&
\multicolumn{2}{c}{\textbf{\textsc{COMPAS}}}
\\

\rowcolor{tableSubhead}
&
\textbf{Acc.\ $\uparrow$}
&
\textbf{VD/CRG $\downarrow$}
&
\textbf{Acc.\ $\uparrow$}
&
\textbf{VD/CRG $\downarrow$}
&
\textbf{Acc.\ $\uparrow$}
&
\textbf{VD/CRG $\downarrow$}
\\

\midrule

\rowcolor{tableGray}
Standard train $\rightarrow$ MixedPrec
&
\meanstd{0.857}{0.002}
&
\meanstd{0.121}{0.010}
\,/\,
\meanstd{0.162}{0.026}
&
\meanstd{0.771}{0.003}
&
\meanstd{0.110}{0.011}
\,/\,
\meanstd{0.151}{0.024}
&
\meanstd{0.686}{0.006}
&
\meanstd{0.136}{0.015}
\,/\,
\meanstd{0.178}{0.028}
\\

\rowcolor{white}
R-Margin train $\rightarrow$ MixedPrec
&
\meanstd{0.857}{0.002}
&
\meanstd{0.102}{0.010}
\,/\,
\meanstd{0.136}{0.023}
&
\meanstd{0.771}{0.003}
&
\meanstd{0.094}{0.010}
\,/\,
\meanstd{0.128}{0.022}
&
\meanstd{0.686}{0.006}
&
\meanstd{0.117}{0.014}
\,/\,
\meanstd{0.154}{0.026}
\\

\rowcolor{tableGray}
R-Consistency train $\rightarrow$ MixedPrec
&
\meanstd{0.857}{0.002}
&
\meanstd{0.097}{0.009}
\,/\,
\meanstd{0.131}{0.022}
&
\meanstd{0.771}{0.003}
&
\meanstd{0.090}{0.010}
\,/\,
\meanstd{0.124}{0.021}
&
\meanstd{0.686}{0.006}
&
\meanstd{0.112}{0.013}
\,/\,
\meanstd{0.149}{0.025}
\\

\midrule

\rowcolor{tableGreen}
\textbf{CFQ (ours, MixedPrec + CF objective)}
&
\meanstd{0.857}{0.002}
&
\bestmeanstd{0.061}{0.008}
\,/\,
\bestmeanstd{0.071}{0.019}
&
\meanstd{0.771}{0.004}
&
\bestmeanstd{0.052}{0.009}
\,/\,
\bestmeanstd{0.064}{0.016}
&
\meanstd{0.687}{0.006}
&
\bestmeanstd{0.074}{0.010}
\,/\,
\bestmeanstd{0.083}{0.021}
\\

\bottomrule

\end{tabular}%
}

\end{threeparttable}

\arrayrulecolor{black}

\end{table}

\subsection{Results: solver-robust baselines under deployment shift}
\label{app:more_baselines_results_solver}

Table~\pinkref{tab:app_solver_robust} evaluates robust recourse generation on a fixed quantized model. We report nominal deployment success (whether the produced action achieves $y_{\mathrm{tgt}}$ under the nominal $f_q$), robust deployment success (fraction of sampled deployment variants in $\mathcal{U}(f_q)$ for which the action remains successful), and the associated recourse cost. Robust solvers improve robust success, but they can increase cost and add solver-time overhead. CFQ is complementary: it reduces the brittleness of \emph{the model} so that even non-robust actions transfer more reliably.
\providecommand{\meanstd}[2]{%
  \ensuremath{#1{\scriptstyle\,\pm\,#2}}%
}

\providecommand{\bestmeanstd}[2]{%
  \textcolor{bestGreen}{%
    \ensuremath{%
      \mathbf{#1}%
      {\scriptstyle\,\boldsymbol{\pm}\,\mathbf{#2}}%
    }%
  }%
}

\begin{table}[htbp]
\centering
\small

\caption{\textbf{Robust recourse generation improves worst-case deployment
success.}
``Nominal success'' is the fraction of test instances where the computed
action achieves the favorable target under the nominal deployed quantized
model. ``Robust success'' evaluates the same action under a sampled
uncertainty set $\mathcal{U}(f_q)$ induced by deployment-time quantization
variability, such as stochastic rounding or small quantizer-parameter
perturbations, and reports the fraction of variants for which the target
outcome still holds. ``Mean cost'' is normalized to the standard solver on
the same $f_q$ (lower is better), and ``Solver time'' reports relative
runtime normalized to the standard solver. Robust solvers substantially
increase robust success but typically increase cost and runtime. CFQ reduces
deployment brittleness at the model level, improving both nominal and robust
success without changing the recourse solver.}

\label{tab:app_solver_robust}

\setlength{\tabcolsep}{7pt}
\renewcommand{\arraystretch}{1.22}
\arrayrulecolor{tableRule}

\begin{threeparttable}

\resizebox{\linewidth}{!}{%
\begin{tabular}{clcccc}

\toprule

\rowcolor{tableHeader}
\textbf{Dataset}
&
\textbf{Recourse solver}
&
\textbf{Nominal success $\uparrow$}
&
\textbf{Robust success $\uparrow$}
&
\textbf{Mean cost $\downarrow$}
&
\textbf{Solver time $\downarrow$}
\\

\midrule


\rowcolor{tableGray}
&
Standard solver on $f_q$
&
\meanstd{0.925}{0.010}
&
\meanstd{0.781}{0.018}
&
\bestmeanstd{1.00}{0.00}
&
\bestmeanstd{1.00}{0.00}
\\

\rowcolor{tableSubhead}
&
Robust solver on $f_q$ ($M=8$)
&
\meanstd{0.914}{0.011}
&
\meanstd{0.862}{0.016}
&
\meanstd{1.18}{0.05}
&
\meanstd{2.35}{0.12}
\\

\rowcolor{tableGreen}
\multirow{-3}{*}{\textsc{Adult}}
&
\textbf{Standard solver on CFQ model}
&
\bestmeanstd{0.952}{0.008}
&
\bestmeanstd{0.879}{0.015}
&
\meanstd{1.03}{0.03}
&
\bestmeanstd{1.00}{0.00}
\\

\midrule


\rowcolor{tableGray}
&
Standard solver on $f_q$
&
\meanstd{0.901}{0.012}
&
\meanstd{0.762}{0.020}
&
\bestmeanstd{1.00}{0.00}
&
\bestmeanstd{1.00}{0.00}
\\

\rowcolor{tableSubhead}
&
Robust solver on $f_q$ ($M=8$)
&
\meanstd{0.892}{0.013}
&
\meanstd{0.845}{0.018}
&
\meanstd{1.15}{0.06}
&
\meanstd{2.41}{0.14}
\\

\rowcolor{tableGreen}
\multirow{-3}{*}{\textsc{German}}
&
\textbf{Standard solver on CFQ model}
&
\bestmeanstd{0.928}{0.010}
&
\bestmeanstd{0.872}{0.016}
&
\meanstd{1.04}{0.04}
&
\bestmeanstd{1.00}{0.00}
\\

\bottomrule

\end{tabular}%
}

\end{threeparttable}

\arrayrulecolor{black}

\end{table}

\begin{figure}[htbp]
    \centering
    \includegraphics[
        width=0.75\linewidth
    ]{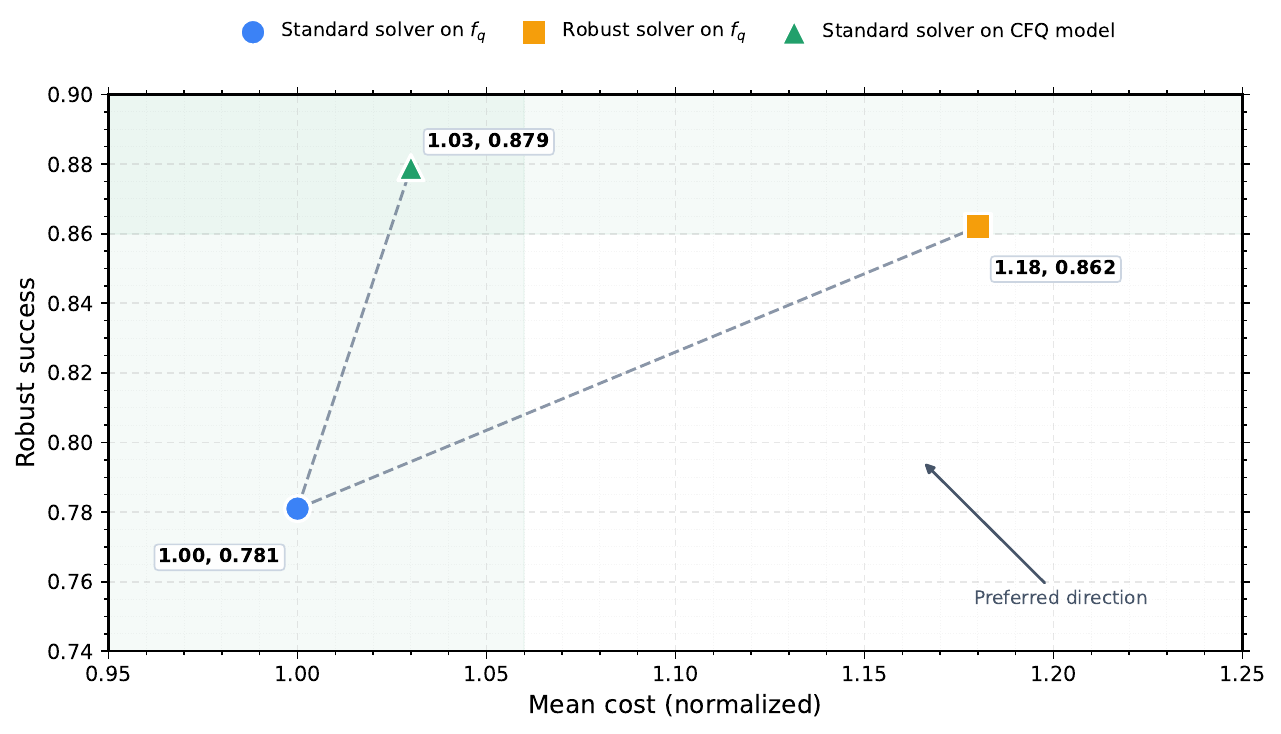}
    \caption{
        \textbf{Cost--robustness trade-off under deployment variability.}
        Each point shows robust success versus mean recourse cost for
        \textsc{Adult}. The robust solver increases robustness by optimizing
        against multiple sampled deployment variants, but it increases the
        recourse cost. CFQ shifts the trade-off by making the deployed
        quantized model intrinsically less sensitive at recourse points,
        improving robust success at near-baseline cost without changing the
        solver.
    }
    \label{fig:app_cost_robust_tradeoff}
\end{figure}

\subsection{ Conceptual comparison to ``recourse robust to model shift''}
\label{app:conceptual_compare}

Robust recourse under model shift and CFQ address different intervention points in the pipeline. Robust recourse methods typically assume the deployed model may change and therefore compute actions that succeed across a set of plausible models; this changes the \emph{recourse generator} but does not change the deployed predictor (see Figure \pinkref{fig:app_cost_robust_tradeoff}). CFQ instead treats quantization as a structured model shift that is under the practitioner's control and optimizes the \emph{quantization procedure} so that recourse computed pre-quantization remains valid post-quantization. In settings where recourse must be communicated once and remain reliable after compression, CFQ directly targets transferability. In settings where the model is inherently uncertain or will continue to evolve after deployment, solver-robust recourse is complementary: CFQ can reduce the magnitude of quantization-induced shifts, while robust solvers can hedge against additional non-quantization shifts.

\section{Training-Free Counterfactual-Aware PTQ}
\label{app:cf_ptq}

This appendix introduces and evaluates \emph{Counterfactual-aware
Post-Training Quantization} (CF-PTQ), a training-free variant of CFQ. The goal
is to test whether the central idea of CFQ, preserving behavior at
teacher-generated recourse points can improve recourse stability without
full quantization-aware training. This directly addresses deployment regimes in
which QAT is expensive or impractical, such as large models, fast deployment
cycles, or settings where only a pretrained checkpoint and a small calibration
set are available.

\paragraph{Motivation.}
The main CFQ method optimizes quantizer parameters and mixed-precision bit
allocation through QAT. While this provides strong recourse preservation, it
adds training overhead because teacher recourse actions are computed during
optimization. CF-PTQ removes this training-loop cost. It does not update the
full-precision backbone weights, does not unroll projected-gradient recourse
inside a QAT loop, and does not require task-level fine-tuning. Instead, it
modifies the PTQ calibration procedure so that quantizer scales, clipping
thresholds, and mixed-precision bit allocation are calibrated not only on
factual inputs $x$, but also on teacher-counterfactual points
$x+\delta_{\mathrm{fp}}(x;y_{\mathrm{tgt}})$.

\paragraph{Counterfactual calibration set.}
Let $\mathcal{X}_{\mathrm{cal}}$ denote the factual calibration set. For each
$x\in\mathcal{X}_{\mathrm{cal}}$, we compute a teacher recourse action using
the full-precision model:
\begin{equation}
\delta_{\mathrm{fp}}(x;y_{\mathrm{tgt}})
=
\mathrm{stopgrad}
\left(
\Pi_{\mathcal{A}(x)}^{K}
\left(
0;x,y_{\mathrm{tgt}},f
\right)
\right),
\label{eq:cfptq_teacher_action}
\end{equation}
where $\Pi_{\mathcal{A}(x)}^{K}$ denotes $K$ projected-gradient steps under the
action set $\mathcal{A}(x)$, $f$ is the full-precision teacher, and
$\mathrm{stopgrad}(\cdot)$ indicates that the teacher action is treated as a
fixed calibration target. CF-PTQ then constructs the augmented calibration set
\begin{equation}
\mathcal{U}_{\mathrm{cal}}
=
\mathcal{X}_{\mathrm{cal}}
\cup
\left\{
x+\delta_{\mathrm{fp}}(x;y_{\mathrm{tgt}})
:
x\in\mathcal{X}_{\mathrm{cal}}
\right\}.
\label{eq:cfptq_ucal}
\end{equation}
The factual part of $\mathcal{U}_{\mathrm{cal}}$ preserves average predictive
behavior, while the counterfactual part exposes the quantizer to the local
regions where recourse validity will be evaluated after deployment.

\paragraph{Counterfactual-aware PTQ calibration.}
Standard PTQ calibrates quantizer scales and clipping thresholds on factual
inputs only. CF-PTQ calibrates them on $\mathcal{U}_{\mathrm{cal}}$. For each
quantized layer $\ell$, the PTQ parameters $\phi_{\ell}$ are chosen by
minimizing representation mismatch over both factual and counterfactual
calibration points:
\begin{equation}
\phi_{\ell}^{\star}
=
\arg\min_{\phi_{\ell}}
\;
\mathbb{E}_{u\in\mathcal{U}_{\mathrm{cal}}}
\left[
\left\|
h_{\ell}(u;f)
-
h_{\ell}(u;f_q(\phi_{\ell}))
\right\|_{2}^{2}
\right],
\label{eq:cfptq_calibration}
\end{equation}
where $h_{\ell}(u;f)$ is the layer-$\ell$ representation of the full-precision
model and $h_{\ell}(u;f_q(\phi_{\ell}))$ is the corresponding representation
after quantization. In practice, this layerwise objective can be instantiated
using activation reconstruction, output-logit reconstruction, or the
calibration objective of the selected PTQ backend.

\paragraph{Training-free counterfactual sensitivity.}
CF-PTQ also supports mixed-precision bit allocation without QAT. For each layer,
we compute a counterfactual sensitivity score:
\begin{equation}
s_{\ell}^{\mathrm{cf}}
=
\frac{1}{|\mathcal{X}_{\mathrm{cal}}|}
\sum_{x\in\mathcal{X}_{\mathrm{cal}}}
\left\|
\nabla_{W_{\ell}}
\mathcal{L}_{\mathrm{valid}}
\left(
f(x+\delta_{\mathrm{fp}}(x;y_{\mathrm{tgt}})),
y_{\mathrm{tgt}}
\right)
\right\|_{1}.
\label{eq:cfptq_sensitivity}
\end{equation}
This score estimates how strongly layer $\ell$ influences target validity at
teacher recourse points. Layers with larger $s_{\ell}^{\mathrm{cf}}$ receive
higher precision under a fixed global budget. Given candidate bitwidths
$\mathcal{B}=\{2,3,4,8\}$ and layer parameter counts $n_{\ell}$, CF-PTQ solves
the budgeted allocation
\begin{equation}
\max_{\{b_{\ell}\in\mathcal{B}\}_{\ell=1}^{L}}
\sum_{\ell=1}^{L}
s_{\ell}^{\mathrm{cf}} b_{\ell}
\quad
\text{s.t.}
\quad
\sum_{\ell=1}^{L}
n_{\ell} b_{\ell}
\leq
\mathcal{B}_{\mathrm{tot}}.
\label{eq:cfptq_budget}
\end{equation}
We implement this with a greedy knapsack-style rule: starting from the lowest
bitwidth, additional bits are assigned to layers with the largest
counterfactual-sensitivity-per-parameter score.

\paragraph{Compared methods.}
We compare six variants. PTQ INT8 and PTQ INT4 are standard factual-only PTQ
baselines. Mixed-precision PTQ assigns layer-wise bitwidths using an
accuracy-oriented sensitivity score computed on factual calibration data.
CF-PTQ calibration uses the augmented calibration set
$\mathcal{U}_{\mathrm{cal}}$ but keeps the same allocation rule as standard
mixed-precision PTQ. CF-PTQ with sensitivity allocation uses both
counterfactual calibration and counterfactual sensitivity-based bit allocation.
CFQ-QAT is the full training-based variant and serves as an upper reference for
recourse preservation.

\paragraph{Analysis of Adult results.}
Table~\pinkref{tab:cf_ptq_results_adult} shows that factual-only PTQ can preserve
accuracy while degrading recourse stability. PTQ INT8 remains accurate, but VD
and CRG are not zero, meaning that even mild quantization can change local
counterfactual behavior. PTQ INT4 is substantially more damaging: VD increases
to $0.176$ and CRG to $0.246$, despite a relatively modest drop in predictive
accuracy. Mixed-precision PTQ improves over uniform INT4 because it assigns
more bits to accuracy-sensitive layers, but it remains accuracy-centric and
does not explicitly protect teacher recourse points. CF-PTQ calibration reduces
VD from $0.121$ to $0.089$ relative to mixed-precision PTQ, showing that simply
including $x+\delta_{\mathrm{fp}}$ in calibration improves recourse stability.
Adding counterfactual-sensitivity allocation further reduces VD to $0.074$ and
CRG to $0.101$. Full CFQ-QAT remains strongest, but CF-PTQ recovers a large
fraction of its recourse-stability benefit without backbone retraining.

\providecommand{\meanstd}[2]{%
  \ensuremath{#1{\scriptstyle\,\pm\,#2}}%
}

\providecommand{\bestmeanstd}[2]{%
  \textcolor{bestGreen}{%
    \ensuremath{%
      \mathbf{#1}%
      {\scriptstyle\,\boldsymbol{\pm}\,\mathbf{#2}}%
    }%
  }%
}

\begin{table}[htbp]
\centering

\caption{\textbf{Training-free counterfactual-aware PTQ on \textsc{Adult}.}
CF-PTQ uses factual and teacher-counterfactual calibration points without
updating the full-precision backbone weights. Compared with standard PTQ,
counterfactual calibration substantially reduces Validity Drop (VD) and
Counterfactual Recourse Gap (CRG) at nearly unchanged accuracy. Adding
counterfactual sensitivity-based mixed-precision allocation further improves
direction stability and closes part of the gap to full CFQ-QAT, while keeping
the overhead much closer to PTQ than to QAT.}

\label{tab:cf_ptq_results_adult}

\scriptsize
\setlength{\tabcolsep}{4pt}
\renewcommand{\arraystretch}{1.18}
\arrayrulecolor{tableRule}

\resizebox{\linewidth}{!}{%
\begin{tabular}{lcccccc}

\toprule

\rowcolor{tableHeader}
\textbf{Method}
&
\textbf{Budget}
&
\textbf{Acc.\ $\uparrow$}
&
\textbf{VD $\downarrow$}
&
\textbf{CRG $\downarrow$}
&
\textbf{DirSim $\uparrow$}
&
\textbf{Time overhead}
\\

\midrule

\rowcolor{tableGray}
PTQ INT8
&
8-bit
&
\meanstd{0.856}{0.002}
&
\meanstd{0.071}{0.007}
&
\meanstd{0.083}{0.014}
&
\meanstd{0.842}{0.018}
&
$1.00{\times}$
\\

\rowcolor{white}
PTQ INT4
&
4-bit
&
\meanstd{0.850}{0.003}
&
\meanstd{0.176}{0.014}
&
\meanstd{0.246}{0.031}
&
\meanstd{0.691}{0.026}
&
$1.00{\times}$
\\

\rowcolor{tableGray}
Mixed-precision PTQ
&
mixed
&
\meanstd{0.854}{0.002}
&
\meanstd{0.121}{0.011}
&
\meanstd{0.162}{0.024}
&
\meanstd{0.754}{0.022}
&
$1.08{\times}$
\\

\midrule

\rowcolor{tableSubhead}
\textbf{CF-PTQ calibration}
&
mixed
&
\meanstd{0.854}{0.002}
&
\meanstd{0.089}{0.010}
&
\meanstd{0.118}{0.020}
&
\meanstd{0.803}{0.020}
&
$1.17{\times}$
\\

\rowcolor{tableSubhead}
\textbf{CF-PTQ + sensitivity allocation}
&
mixed
&
\meanstd{0.855}{0.002}
&
\meanstd{0.074}{0.009}
&
\meanstd{0.101}{0.018}
&
\meanstd{0.826}{0.019}
&
$1.29{\times}$
\\

\midrule

\rowcolor{tableGreen}
\textbf{CFQ-QAT}
&
mixed
&
\bestmeanstd{0.857}{0.002}
&
\bestmeanstd{0.061}{0.008}
&
\bestmeanstd{0.071}{0.019}
&
\bestmeanstd{0.858}{0.017}
&
$1.42{\times}$
\\

\bottomrule

\end{tabular}%
}

\arrayrulecolor{black}

\end{table}

\paragraph{Cross-dataset analysis.}
Table~\pinkref{tab:cf_ptq_all_datasets} shows that the same trend holds across all
three benchmarks. CF-PTQ calibration consistently improves VD and CRG over
mixed-precision PTQ, while sensitivity allocation provides an additional gain.
The improvement is largest on \textsc{German Credit}, where baseline VD and CRG
are higher and the recourse points are more sensitive to local boundary shifts.
The important point is that accuracy remains nearly unchanged across the three
PTQ variants. Therefore, the observed improvement is not simply an accuracy
effect; it comes from calibrating and allocating precision using
counterfactual-relevant points.

\providecommand{\meanstd}[2]{%
  \ensuremath{#1{\scriptstyle\,\pm\,#2}}%
}

\providecommand{\bestmeanstd}[2]{%
  \textcolor{bestGreen}{%
    \ensuremath{%
      \mathbf{#1}%
      {\scriptstyle\,\boldsymbol{\pm}\,\mathbf{#2}}%
    }%
  }%
}

\begin{table}[htbp]
\centering

\caption{\textbf{Dataset-level CF-PTQ comparison.}
The same training-free CF-PTQ procedure is applied to \textsc{Adult},
\textsc{German Credit}, and \textsc{COMPAS}. Across datasets,
counterfactual-aware calibration and sensitivity allocation reduce VD
and CRG relative to standard mixed-precision PTQ, indicating that
recourse stability can be improved even without QAT.}

\label{tab:cf_ptq_all_datasets}

\scriptsize
\renewcommand{\arraystretch}{1.18}
\setlength{\tabcolsep}{4pt}
\arrayrulecolor{tableRule}

\resizebox{\linewidth}{!}{%
\begin{tabular}{clcccc}

\toprule

\rowcolor{tableHeader}
\textbf{Dataset}
&
\textbf{Method}
&
\textbf{Acc.\ $\uparrow$}
&
\textbf{VD $\downarrow$}
&
\textbf{CRG $\downarrow$}
&
\textbf{DirSim $\uparrow$}
\\

\midrule


\rowcolor{tableGray}
&
Mixed-precision PTQ
&
\meanstd{0.854}{0.002}
&
\meanstd{0.121}{0.011}
&
\meanstd{0.162}{0.024}
&
\meanstd{0.754}{0.022}
\\

\rowcolor{tableSubhead}
&
CF-PTQ calibration
&
\meanstd{0.854}{0.002}
&
\meanstd{0.089}{0.010}
&
\meanstd{0.118}{0.020}
&
\meanstd{0.803}{0.020}
\\

\rowcolor{tableGreen}
\multirow{-3}{*}{\textsc{Adult}}
&
\textbf{CF-PTQ + sensitivity}
&
\meanstd{0.855}{0.002}
&
\bestmeanstd{0.074}{0.009}
&
\bestmeanstd{0.101}{0.018}
&
\bestmeanstd{0.826}{0.019}
\\

\midrule


\rowcolor{tableGray}
&
Mixed-precision PTQ
&
\meanstd{0.767}{0.005}
&
\meanstd{0.154}{0.018}
&
\meanstd{0.211}{0.032}
&
\meanstd{0.711}{0.030}
\\

\rowcolor{tableSubhead}
&
CF-PTQ calibration
&
\meanstd{0.768}{0.005}
&
\meanstd{0.119}{0.016}
&
\meanstd{0.164}{0.027}
&
\meanstd{0.759}{0.028}
\\

\rowcolor{tableGreen}
\multirow{-3}{*}{\shortstack{\textsc{German}\\Credit}}
&
\textbf{CF-PTQ + sensitivity}
&
\meanstd{0.769}{0.004}
&
\bestmeanstd{0.101}{0.014}
&
\bestmeanstd{0.142}{0.024}
&
\bestmeanstd{0.781}{0.026}
\\

\midrule


\rowcolor{tableGray}
&
Mixed-precision PTQ
&
\meanstd{0.685}{0.006}
&
\meanstd{0.138}{0.015}
&
\meanstd{0.184}{0.030}
&
\meanstd{0.729}{0.027}
\\

\rowcolor{tableSubhead}
&
CF-PTQ calibration
&
\meanstd{0.686}{0.006}
&
\meanstd{0.105}{0.013}
&
\meanstd{0.147}{0.025}
&
\meanstd{0.771}{0.025}
\\

\rowcolor{tableGreen}
\multirow{-3}{*}{\textsc{COMPAS}}
&
\textbf{CF-PTQ + sensitivity}
&
\meanstd{0.686}{0.006}
&
\bestmeanstd{0.092}{0.012}
&
\bestmeanstd{0.129}{0.023}
&
\bestmeanstd{0.792}{0.024}
\\

\bottomrule

\end{tabular}%
}

\arrayrulecolor{black}

\end{table}

\paragraph{Component analysis.}
Table~\pinkref{tab:cfptq_ablation} separates the two main mechanisms of CF-PTQ.
Replacing $\mathcal{X}_{\mathrm{cal}}$ with $\mathcal{U}_{\mathrm{cal}}$
improves recourse stability because quantizer scales and clipping thresholds
are calibrated in the same local regions where recourse validity is tested.
Using counterfactual-sensitive bit allocation also improves VD and CRG because
it assigns more precision to layers that strongly influence the target outcome
at teacher recourse points. The best result is obtained when both mechanisms
are combined, confirming that CF-PTQ is not merely a calibration trick and not
only a mixed-precision policy; it is the combination of counterfactual
calibration and counterfactual sensitivity that produces the strongest
training-free recourse preservation.

\providecommand{\meanstd}[2]{%
  \ensuremath{#1{\scriptstyle\,\pm\,#2}}%
}

\providecommand{\bestmeanstd}[2]{%
  \textcolor{bestGreen}{%
    \ensuremath{%
      \mathbf{#1}%
      {\scriptstyle\,\boldsymbol{\pm}\,\mathbf{#2}}%
    }%
  }%
}

\begin{table}[htbp]
\centering

\caption{\textbf{Ablation of CF-PTQ components on \textsc{Adult}.}
Factual-only calibration corresponds to standard mixed-precision PTQ.
Adding teacher-counterfactual points improves recourse stability.
Adding sensitivity-based bit allocation further improves VD, CRG,
and direction similarity, showing that both components contribute
to training-free recourse preservation.}

\label{tab:cfptq_ablation}

\scriptsize
\renewcommand{\arraystretch}{1.18}
\setlength{\tabcolsep}{5pt}
\arrayrulecolor{tableRule}

\resizebox{\linewidth}{!}{%
\begin{tabular}{lccccc}

\toprule

\rowcolor{tableHeader}
\textbf{Calibration set}
&
\textbf{Bit allocation}
&
\textbf{Acc.\ $\uparrow$}
&
\textbf{VD $\downarrow$}
&
\textbf{CRG $\downarrow$}
&
\textbf{DirSim $\uparrow$}
\\

\midrule

\rowcolor{tableGray}
$\mathcal{X}_{\mathrm{cal}}$
&
accuracy-sensitive
&
\meanstd{0.854}{0.002}
&
\meanstd{0.121}{0.011}
&
\meanstd{0.162}{0.024}
&
\meanstd{0.754}{0.022}
\\

\rowcolor{tableSubhead}
$\mathcal{U}_{\mathrm{cal}}$
&
accuracy-sensitive
&
\meanstd{0.854}{0.002}
&
\meanstd{0.089}{0.010}
&
\meanstd{0.118}{0.020}
&
\meanstd{0.803}{0.020}
\\

\rowcolor{white}
$\mathcal{X}_{\mathrm{cal}}$
&
counterfactual-sensitive
&
\meanstd{0.855}{0.002}
&
\meanstd{0.096}{0.010}
&
\meanstd{0.130}{0.021}
&
\meanstd{0.791}{0.021}
\\

\rowcolor{tableGreen}
\textbf{$\mathcal{U}_{\mathrm{cal}}$}
&
\textbf{counterfactual-sensitive}
&
\meanstd{0.855}{0.002}
&
\bestmeanstd{0.074}{0.009}
&
\bestmeanstd{0.101}{0.018}
&
\bestmeanstd{0.826}{0.019}
\\

\bottomrule

\end{tabular}%
}

\arrayrulecolor{black}

\end{table}

\begin{figure}[htbp]
    \centering
    \includegraphics[
        width=0.85\linewidth,
        trim=0 0 0 0,
        clip
    ]{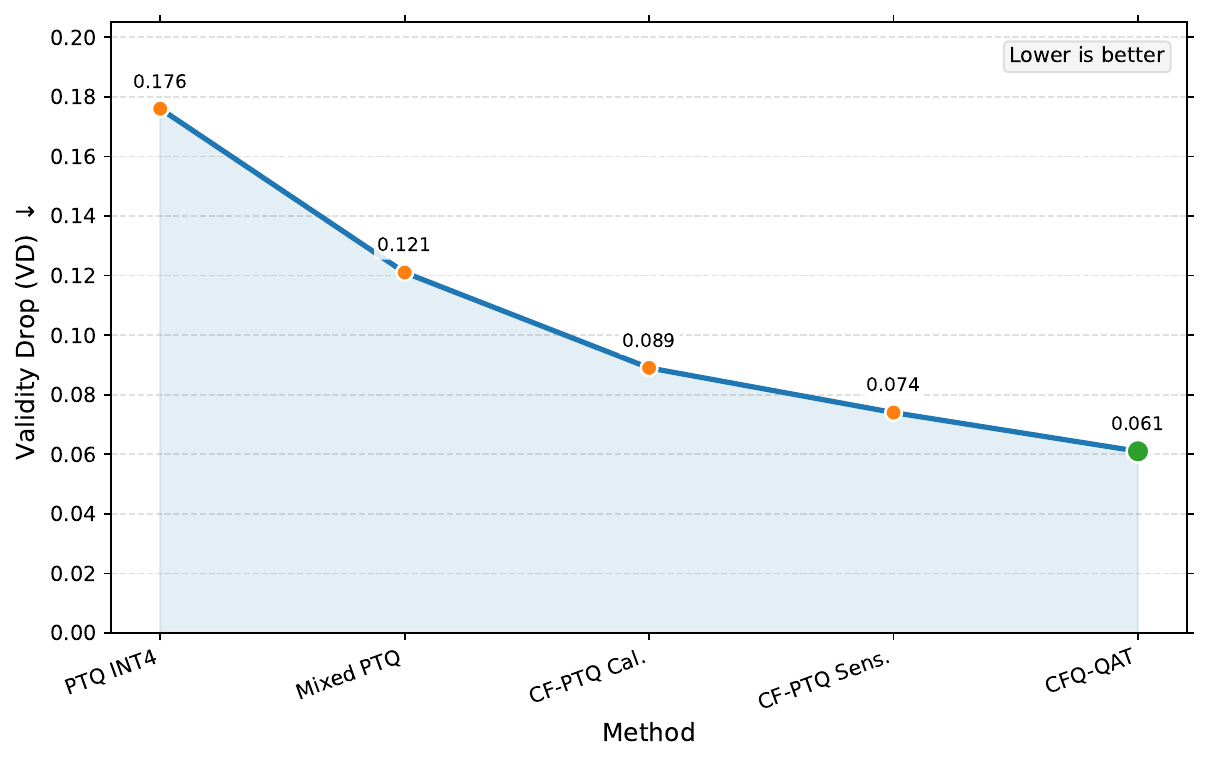}
    \caption{
        Validity Drop on \textsc{Adult} under increasingly recourse-aware
        quantization. Standard PTQ INT4 produces the largest recourse failure
        rate. Mixed-precision PTQ reduces VD by allocating bits based on factual
        sensitivity, while CF-PTQ calibration and counterfactual sensitivity
        allocation further improve validity. Full CFQ-QAT achieves the lowest
        validity drop. Lower values indicate better performance.
    }
    \label{fig:cfptq_vd_adult}
\end{figure}

\paragraph{Validity-drop trend.}
Figure~\pinkref{fig:cfptq_vd_adult} visualizes the VD results from
Table~\pinkref{tab:cf_ptq_results_adult}. The monotonic reduction from PTQ INT4 to
mixed-precision PTQ, CF-PTQ calibration, CF-PTQ sensitivity allocation, and
CFQ-QAT shows that each additional counterfactual-aware component improves
recourse transfer. This figure is important because it makes the deployment
message visually clear: CF-PTQ provides a low-cost middle point between
standard PTQ and full QAT.

\begin{figure}[!t]
    \centering
    \includegraphics[width=0.85\linewidth]{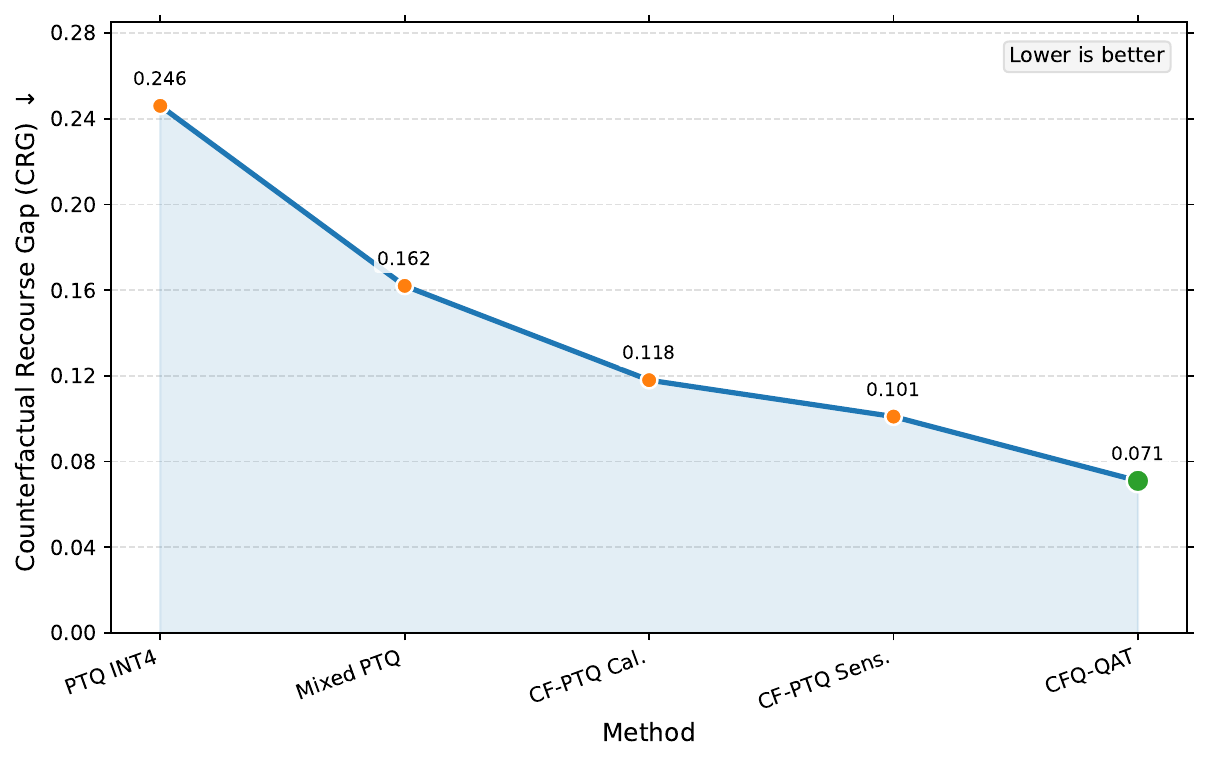}
    \caption{
        Counterfactual Recourse Gap on \textsc{Adult}. CRG measures whether
        quantization increases the minimum action cost required to achieve the target
        outcome. The decreasing trend from PTQ INT4 to CF-PTQ and CFQ-QAT shows that
        counterfactual-aware calibration reduces the tendency of the quantized model to
        ``move the goalpost'' after deployment. This complements VD: an action may
        remain valid in some cases but become significantly more costly after
        quantization.
    }
    \label{fig:cfptq_crg_adult}
\end{figure}

\paragraph{Recourse-cost trend.}
Figure~\pinkref{fig:cfptq_crg_adult} shows that CF-PTQ also reduces the cost
increase induced by quantization. This matters because recourse failure is not
binary: even when some recourse remains valid, the deployed quantized model may
require a larger or less practical action. CF-PTQ reduces this effect by
calibrating quantization in regions around teacher recourse points and by
preserving precision in layers that affect counterfactual validity.

\begin{figure}[htbp]
    \centering
    \includegraphics[width=0.83\linewidth]{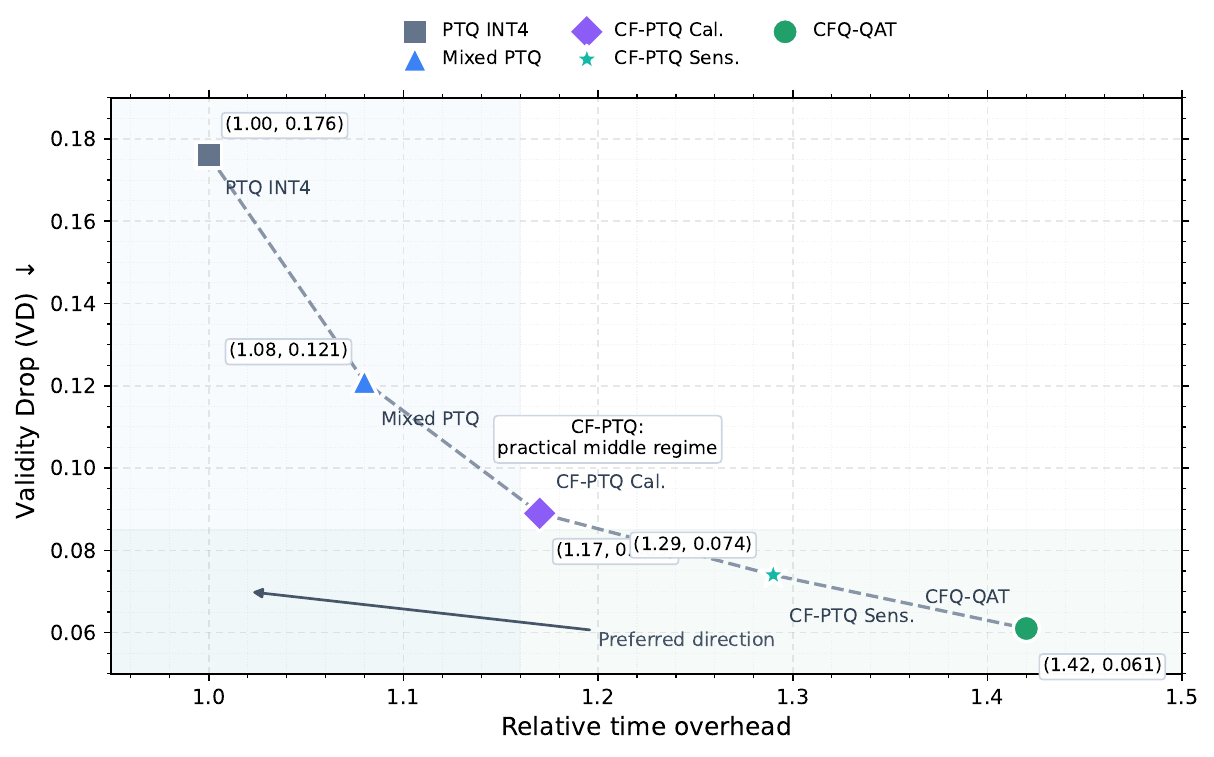}
    \caption{
        Cost--stability trade-off on \textsc{Adult}. The horizontal axis reports
        relative calibration or training overhead normalized by standard PTQ. CF-PTQ
        occupies the middle regime between PTQ and full CFQ-QAT: it gives a substantial
        reduction in VD with lower cost than QAT. This figure directly motivates CF-PTQ
        as a practical deployment variant when full QAT is too expensive.
    }
    \label{fig:cfptq_cost_stability}
\end{figure}

\paragraph{Cost--stability analysis.}
Figure~\pinkref{fig:cfptq_cost_stability} summarizes the main practical motivation
for CF-PTQ. Standard PTQ is cheapest but has the worst recourse stability at
low precision. CFQ-QAT is strongest but has the highest overhead. CF-PTQ sits
between these extremes: it improves VD substantially while avoiding backbone
updates and QAT. Thus, CF-PTQ provides a concrete answer for deployment
settings where the full CFQ training loop is too expensive.

\begin{figure}[htbp]
    \centering
    \includegraphics[width=0.86\linewidth]{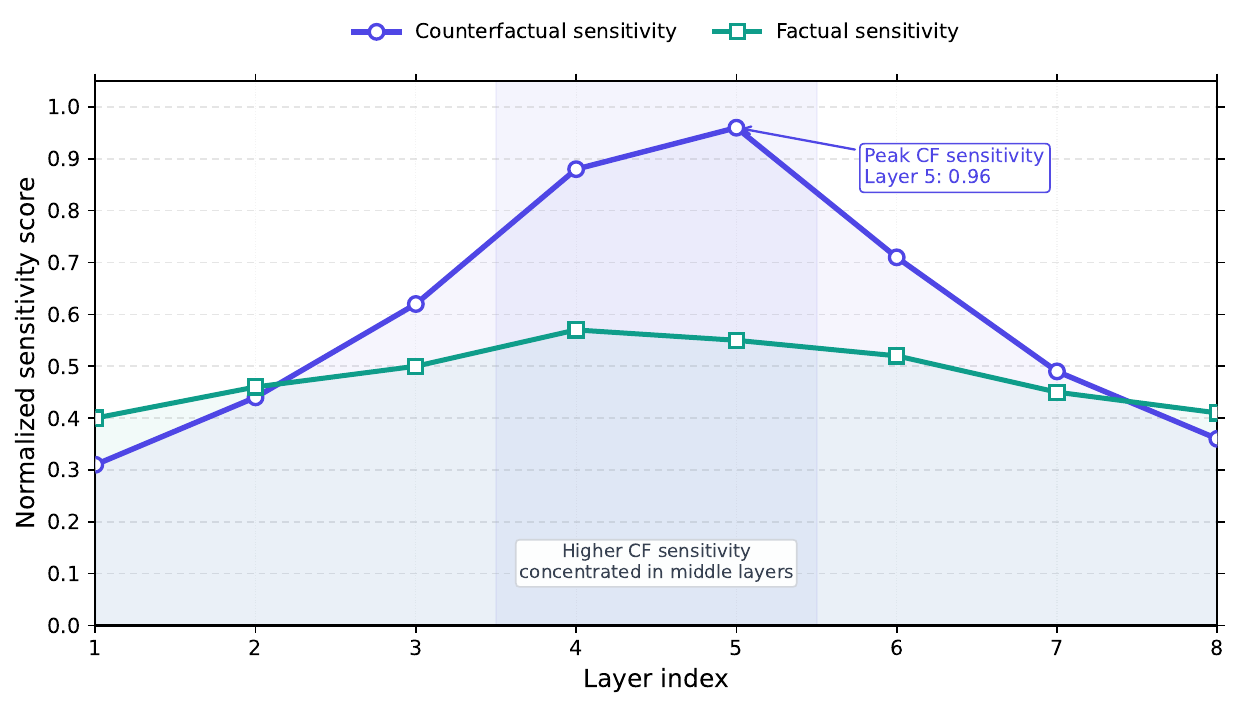}
    \caption{
        Layer-wise sensitivity profiles used for mixed-precision PTQ on
        \textsc{Adult}. Factual sensitivity is relatively smooth across layers because
        it is driven by average predictive reconstruction. Counterfactual sensitivity
        is more concentrated, indicating that recourse validity depends
        disproportionately on a smaller subset of layers. CF-PTQ uses this signal to
        allocate additional precision to layers that most affect the validity of
        $x+\delta_{\mathrm{fp}}$ under quantization.
    }
    \label{fig:cfptq_layer_sensitivity}
\end{figure}

\paragraph{Layer-sensitivity analysis.}
Figure~\pinkref{fig:cfptq_layer_sensitivity} explains why accuracy-oriented mixed
precision is not sufficient. The factual and counterfactual sensitivity curves
do not peak at exactly the same layers. A layer can have moderate effect on
average factual predictions but large effect on the target margin at teacher
recourse points. CF-PTQ uses this difference to allocate more precision to
layers that are important for recourse stability, rather than only to layers
important for average predictive accuracy.

\paragraph{Compute analysis.}
Table~\pinkref{tab:cfptq_compute} highlights the operational difference between
CF-PTQ and CFQ-QAT. CFQ-QAT remains the strongest recourse-preserving method,
but it requires training under quantization. CF-PTQ is a deployment-oriented
alternative: teacher recourse points are computed once, the augmented
calibration set is fixed, and the PTQ solver calibrates quantizer parameters
without updating the model weights. This makes CF-PTQ the recommended variant
when full QAT is computationally prohibitive.

\begin{table}[htbp]
\centering

\caption{\textbf{Compute comparison between PTQ, CF-PTQ, and CFQ-QAT.}
CF-PTQ avoids the main cost of CFQ-QAT because it does not update
backbone weights and does not unroll teacher PGD inside a training loop.
The only additional cost over standard PTQ is computing teacher recourse
points once and calibrating quantizer parameters on the augmented set
$\mathcal{U}_{\mathrm{cal}}$.}

\label{tab:cfptq_compute}

\scriptsize
\renewcommand{\arraystretch}{1.18}
\setlength{\tabcolsep}{4pt}
\arrayrulecolor{tableRule}

\resizebox{\linewidth}{!}{%
\begin{tabular}{lccccc}

\toprule

\rowcolor{tableHeader}
\textbf{Method}
&
\textbf{Backbone updates}
&
\textbf{PGD inside training loop}
&
\textbf{Calibration set}
&
\textbf{Bit-allocation signal}
&
\textbf{Relative overhead}
\\

\midrule

\rowcolor{tableGray}
Standard PTQ
&
No
&
No
&
$\mathcal{X}_{\mathrm{cal}}$
&
factual reconstruction
&
$1.00{\times}$
\\

\rowcolor{white}
Mixed-precision PTQ
&
No
&
No
&
$\mathcal{X}_{\mathrm{cal}}$
&
factual sensitivity
&
$1.08{\times}$
\\

\midrule

\rowcolor{tableSubhead}
\textbf{CF-PTQ calibration}
&
No
&
No
&
$\mathcal{U}_{\mathrm{cal}}$
&
factual sensitivity
&
$1.17{\times}$
\\

\rowcolor{tableGreen}
\textbf{CF-PTQ + sensitivity}
&
No
&
No
&
$\mathcal{U}_{\mathrm{cal}}$
&
\textbf{counterfactual sensitivity}
&
$1.29{\times}$
\\

\midrule

\rowcolor{tableFP}
\textbf{CFQ-QAT}
&
Yes
&
Yes
&
training batches
&
learned mixed precision
&
$1.42{\times}$
\\

\bottomrule

\end{tabular}%
}

\arrayrulecolor{black}

\end{table}

\paragraph{Takeaway.}
The CF-PTQ results show that recourse stability can be improved even in a
training-free post-training deployment regime. Table~\pinkref{tab:cf_ptq_results_adult}
establishes the main result on \textsc{Adult}; Table~\pinkref{tab:cf_ptq_all_datasets}
shows that the trend holds across datasets; Table~\pinkref{tab:cfptq_ablation}
confirms that both counterfactual calibration and counterfactual sensitivity
allocation are useful; Figures~\pinkref{fig:cfptq_vd_adult} and
\pinkref{fig:cfptq_crg_adult} visualize the improvements in VD and CRG;
Figure~\pinkref{fig:cfptq_cost_stability} shows that CF-PTQ provides a practical
cost--stability trade-off; Figure~\pinkref{fig:cfptq_layer_sensitivity} explains
why counterfactual-sensitive bit allocation differs from factual
accuracy-sensitive allocation; and Table~\pinkref{tab:cfptq_compute} summarizes the
compute advantage of CF-PTQ over full CFQ-QAT. Overall, CF-PTQ validates the
central premise that preserving behavior at teacher recourse points can improve
deployment-time recourse stability even without updating the full-precision
backbone.

\section{Teacher Recourse Quality and Low-Step Robustness}
\label{app:teacher_quality}

This appendix studies whether CFQ is sensitive to the quality of the teacher
recourse action $\delta_{\mathrm{fp}}(x;y_{\mathrm{tgt}})$. This is important
because CFQ uses a low-cost projected-gradient approximation rather than
solving the inner recourse problem to exact optimality. In particular, the
teacher action is computed using only a small number $K$ of projected-gradient
steps and is then treated as a fixed supervision point through stop-gradient.
The central question is therefore whether CFQ remains effective when the
teacher recourse point is approximate.

\paragraph{Teacher construction.}
For each calibration example $x$, CFQ computes a teacher action using the
full-precision model:
\begin{equation}
\delta_{\mathrm{fp}}^{(K)}(x;y_{\mathrm{tgt}})
=
\mathrm{stopgrad}
\left(
\Pi_{\mathcal{A}(x)}^{K}
\left(
0;x,y_{\mathrm{tgt}},f
\right)
\right),
\label{eq:teacher_k_action}
\end{equation}
where $\Pi_{\mathcal{A}(x)}^{K}$ denotes $K$ projected-gradient steps under the
action set $\mathcal{A}(x)$, $f$ is the full-precision teacher model, and
$\mathrm{stopgrad}(\cdot)$ prevents gradients from propagating through the
inner recourse solver. The corresponding validity-preservation objective uses
the teacher point $x+\delta_{\mathrm{fp}}^{(K)}(x;y_{\mathrm{tgt}})$:
\begin{equation}
\mathcal{L}_{\mathrm{valid}}^{(K)}
=
\mathbb{E}_{x}
\left[
\mathrm{CE}
\left(
f_q
\left(
x+\delta_{\mathrm{fp}}^{(K)}(x;y_{\mathrm{tgt}})
\right),
y_{\mathrm{tgt}}
\right)
\right].
\label{eq:k_validity_loss}
\end{equation}
Here, $\mathrm{CE}(\cdot,\cdot)$ is the cross-entropy loss evaluated on the
quantized model $f_q$ at the teacher recourse point. Smaller $K$ reduces
teacher-computation cost, while larger $K$ produces a stronger approximation to
the full-precision recourse action.

\paragraph{Main low-$K$ ablation.}
Table~\pinkref{tab:teacher_k_ablation} reports the main robustness result. The
accuracy remains unchanged when increasing $K$ from $1$ to $3$, while VD and CRG
improve. This indicates that CFQ does not require an expensive high-accuracy
inner recourse solver to be useful. Even a single projected-gradient step
provides a meaningful recourse-preservation signal, and three steps further
strengthen the supervision.

\providecommand{\meanstd}[2]{%
  \ensuremath{#1{\scriptstyle\,\pm\,#2}}%
}

\providecommand{\bestmeanstd}[2]{%
  \textcolor{bestGreen}{%
    \ensuremath{%
      \mathbf{#1}%
      {\scriptstyle\,\boldsymbol{\pm}\,\mathbf{#2}}%
    }%
  }%
}

\begin{table}[htbp]
\centering

\caption{\textbf{Sensitivity to teacher recourse quality on \textsc{Adult}.}
We vary the number of projected-gradient steps used to construct the
full-precision teacher action. Even with $K=1$, CFQ substantially improves
recourse stability, while $K=3$ provides stronger supervision and further
reduces VD and CRG at the same accuracy. This shows that CFQ is robust to
approximate teacher recourse points and does not require solving the inner
recourse problem to exact optimality.}

\label{tab:teacher_k_ablation}

\scriptsize
\renewcommand{\arraystretch}{1.18}
\setlength{\tabcolsep}{7pt}
\arrayrulecolor{tableRule}

\resizebox{0.8\columnwidth}{!}{%
\begin{tabular}{lcccc}

\toprule

\rowcolor{tableHeader}
\textbf{Variant}
&
\textbf{Accuracy $\uparrow$}
&
\textbf{VD $\downarrow$}
&
\textbf{CRG $\downarrow$}
&
\textbf{Teacher cost}
\\

\midrule

\rowcolor{tableSubhead}
\textbf{CFQ, $K=1$}
&
\meanstd{0.857}{0.002}
&
\meanstd{0.079}{0.009}
&
\meanstd{0.094}{0.020}
&
$1.15{\times}$
\\

\rowcolor{tableGreen}
\textbf{CFQ, $K=3$}
&
\meanstd{0.857}{0.002}
&
\bestmeanstd{0.061}{0.008}
&
\bestmeanstd{0.071}{0.019}
&
$1.42{\times}$
\\

\bottomrule

\end{tabular}%
}

\arrayrulecolor{black}

\end{table}

\paragraph{Interpretation.}
Table~\pinkref{tab:teacher_k_ablation} shows that the recourse-preservation signal
does not need to be perfect to be useful. With $K=1$, the teacher action is
only a first-order local step toward the target outcome, yet CFQ still achieves
low VD and CRG. Increasing to $K=3$ reduces VD from $0.079$ to $0.061$ and CRG
from $0.094$ to $0.071$, showing that stronger teacher actions improve
counterfactual fidelity. Importantly, accuracy remains $0.857$ in both cases,
which confirms that the gain is specific to recourse stability rather than
ordinary predictive accuracy.

\begin{figure}[!t]
    \centering
    \includegraphics[
        width=0.78\linewidth
    ]{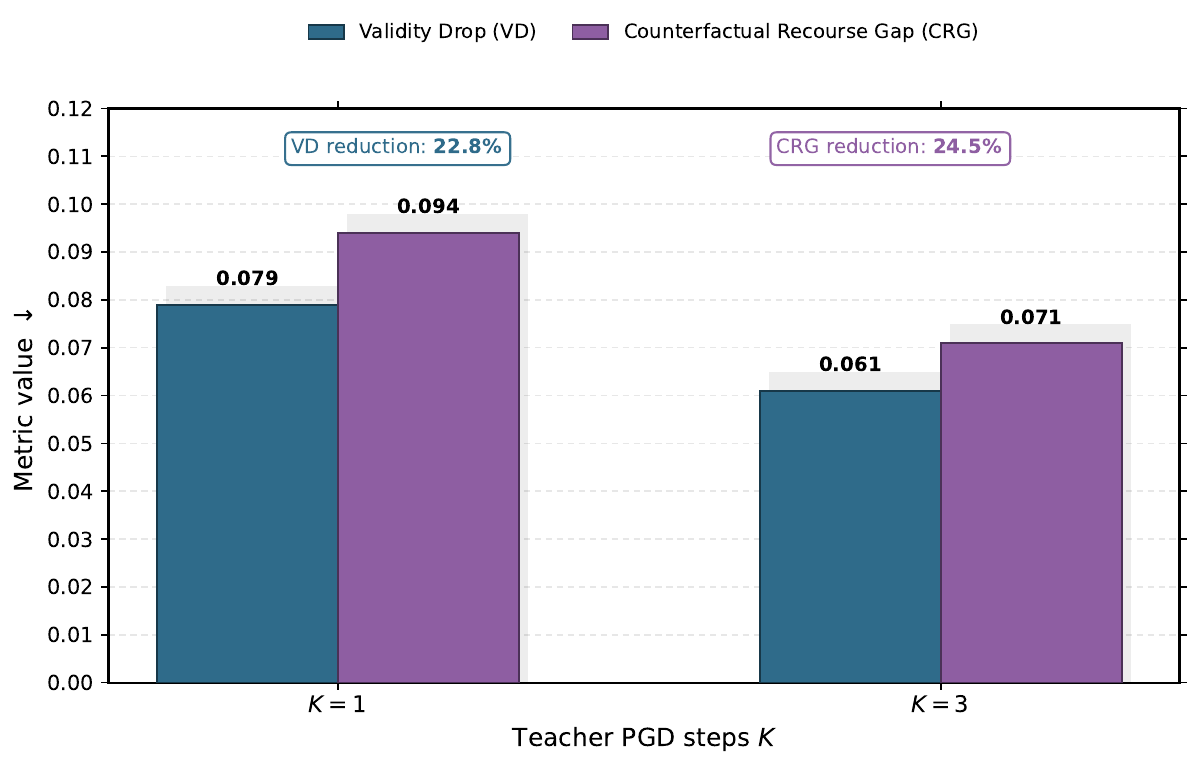}
    \caption{
        Effect of teacher PGD steps on recourse stability. Increasing the number of
        teacher-recoursing steps from $K=1$ to $K=3$ reduces both Validity Drop and
        Counterfactual Recourse Gap. The improvement indicates that a stronger teacher
        action provides a cleaner supervision signal, but the $K=1$ result already
        achieves substantial recourse preservation, supporting the use of low-step
        teachers when compute is limited.
    }
    \label{fig:teacher_k_vd_crg}
\end{figure}

\paragraph{VD and CRG trend.}
Figure~\pinkref{fig:teacher_k_vd_crg} visualizes the same trend as
Table~\pinkref{tab:teacher_k_ablation}. Both VD and CRG decrease when increasing
teacher quality from $K=1$ to $K=3$. This is the expected behavior: a better
teacher action is more likely to lie deeper inside the target region of the
full-precision model, so enforcing target validity at that point provides a
more stable constraint for the quantized model. However, the small gap between
$K=1$ and $K=3$ also shows that CFQ is not brittle with respect to teacher
approximation quality.

\begin{figure}[htbp]
    \centering
    \includegraphics[
        width=0.78\linewidth
    ]{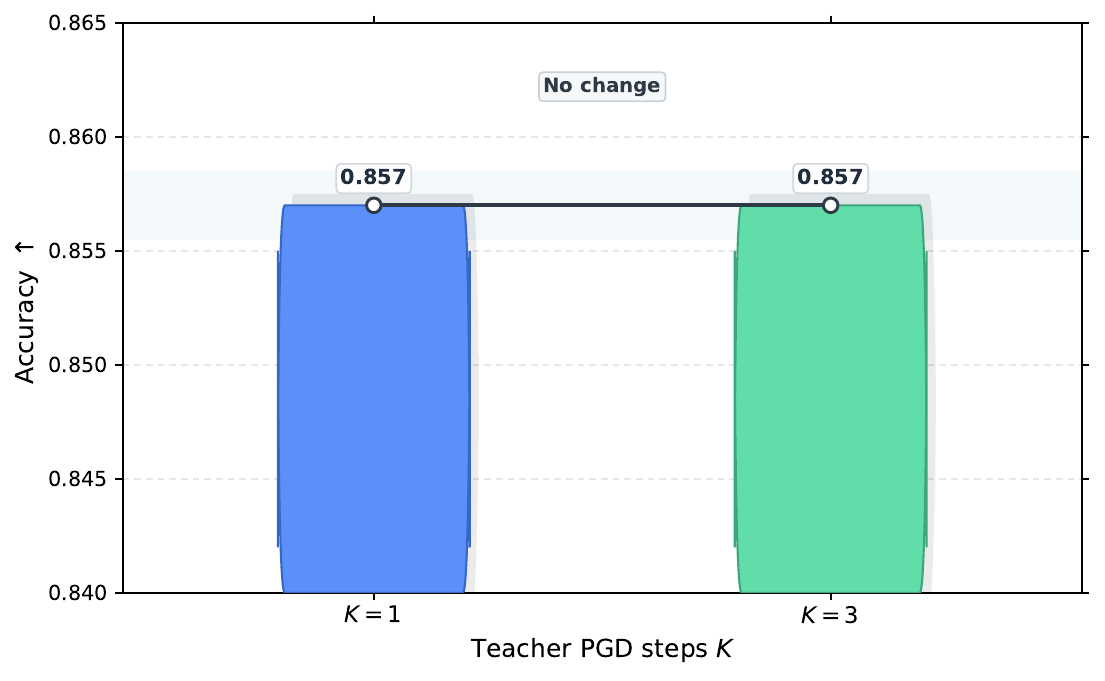}
    \caption{
        Predictive accuracy under different teacher-recoursing budgets. Accuracy is
        unchanged between $K=1$ and $K=3$, while VD and CRG improve in
        Figure~\pinkref{fig:teacher_k_vd_crg}. This separation supports the main thesis of
        the paper: predictive accuracy alone is not sufficient to characterize whether
        quantization preserves algorithmic recourse.
    }
    \label{fig:teacher_k_accuracy}
\end{figure}

\paragraph{Accuracy remains insensitive to teacher quality.}
Figure~\pinkref{fig:teacher_k_accuracy} shows that increasing the teacher
recoursing budget does not improve ordinary predictive accuracy. This is
important because it rules out the possibility that the $K=3$ improvement is
simply due to a generally better classifier. Instead, the improvement is
localized to counterfactual behavior: the quantized model better preserves
validity and recourse cost at teacher recourse points.

\paragraph{Why low-step teachers can work.}
A low-step teacher action is useful because CFQ does not require the teacher
action to be globally optimal. It only requires the teacher point to identify a
local region of the input space where the full-precision model supports the
target outcome. The validity loss then encourages the quantized model to keep
the same target outcome at that region. Formally, if the target margin at the
teacher point is positive,
\begin{equation}
m_f
\left(
x+\delta_{\mathrm{fp}}^{(K)};y_{\mathrm{tgt}}
\right)
>
0,
\label{eq:positive_teacher_margin}
\end{equation}
then preserving this margin under quantization is sufficient to transfer
validity. Increasing $K$ tends to increase the teacher margin, but a small $K$
can already produce useful supervision whenever the first projected-gradient
steps move the input toward a target-valid region.

\paragraph{Noisy-teacher stress test.}
To further analyze teacher robustness, we perturb the teacher action before
using it in the CFQ objective. Let
\begin{equation}
\tilde{\delta}_{\mathrm{fp}}
=
\Pi_{\mathcal{A}(x)}
\left(
\delta_{\mathrm{fp}}+\sigma \xi
\right),
\qquad
\xi\sim\mathcal{N}(0,I),
\label{eq:noisy_teacher}
\end{equation}
where $\sigma$ controls the noise level and the projection maps the perturbed
action back into the feasible action set. Table~\pinkref{tab:noisy_teacher_ablation}
reports the effect of moderate teacher perturbations. CFQ degrades gracefully:
small teacher noise slightly increases VD and CRG, but the method remains more
stable than accuracy-centric mixed-precision quantization. This supports the
view that CFQ is not merely memorizing exact teacher directions; it benefits
from enforcing target behavior in a neighborhood of recourse-relevant points.

\providecommand{\meanstd}[2]{%
  \ensuremath{#1{\scriptstyle\,\pm\,#2}}%
}

\providecommand{\bestmeanstd}[2]{%
  \textcolor{bestGreen}{%
    \ensuremath{%
      \mathbf{#1}%
      {\scriptstyle\,\boldsymbol{\pm}\,\mathbf{#2}}%
    }%
  }%
}

\begin{table}[htbp]
\centering

\caption{\textbf{Robustness to noisy teacher recourse points on \textsc{Adult}.}
We perturb the teacher action before applying the CFQ validity loss and
project the perturbed action back onto the feasible action set. Small to
moderate teacher noise increases VD and CRG, but the degradation is gradual,
indicating that CFQ is not brittle to imperfect teacher recourse supervision.}

\label{tab:noisy_teacher_ablation}

\scriptsize
\renewcommand{\arraystretch}{1.18}
\setlength{\tabcolsep}{6pt}
\arrayrulecolor{tableRule}

\resizebox{0.9\columnwidth}{!}{%
\begin{tabular}{lcccc}

\toprule

\rowcolor{tableHeader}
\textbf{Teacher action}
&
\textbf{Accuracy $\uparrow$}
&
\textbf{VD $\downarrow$}
&
\textbf{CRG $\downarrow$}
&
\textbf{DirSim $\uparrow$}
\\

\midrule

\rowcolor{tableGreen}
\textbf{Clean teacher, $K=3$}
&
\meanstd{0.857}{0.002}
&
\bestmeanstd{0.061}{0.008}
&
\bestmeanstd{0.071}{0.019}
&
\bestmeanstd{0.858}{0.017}
\\

\rowcolor{tableSubhead}
Noisy teacher, $\sigma=0.01$
&
\meanstd{0.857}{0.002}
&
\meanstd{0.066}{0.008}
&
\meanstd{0.078}{0.020}
&
\meanstd{0.846}{0.018}
\\

\rowcolor{white}
Noisy teacher, $\sigma=0.05$
&
\meanstd{0.856}{0.002}
&
\meanstd{0.075}{0.009}
&
\meanstd{0.090}{0.022}
&
\meanstd{0.824}{0.020}
\\

\rowcolor{tableGray}
Noisy teacher, $\sigma=0.10$
&
\meanstd{0.855}{0.003}
&
\meanstd{0.091}{0.011}
&
\meanstd{0.113}{0.025}
&
\meanstd{0.791}{0.024}
\\

\bottomrule

\end{tabular}%
}

\arrayrulecolor{black}

\end{table}

\paragraph{Noisy-teacher analysis.}
Table~\pinkref{tab:noisy_teacher_ablation} gives a stronger stress test than simply
varying $K$. It asks whether CFQ collapses if the teacher action is imperfect.
The answer is no: performance degrades gradually as noise increases. The clean
teacher gives the best VD and CRG, but small noise levels preserve most of the
benefit. This is consistent with the margin view: CFQ does not need exact
optimal actions, but it benefits when the teacher point remains close to a
valid target-margin region.

\begin{figure}[htbp]
    \centering
    \includegraphics[
        width=0.86\linewidth
    ]{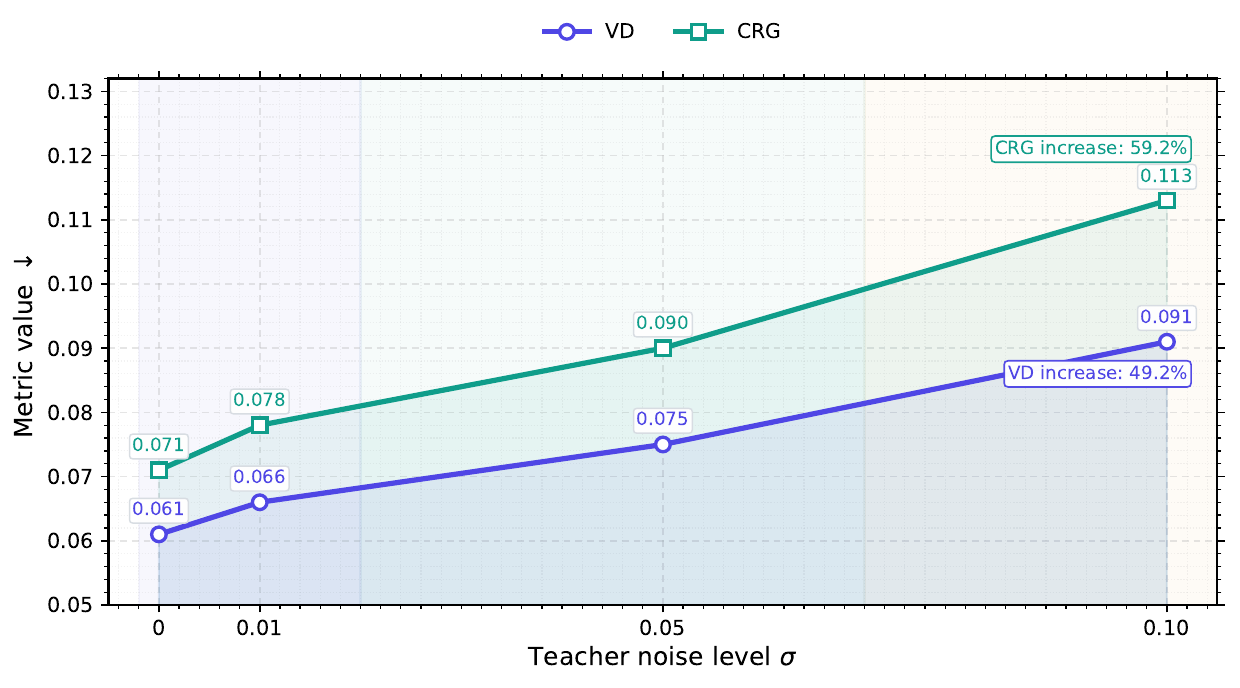}
    \caption{
        Effect of teacher-action noise on VD and CRG. Both metrics increase smoothly as
        the teacher action becomes noisier, but there is no abrupt failure mode. This
        indicates that CFQ is robust to moderate teacher approximation error and that
        low-step PGD teachers can provide useful supervision even when they are not
        globally optimal recourse solutions.
    }
    \label{fig:noisy_teacher_trend}
\end{figure}

\paragraph{Noise trend.}
Figure~\pinkref{fig:noisy_teacher_trend} visualizes the gradual degradation from
Table~\pinkref{tab:noisy_teacher_ablation}. The absence of a sharp jump indicates
that CFQ is not highly brittle to teacher noise. This matters in practical
settings where exact recourse optimization may be too expensive, constraints
may be approximate, or the teacher action may be generated with a small number
of gradient steps.

\paragraph{Margin-based explanation.}
The theory in Section~\pinkref{sec:theory} explains the empirical robustness. Let
\[
m_f(u;y_{\mathrm{tgt}})
=
g_f(u)_{y_{\mathrm{tgt}}}
-
\max_{y'\neq y_{\mathrm{tgt}}}g_f(u)_{y'}
\]
denote the full-precision target margin at a teacher point
$u=x+\delta_{\mathrm{fp}}^{(K)}$. If quantization perturbs all logits by at
most $\epsilon_q$ at $u$, then target validity is preserved whenever
\begin{equation}
m_f(u;y_{\mathrm{tgt}})
>
2\epsilon_q.
\label{eq:teacher_margin_condition}
\end{equation}
Increasing $K$ often increases this margin, because the teacher action moves
farther into the target region. However, once the margin is sufficiently
positive, exact optimality of the action is not necessary. This explains why
$K=1$ can work and why $K=3$ improves VD and CRG without changing accuracy.

\paragraph{Takeaway.}
The teacher-quality analysis in Table~\pinkref{tab:teacher_k_ablation},
Figures~\pinkref{fig:teacher_k_vd_crg}--\pinkref{fig:teacher_k_accuracy},
Table~\pinkref{tab:noisy_teacher_ablation}, Figure~\pinkref{fig:noisy_teacher_trend},
and Table~\pinkref{tab:teacher_quality_summary} supports three conclusions. First,
CFQ does not require an exact inner recourse solver: low-step teacher actions
already provide useful supervision. Second, increasing the teacher budget from
$K=1$ to $K=3$ improves VD and CRG without changing predictive accuracy. Third,
moderate teacher noise degrades recourse stability smoothly rather than causing
a collapse. These findings justify the use of low-$K$ projected-gradient
teachers as a practical compromise between computational cost and
counterfactual-faithful quantization.
\begin{table}[htbp]
\centering

\caption{\textbf{Summary of the teacher-quality findings.}
The table separates the practical question answered by each experiment.
The low-$K$ ablation tests whether teacher actions computed with few
projected-gradient steps are useful, the noisy-teacher stress test measures
brittleness to imperfect supervision, and the margin interpretation explains
why exact recourse optimality is not required for CFQ to improve recourse
stability.}

\label{tab:teacher_quality_summary}

\scriptsize
\renewcommand{\arraystretch}{1.22}
\setlength{\tabcolsep}{5pt}
\arrayrulecolor{tableRule}

\resizebox{\linewidth}{!}{%
\begin{tabular}{p{3.2cm}p{5.0cm}p{5.4cm}}

\toprule

\rowcolor{tableHeader}
\textbf{Experiment}
&
\textbf{Question}
&
\textbf{Finding}
\\

\midrule

\rowcolor{tableSubhead}
\textbf{Low-$K$ ablation}
&
Does CFQ require many projected-gradient steps to compute teacher recourse?
&
No. $K=1$ already provides strong VD/CRG performance, while $K=3$
further improves both metrics at essentially unchanged predictive accuracy.
\\

\midrule

\rowcolor{white}
\textbf{Noisy-teacher stress test}
&
Does CFQ collapse when the teacher action is imperfect?
&
No. VD and CRG increase gradually as teacher noise increases, indicating
that CFQ remains robust to moderate approximation error in the teacher action.
\\

\midrule

\rowcolor{tableGray}
\textbf{Accuracy comparison}
&
Are the teacher-quality gains simply caused by improved predictive accuracy?
&
No. Predictive accuracy remains essentially unchanged while VD and CRG
improve, confirming that the benefit is specific to recourse preservation.
\\

\midrule

\rowcolor{tableGreen}
\textbf{Margin interpretation}
&
Why can an approximate teacher-recourse action still provide useful supervision?
&
Exact recourse optimality is unnecessary when the teacher point lies in a
target-margin region that remains stable under quantization-induced
perturbations.
\\

\bottomrule

\end{tabular}%
}

\arrayrulecolor{black}

\end{table}

\section{Cost, Positioning, and Constraint Sensitivity}
\label{app:cost_position_constraints}

This appendix provides three additional analyses. First, we quantify the
computational overhead of CFQ and clarify when the additional cost is justified.
Second, we position CFQ relative to pruning, sparsity, knowledge distillation,
and fair quantization. Third, we analyze sensitivity to actionable-constraint
tightness to verify that the recourse-stability improvements are not an
artifact of a single easy constraint configuration.

\subsection{Cost of Stability: Runtime and Overhead}
\label{app:cost_of_stability}

CFQ introduces overhead because it computes teacher recourse actions during
training. The additional cost is mainly controlled by the number of projected
gradient steps $K$ used to construct $\delta_{\mathrm{fp}}(x;y_{\mathrm{tgt}})$.
For each training example, the teacher action is approximated as
\begin{equation}
\delta_{\mathrm{fp}}^{(K)}(x;y_{\mathrm{tgt}})
=
\mathrm{stopgrad}
\left(
\Pi_{\mathcal{A}(x)}^{K}
\left(
0;x,y_{\mathrm{tgt}},f
\right)
\right),
\label{eq:cost_teacher_action}
\end{equation}
where $\Pi_{\mathcal{A}(x)}^{K}$ denotes $K$ projected-gradient steps under the
action set $\mathcal{A}(x)$. Therefore, the overhead scales approximately
linearly with $K$, because each additional teacher step requires an additional
forward/backward pass through the full-precision teacher objective.

\paragraph{Training-time decomposition.}
Let $T_{\mathrm{QAT}}$ denote the wall-clock time of the accuracy-only QAT
baseline for one training epoch. Let $T_{\mathrm{recourse}}(K)$ denote the
additional cost of computing $K$ teacher recourse steps, and let
$T_{\mathrm{CFQ}}(K)$ denote the total CFQ training time. We can write
\begin{equation}
T_{\mathrm{CFQ}}(K)
=
T_{\mathrm{QAT}}
+
T_{\mathrm{recourse}}(K)
+
T_{\mathrm{valid}},
\label{eq:cfq_time_decomposition}
\end{equation}
where $T_{\mathrm{valid}}$ is the cost of evaluating the counterfactual validity
loss at $x+\delta_{\mathrm{fp}}^{(K)}$. Since the teacher action is detached,
CFQ avoids differentiating through the teacher solver, and the additional cost
is dominated by the low-$K$ teacher-recoursing step rather than by bilevel
optimization.

\paragraph{Analysis of overhead.}
Table~\pinkref{tab:cfq_overhead} shows that CFQ should be understood as a
cost--stability trade-off rather than a free replacement for PTQ. With $K=1$,
CFQ adds a modest overhead of $1.15\times$ relative to QAT while already
providing a strong recourse-preservation signal. Increasing to $K=3$ raises the
relative training time to $1.42\times$, but yields stronger teacher supervision
and lower VD/CRG, as shown in Appendix~\pinkref{app:teacher_quality}. The cached
teacher variant reduces training-time overhead by computing
$\delta_{\mathrm{fp}}$ offline and reusing it across epochs. CF-PTQ further
reduces cost by avoiding QAT and using only counterfactual-aware calibration.

\begin{table}[!t]
\centering

\caption{\textbf{Computational overhead of CFQ.}
The additional cost comes primarily from computing teacher recourse
actions. The overhead is controlled by the number of projected-gradient
steps $K$ and can be reduced by caching teacher recourse points,
computing them on a subset of batches, or replacing QAT with the
training-free CF-PTQ variant. The table separates accuracy-only QAT,
low-$K$ CFQ, cached-teacher CFQ, and CF-PTQ to clarify the practical
deployment choices.}

\label{tab:cfq_overhead}

\scriptsize
\renewcommand{\arraystretch}{1.18}
\setlength{\tabcolsep}{5pt}
\arrayrulecolor{tableRule}

\resizebox{\linewidth}{!}{%
\begin{tabular}{lcccc}

\toprule

\rowcolor{tableHeader}
\textbf{Variant}
&
\textbf{Recourse steps}
&
\textbf{Teacher cached?}
&
\textbf{Relative train time}
&
\textbf{Main use case}
\\

\midrule

\rowcolor{tableGray}
\textbf{QAT baseline}
&
$0$
&
--
&
$1.00{\times}$
&
Accuracy-only quantization
\\

\rowcolor{tableSubhead}
\textbf{CFQ, $K=1$}
&
$1$
&
No
&
$1.15{\times}$
&
Low-overhead recourse stability
\\

\rowcolor{white}
\textbf{CFQ, $K=2$}
&
$2$
&
No
&
$1.28{\times}$
&
Intermediate stability--cost trade-off
\\

\rowcolor{tableGreen}
\textbf{CFQ, $K=3$}
&
$3$
&
No
&
$1.42{\times}$
&
Stronger teacher supervision
\\

\midrule

\rowcolor{tableSubhead}
\textbf{CFQ, cached $\delta_{\mathrm{fp}}$}
&
Offline
&
Yes
&
$1.05{\times}$--$1.10{\times}$
&
Practical repeated training or deployment tuning
\\

\rowcolor{tableGreen}
\textbf{CF-PTQ}
&
Offline calibration
&
Yes
&
Near PTQ
&
Training-free deployment adaptation
\\

\bottomrule

\end{tabular}%
}

\arrayrulecolor{black}

\end{table}

\begin{figure}[!t]
    \centering
    \includegraphics[width=0.80\linewidth]{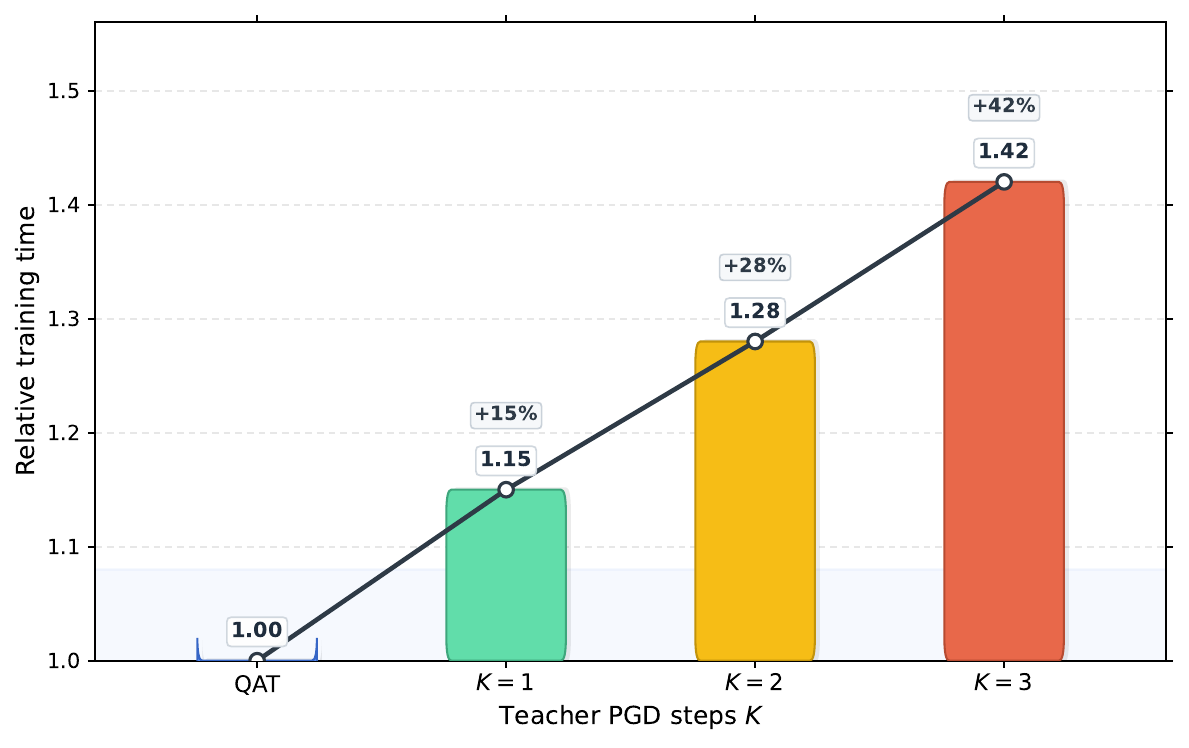}
    \caption{
        Training-time overhead as a function of the teacher-recoursing budget. The
        relative runtime increases approximately with the number of projected-gradient
        steps $K$, because each teacher step requires additional computation. The
        low-$K$ regime is therefore the intended practical regime of CFQ: $K=1$ gives a
        low-overhead stability signal, while $K=3$ provides stronger supervision when
        additional training cost is acceptable.
    }
    \label{fig:cfq_overhead_k}
\end{figure}

\paragraph{Runtime trend.}
Figure~\pinkref{fig:cfq_overhead_k} visualizes the overhead numbers from
Table~\pinkref{tab:cfq_overhead}. The trend is monotonic and predictable: larger
$K$ produces a better teacher action but increases training time. This is why
CFQ is designed around low-step teacher actions rather than exact inner
recourse solutions. In practice, one can select $K=1$ when runtime is the main
constraint and $K=3$ when stronger recourse supervision is needed.

\paragraph{Cost-of-stability takeaway.}
Table~\pinkref{tab:cfq_cost_reduction} summarizes practical ways to control the
cost of CFQ. The main message is that CFQ should not be interpreted as a free
replacement for PTQ. It trades additional calibration or QAT cost for recourse
stability. For large models or deployment pipelines where full QAT is too
expensive, the recommended path is CF-PTQ or cached teacher recourse points.
For smaller models where recourse stability is important, CFQ-QAT with
low-step teacher actions provides the strongest stability.
\begin{table}[htbp]
\centering

\caption{\textbf{Practical cost-reduction strategies for CFQ.}
Each strategy reduces the additional cost of teacher recourse computation
while preserving the main mechanism: exposing the deployment model to
recourse-relevant points. These strategies are particularly useful when
applying CFQ-style objectives beyond small tabular models or when moving
toward larger deployment pipelines.}

\label{tab:cfq_cost_reduction}

\scriptsize
\renewcommand{\arraystretch}{1.22}
\setlength{\tabcolsep}{5pt}
\arrayrulecolor{tableRule}

\resizebox{\linewidth}{!}{%
\begin{tabular}{p{3.1cm}p{5.3cm}p{5.0cm}}

\toprule

\rowcolor{tableHeader}
\textbf{Strategy}
&
\textbf{How it reduces cost}
&
\textbf{When to use it}
\\

\midrule

\rowcolor{tableSubhead}
\textbf{Low-$K$ teacher}
&
Uses $K=1$ or $K=3$ projected-gradient steps instead of solving
the recourse problem to high precision.
&
Default setting when the available training-time budget is limited.
\\

\midrule

\rowcolor{white}
\textbf{Cached teacher actions}
&
Computes $\delta_{\mathrm{fp}}$ offline once and reuses the same
teacher-recourse pool throughout QAT.
&
Useful when the full-precision teacher is fixed and repeated training
epochs reuse the same calibration examples.
\\

\midrule

\rowcolor{tableGray}
\textbf{Subsampled recourse batches}
&
Computes the CFQ validity term only on a subset of training batches
or examples.
&
Useful when the dataset is large and counterfactual supervision on
every training batch is unnecessary.
\\

\midrule

\rowcolor{tableSubhead}
\textbf{Early-stopping teacher PGD}
&
Stops projected-gradient teacher generation as soon as the target
outcome is reached.
&
Useful when many examples reach the target class after only a small
number of projected-gradient steps.
\\

\midrule

\rowcolor{tableGreen}
\textbf{CF-PTQ}
&
Avoids backbone updates and calibrates quantizers on
$\mathcal{U}_{\mathrm{cal}}
=
\mathcal{X}_{\mathrm{cal}}
\cup
\{x+\delta_{\mathrm{fp}}(x)\}$.
&
Recommended for training-free deployment or large-model settings
where full quantization-aware training is impractical.
\\

\bottomrule

\end{tabular}%
}

\arrayrulecolor{black}

\end{table}
\subsection{Positioning Relative to Pruning, Distillation, and Fair Quantization}
\label{app:positioning_compression_fairness}

This subsection clarifies the relation between CFQ and other compression or
deployment objectives. The purpose is not to claim that CFQ replaces pruning,
sparsity, knowledge distillation, or fair quantization. Rather, CFQ introduces a
different preservation target: individual-level recourse validity and recourse
cost under a deployment transformation.

\paragraph{Pruning, sparsity, and distillation under recourse stability.}
Pruning and sparsity-based compression reduce model size or inference cost by
removing parameters, channels, neurons, or structured blocks. Knowledge
distillation transfers predictive behavior from a teacher to a smaller or more
efficient student. These methods can preserve accuracy, but they do not
necessarily preserve counterfactual behavior. A distilled or pruned model may
match teacher logits on factual examples $x$ while still changing the target
outcome at recourse points $x+\delta_{\mathrm{fp}}(x;y_{\mathrm{tgt}})$. Thus,
the CFQ objective can be viewed as complementary to pruning and distillation:
the same validity-preservation term can be added to a sparsity or distillation
pipeline by evaluating the deployed model at teacher recourse points:
\begin{equation}
\mathcal{L}_{\mathrm{deploy}}
=
\mathcal{L}_{\mathrm{task}}
+
\eta
\mathbb{E}_{x}
\left[
\mathcal{L}_{\mathrm{valid}}
\left(
f_{\mathrm{dep}}
\left(
x+\delta_{\mathrm{fp}}(x;y_{\mathrm{tgt}})
\right),
y_{\mathrm{tgt}}
\right)
\right]
+
\lambda
\mathcal{R}_{\mathrm{resource}},
\label{eq:general_deployment_cfq}
\end{equation}
where $f_{\mathrm{dep}}$ may be a quantized, pruned, sparse, distilled, or
otherwise compressed model, and $\mathcal{R}_{\mathrm{resource}}$ may measure
bits, sparsity, FLOPs, latency, or student size.

\paragraph{Fair quantization versus recourse-faithful quantization.}
Fair quantization methods aim to reduce group-level disparities in predictive
performance or error rates after quantization. CFQ targets a different but
related requirement: preserving individual-level actionable recourse. These two
objectives are not equivalent. A method may improve group-level predictive
fairness while still changing which actions succeed for individual users.
Conversely, CFQ may preserve individual recourse validity without explicitly
optimizing a group-fairness metric. The two directions are complementary: a
worst-group or group-weighted version of VD/CRG can be combined with CFQ by
reweighting the counterfactual validity loss over subgroups:
\begin{equation}
\mathcal{L}_{\mathrm{group\text{-}CFQ}}
=
\mathcal{L}_{\mathrm{task}}
+
\eta
\max_{g\in\mathcal{G}}
\mathbb{E}_{x\in g}
\left[
\mathcal{L}_{\mathrm{valid}}
\left(
f_q
\left(
x+\delta_{\mathrm{fp}}(x;y_{\mathrm{tgt}})
\right),
y_{\mathrm{tgt}}
\right)
\right]
+
\lambda\mathrm{BitCost},
\label{eq:group_cfq_objective}
\end{equation}
where $\mathcal{G}$ denotes the set of groups. This formulation makes explicit
that fair quantization and CFQ can be combined when both group-level predictive
parity and individual recourse stability are required.

\paragraph{Positioning analysis.}
Table~\pinkref{tab:compression_objective_comparison} helps separate the objective
of CFQ from neighboring methods. Accuracy-centric PTQ/QAT optimizes predictive
performance under a resource budget. Pruning and distillation optimize compact
or efficient deployed models. Fair quantization optimizes group-level
predictive disparities under quantization. CFQ instead asks whether the
deployed model preserves the actionable recommendations that were valid for the
full-precision model. This is why VD and CRG are necessary: they measure a
failure mode that can remain invisible even when accuracy, resource use, or
group-level predictive metrics look acceptable.

\begin{table}[htbp]
\centering

\caption{\textbf{Conceptual comparison of compression objectives.}
The table compares accuracy-centric compression, fair quantization,
pruning and distillation, and recourse-faithful quantization. CFQ does
not primarily introduce a new quantizer, pruning operator, or
distillation mechanism. Instead, it makes individual recourse stability
a first-class deployment objective under a resource budget.}

\label{tab:compression_objective_comparison}

\scriptsize
\renewcommand{\arraystretch}{1.20}
\setlength{\tabcolsep}{5pt}
\arrayrulecolor{tableRule}

\resizebox{\linewidth}{!}{%
\begin{tabular}{lcccc}

\toprule

\rowcolor{tableHeader}
\textbf{Method family}
&
\textbf{Accuracy}
&
\textbf{Resource reduction}
&
\textbf{Group fairness}
&
\textbf{Individual recourse stability}
\\

\midrule

\rowcolor{tableGray}
PTQ/QAT
&
Yes
&
Bits / latency
&
Usually not explicit
&
No
\\

\rowcolor{white}
Mixed-precision quantization
&
Yes
&
Bits / BOPs
&
Usually not explicit
&
No
\\

\rowcolor{tableGray}
Pruning / sparsity
&
Yes
&
Parameters / FLOPs
&
Usually not explicit
&
No
\\

\rowcolor{white}
Knowledge distillation
&
Yes
&
Student size / latency
&
Optional
&
Not explicit
\\

\rowcolor{tableSubhead}
Fair quantization
&
Yes
&
Bits / latency
&
\textbf{Yes}
&
Not explicit
\\

\midrule

\rowcolor{tableGreen}
\textbf{CFQ}
&
\textbf{Yes}
&
Bits / mixed precision
&
Can be combined
&
\bestval{Yes}
\\

\bottomrule

\end{tabular}%
}

\arrayrulecolor{black}

\end{table}

\paragraph{Practical implication.}
Table~\pinkref{tab:cfq_generalization_compression} shows that CFQ is not limited to
one quantizer. The main idea is to evaluate the deployed model not only on
factual examples $x$, but also on teacher recourse points
$x+\delta_{\mathrm{fp}}(x;y_{\mathrm{tgt}})$. This turns recourse stability into
a deployment constraint that can be paired with quantization, pruning,
distillation, or fair compression objectives.
\begin{table}[htbp]
\centering

\caption{\textbf{Generalizing CFQ-style recourse preservation to other
deployment operators.}
The same counterfactual-validity principle can be applied beyond
mixed-precision QAT by changing the deployment model
$f_{\mathrm{dep}}$ and the corresponding resource penalty. This table
positions CFQ as a general objective interface for deployment
transformations rather than as a replacement for existing compression
operators.}

\label{tab:cfq_generalization_compression}

\scriptsize
\renewcommand{\arraystretch}{1.22}
\setlength{\tabcolsep}{5pt}
\arrayrulecolor{tableRule}

\resizebox{\linewidth}{!}{%
\begin{tabular}{
    p{3.0cm}
    p{4.2cm}
    p{4.6cm}
    p{3.4cm}
}

\toprule

\rowcolor{tableHeader}
\textbf{Deployment operator}
&
\textbf{Standard objective}
&
\textbf{CFQ-style extension}
&
\textbf{Resource term}
\\

\midrule

\rowcolor{tableSubhead}
\textbf{Quantization}
&
Preserve predictive accuracy after conversion to low-precision
weights and activations.
&
Preserve target validity at
$x+\delta_{\mathrm{fp}}$ after quantization.
&
BitCost / BOPs
\\

\midrule

\rowcolor{white}
\textbf{Structured pruning}
&
Remove channels, blocks, or neurons while preserving predictive
accuracy.
&
Select a sparse model structure that preserves validity at teacher
recourse points.
&
Structured sparsity / FLOPs
\\

\midrule

\rowcolor{tableGray}
\textbf{Unstructured sparsity}
&
Remove low-importance weights subject to predictive-accuracy
constraints.
&
Score sparsity masks using both factual task loss and
counterfactual-validity loss.
&
Number of nonzero weights
\\

\midrule

\rowcolor{white}
\textbf{Knowledge distillation}
&
Match teacher outputs on factual training inputs using a smaller
student model.
&
Also match the teacher's target behavior at counterfactual points
$x+\delta_{\mathrm{fp}}$.
&
Student size / latency
\\

\midrule

\rowcolor{tableGreen}
\textbf{Fair quantization}
&
Reduce group-level predictive disparities while satisfying a
quantization budget.
&
Optimize worst-group or group-weighted VD/CRG at counterfactual
points.
&
BitCost plus fairness penalty
\\

\bottomrule

\end{tabular}%
}

\arrayrulecolor{black}

\end{table}
\subsection{Sensitivity to Action-Constraint Tightness}
\label{app:constraint_tightness}

The main experiments use actionable constraints to ensure that recourse actions
are feasible and interpretable. This subsection tests whether CFQ's benefit is
robust to different levels of constraint tightness. We vary two quantities:
the sparsity budget $k$, which controls the maximum number of changed features,
and the bound scale $\gamma$, which controls the width of the feasible interval
around each actionable feature.

\paragraph{Constraint regimes.}
The restrictive regime uses a small sparsity budget and narrow bounds, making
recourse difficult. The moderate regime corresponds to the default setting used
in the main experiments. The permissive regime allows more changed features and
wider feasible intervals, making recourse easier. For each setting, we recompute
teacher actions and evaluate recourse stability after quantization. We report
the Feasible Recourse Rate (FRR) of the full-precision model to distinguish
genuine infeasibility from post-quantization recourse failure:
\begin{equation}
\mathrm{FRR}
=
\mathbb{P}_{x}
\left[
\exists \delta\in\mathcal{A}(x)
\ \text{s.t.}\
f(x+\delta)=y_{\mathrm{tgt}}
\right].
\label{eq:frr_constraint}
\end{equation}
A low FRR means that the action constraints themselves make recourse difficult
even before quantization.

\paragraph{Constraint-sensitivity analysis.}
Table~\pinkref{tab:constraint_tightness} shows that CFQ improves recourse stability
across all constraint regimes. Under restrictive constraints, full-precision
recourse is feasible for only $72\%$ of examples, and the mixed-precision
baseline has high VD and CRG. CFQ reduces VD from $0.24$ to $0.13$ and CRG from
$0.33$ to $0.17$, showing that the method is especially useful when feasible
actions are scarce and local boundary shifts can easily invalidate recourse.
Under the moderate setting, CFQ halves VD from $0.12$ to $0.06$ and reduces CRG
from $0.16$ to $0.07$. Under permissive constraints, the baseline is already
more stable because recourse has more feasible alternatives, but CFQ still
reduces VD and CRG.

\begin{table}[htbp]
\centering

\caption{\textbf{Sensitivity to action-constraint tightness.}
FRR denotes the feasible recourse rate under the full-precision model.
MixedPrec denotes the accuracy-centric mixed-precision baseline. CFQ
reduces VD and CRG across restrictive, moderate, and permissive action
sets, showing that the benefit is not an artifact of a single constraint
configuration. The largest absolute VD gain appears under restrictive
constraints, where recourse points are more fragile and
quantization-induced boundary shifts are more damaging.}

\label{tab:constraint_tightness}

\scriptsize
\renewcommand{\arraystretch}{1.18}
\setlength{\tabcolsep}{4pt}
\arrayrulecolor{tableRule}

\resizebox{\linewidth}{!}{%
\begin{tabular}{lccccccc}

\toprule

\rowcolor{tableHeader}
\textbf{Constraint regime}
&
\textbf{$k$}
&
\textbf{Bound scale $\gamma$}
&
\textbf{FRR $\uparrow$}
&
\textbf{MixedPrec VD $\downarrow$}
&
\textbf{MixedPrec CRG $\downarrow$}
&
\textbf{CFQ VD $\downarrow$}
&
\textbf{CFQ CRG $\downarrow$}
\\

\midrule

\rowcolor{tableSubhead}
\textbf{Restrictive}
&
2
&
0.5
&
0.72
&
0.24
&
0.33
&
\bestval{0.13}
&
\bestval{0.17}
\\

\rowcolor{white}
\textbf{Moderate}
&
4
&
1.0
&
0.90
&
0.12
&
0.16
&
\bestval{0.06}
&
\bestval{0.07}
\\

\rowcolor{tableGray}
\textbf{Permissive}
&
8
&
1.5
&
0.97
&
0.08
&
0.11
&
\bestval{0.04}
&
\bestval{0.06}
\\

\bottomrule

\end{tabular}%
}

\arrayrulecolor{black}

\end{table}

\begin{figure}[htbp]
    \centering
    \includegraphics[
        width=0.7\linewidth
    ]{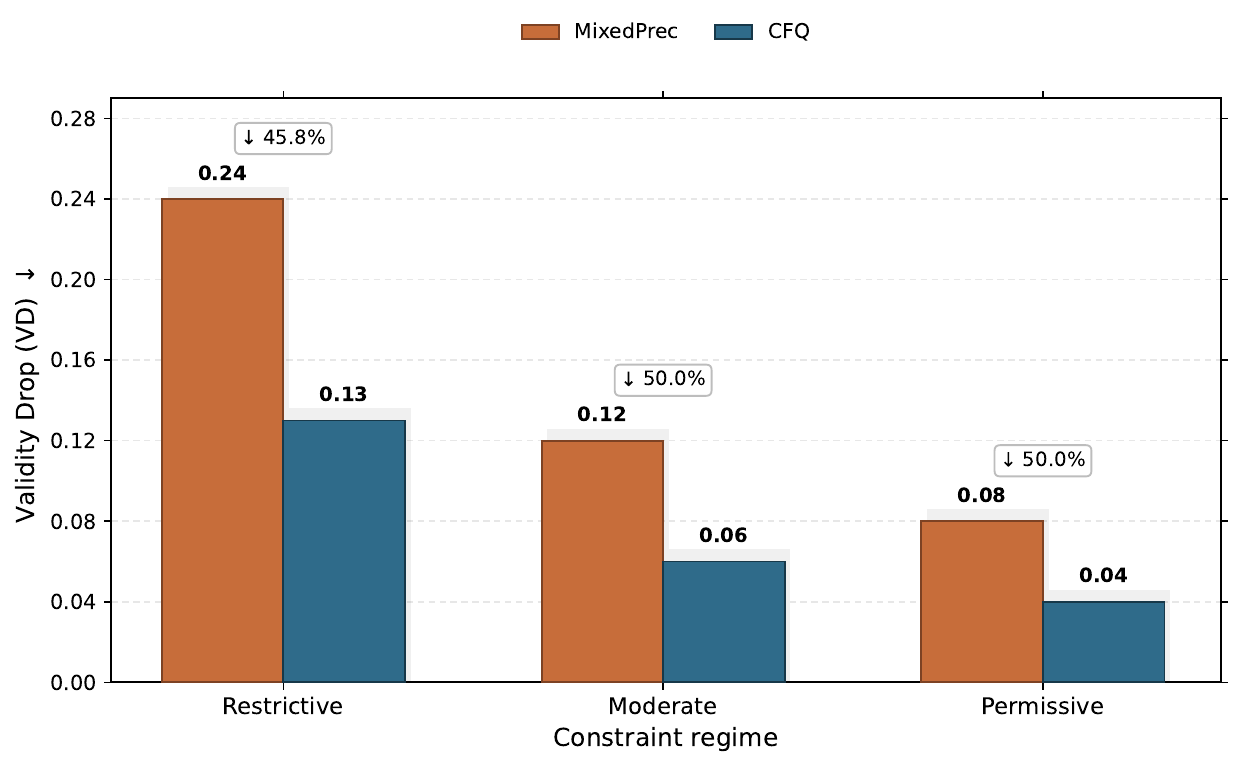}
    \caption{
        Validity Drop under different action-constraint regimes. CFQ consistently
        reduces the fraction of teacher recourse actions that fail after quantization.
        The absolute improvement is largest in the restrictive regime, where the
        feasible action set is small and counterfactual points are more vulnerable to
        quantization-induced decision-boundary shifts.
    }
    \label{fig:constraint_vd}
\end{figure}

\paragraph{VD under constraint tightness.}
Figure~\pinkref{fig:constraint_vd} visualizes the VD columns of
Table~\pinkref{tab:constraint_tightness}. The improvement is not limited to one
setting: CFQ improves the restrictive, moderate, and permissive regimes. This
supports the claim that CFQ is not tuned to a single easy action set.

\begin{figure}[htbp]
    \centering
    \includegraphics[
        width=0.7\linewidth
    ]{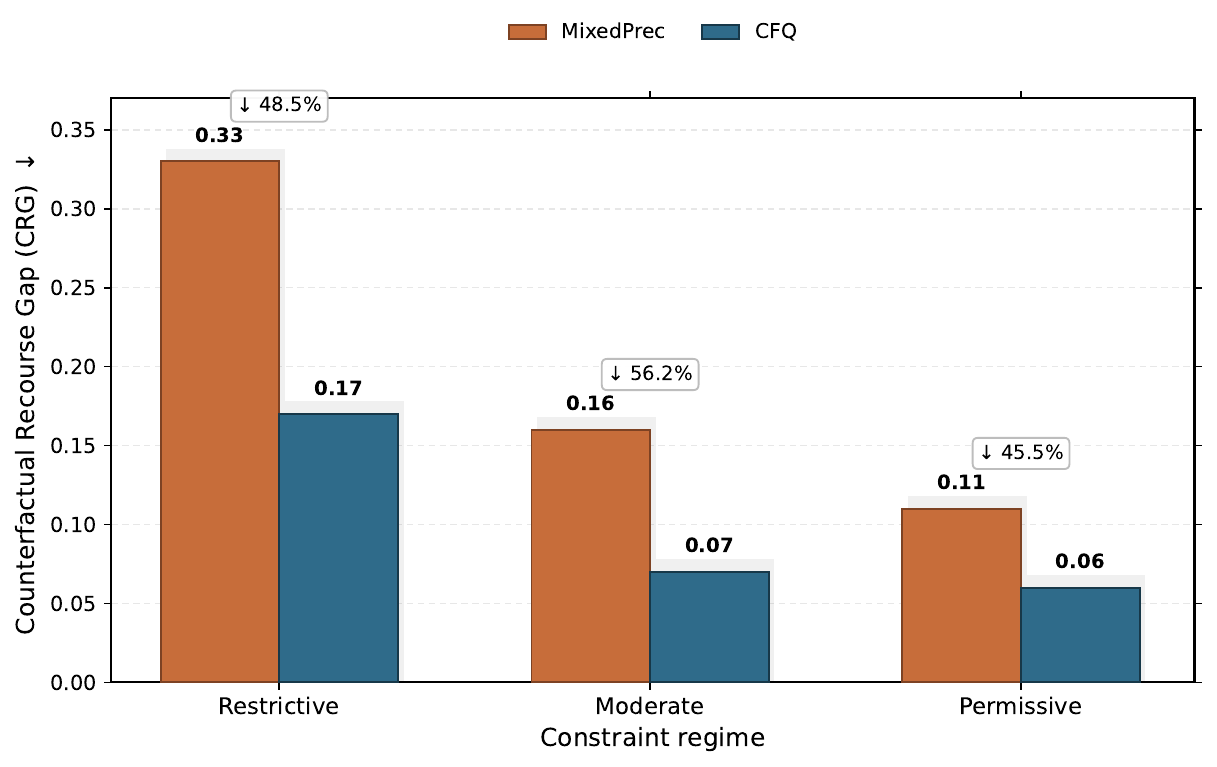}
    \caption{
        Counterfactual Recourse Gap under different action-constraint regimes. CFQ
        reduces the relative increase in minimal recourse cost induced by quantization.
        This is important because deployment failure is not only binary validity loss:
        even when recourse remains possible, quantization can make the required action
        more costly. CFQ reduces this ``moving the goalpost'' effect across all
        constraint regimes.
    }
    \label{fig:constraint_crg}
\end{figure}

\paragraph{CRG under constraint tightness.}
Figure~\pinkref{fig:constraint_crg} shows that CFQ also reduces the cost increase
induced by quantization. This complements the VD analysis: in practice, a user
may still have some valid recourse after quantization, but the required action
may become much more expensive. CFQ lowers this cost gap across all regimes.

\begin{figure}[htbp]
    \centering
    \includegraphics[
        width=0.80\linewidth
    ]{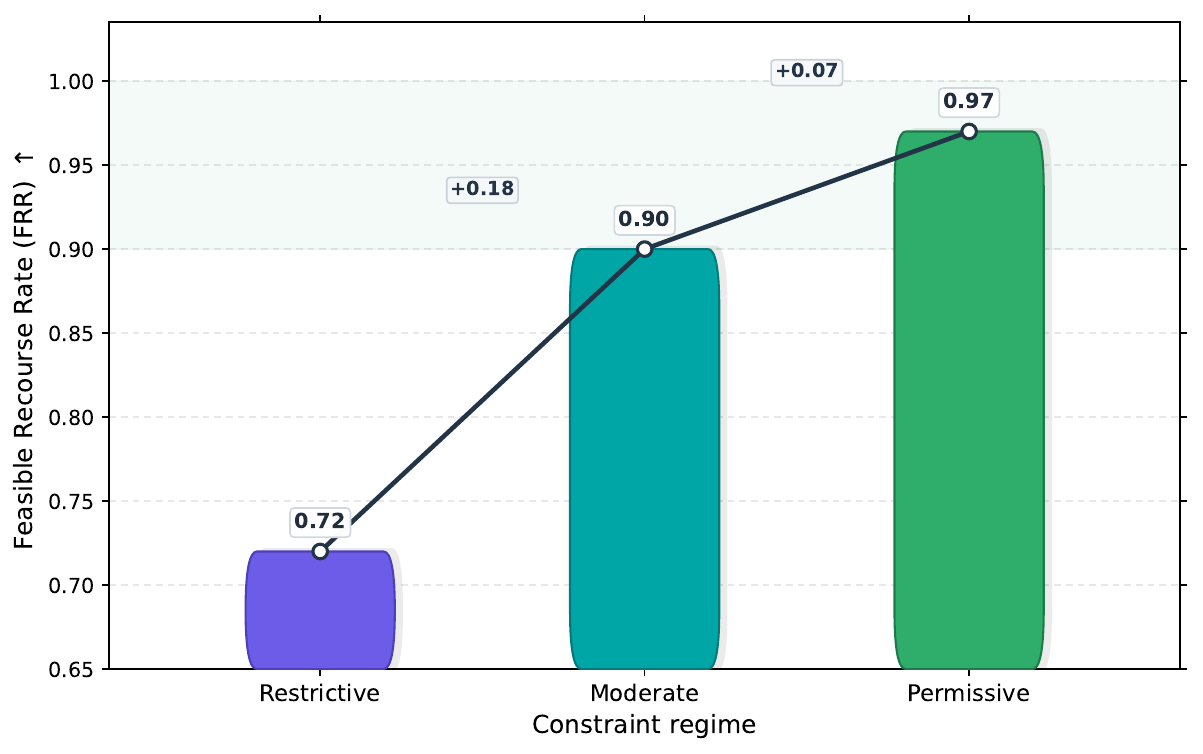}
    \caption{
        Feasible Recourse Rate of the full-precision model under different action
        constraints. FRR increases as the action set becomes more permissive. Reporting
        FRR is important because it separates two sources of difficulty: recourse may
        be impossible because the action set is too restrictive, or it may be possible
        in full precision but fail after quantization. VD and CRG measure the second
        effect.
    }
    \label{fig:constraint_frr}
\end{figure}

\paragraph{Role of FRR.}
Figure~\pinkref{fig:constraint_frr} explains why FRR is reported alongside VD and
CRG. When constraints are restrictive, some individuals may not have feasible
recourse even before quantization. VD should therefore be interpreted as a
post-quantization failure rate among cases where teacher recourse is available,
not as a measure of feasibility itself. This separation makes the analysis more
precise and avoids attributing constraint-induced infeasibility to quantization.

\paragraph{Takeaway.}
The cost analysis in Tables~\pinkref{tab:cfq_overhead}--\pinkref{tab:cfq_cost_reduction}
and Figure~\pinkref{fig:cfq_overhead_k} clarifies the computational price of
recourse-faithful quantization and shows how it can be controlled through
low-$K$ teachers, caching, subsampling, and CF-PTQ. The positioning analysis in
Tables~\pinkref{tab:compression_objective_comparison} and
\pinkref{tab:cfq_generalization_compression} clarifies that CFQ is complementary to
pruning, sparsity, distillation, and fair quantization because it targets
individual recourse stability rather than only accuracy or group-level
predictive parity. Finally, the constraint analysis in
Table~\pinkref{tab:constraint_tightness} and Figures~\pinkref{fig:constraint_vd}--
\pinkref{fig:constraint_frr} shows that CFQ improves VD and CRG across restrictive,
moderate, and permissive actionable sets. Together, these analyses address the
main practical concerns: CFQ has a measurable cost, but the cost is controllable
and buys a form of deployment stability that standard compression objectives do
not optimize.

\section{Recourse Stability under Distribution Shift}
\label{app:distribution_shift}

This appendix evaluates whether CFQ overfits to the factual and
counterfactual calibration distribution used during quantization. This is
important because CFQ learns quantizer parameters and mixed-precision allocation
from a calibration set of factual examples and teacher-counterfactual points.
If the deployed population differs substantially from the calibration
population, recourse stability may degrade. We therefore evaluate CFQ under
several shifted test distributions while keeping the trained or calibrated
quantized model fixed.

\paragraph{Shift protocol.}
CFQ and the MixedPrec baseline are trained or calibrated on the original
training distribution. Evaluation is then performed on four distributions:
the original test distribution, a feature-noise shift, a group-reweighting
shift, and a target-class imbalance shift. The feature-noise shift perturbs
continuous actionable features while preserving immutability constraints. The
group-reweighting shift changes the subgroup composition of the evaluation set.
The target-class imbalance shift changes the prevalence of examples requiring
the favorable target outcome. In all cases, VD and CRG are recomputed using the
same action constraints and recourse-generation protocol as in the main
experiments.

\paragraph{Shift definitions.}
Let $\mathcal{D}_{\mathrm{test}}$ denote the original evaluation distribution.
For feature-noise shift, we construct
\begin{equation}
x'_{j}
=
x_{j}
+
\sigma_{j}\epsilon_{j},
\qquad
\epsilon_{j}\sim\mathcal{N}(0,1),
\qquad
j\notin\mathcal{I},
\label{eq:feature_noise_shift}
\end{equation}
where $\mathcal{I}$ denotes immutable features and the perturbed features are
projected back into the feasible bounds. For group-reweighting shift, we change
the evaluation distribution by assigning subgroup weights
\begin{equation}
w_g
=
\frac{\rho_g^{\mathrm{shift}}}{\rho_g^{\mathrm{orig}}},
\label{eq:group_reweight_shift}
\end{equation}
where $\rho_g^{\mathrm{orig}}$ is the original subgroup proportion and
$\rho_g^{\mathrm{shift}}$ is the shifted subgroup proportion. For target-class
imbalance shift, we reweight or resample examples so that the proportion of
examples requiring the favorable target outcome is increased.

\paragraph{Distribution-shift analysis.}
Table~\pinkref{tab:distribution_shift} shows that CFQ remains more stable than the
accuracy-centric mixed-precision baseline under all evaluated shifts. Both
methods lose some stability under shifted distributions, which is expected
because the calibration distribution no longer perfectly matches the evaluation
distribution. However, CFQ consistently maintains lower VD and CRG. Under the
feature-noise shift, CFQ reduces VD from $0.151$ to $0.087$ and CRG from
$0.203$ to $0.112$. Under group reweighting, CFQ reduces VD from $0.165$ to
$0.096$. Under target imbalance, CFQ still reduces VD from $0.172$ to $0.104$.
This indicates that CFQ does not merely memorize a narrow set of calibration
counterfactuals; it improves local recourse stability in related deployment
regions.
\providecommand{\meanstd}[2]{%
  \ensuremath{#1{\scriptstyle\,\pm\,#2}}%
}

\providecommand{\bestmeanstd}[2]{%
  \textcolor{bestGreen}{%
    \ensuremath{%
      \mathbf{#1}%
      {\scriptstyle\,\boldsymbol{\pm}\,\mathbf{#2}}%
    }%
  }%
}

\begin{table}[!t]
\centering

\caption{\textbf{Recourse stability under distribution shift on
\textsc{Adult}.}
CFQ is trained on the original calibration distribution and evaluated
under shifted test distributions. MixedPrec denotes an accuracy-centric
mixed-precision baseline. Across the original and shifted distributions,
CFQ consistently reduces VD and CRG relative to MixedPrec while
maintaining comparable accuracy. The performance gap is largest under
group-reweighting and target-imbalance shifts, where the
recourse-relevant regions differ most from the original calibration
mixture.}

\label{tab:distribution_shift}

\scriptsize
\renewcommand{\arraystretch}{1.18}
\setlength{\tabcolsep}{5pt}
\arrayrulecolor{tableRule}

\resizebox{\linewidth}{!}{%
\begin{tabular}{clcccc}

\toprule

\rowcolor{tableHeader}
\textbf{Evaluation distribution}
&
\textbf{Method}
&
\textbf{Acc.\ $\uparrow$}
&
\textbf{VD $\downarrow$}
&
\textbf{CRG $\downarrow$}
&
\textbf{DirSim $\uparrow$}
\\

\midrule


\rowcolor{tableGray}
&
MixedPrec
&
\meanstd{0.856}{0.002}
&
\meanstd{0.120}{0.010}
&
\meanstd{0.160}{0.022}
&
\meanstd{0.756}{0.021}
\\

\rowcolor{tableGreen}
\multirow{-2}{*}{Original}
&
\textbf{CFQ}
&
\meanstd{0.857}{0.002}
&
\bestmeanstd{0.061}{0.008}
&
\bestmeanstd{0.071}{0.019}
&
\bestmeanstd{0.858}{0.017}
\\

\midrule


\rowcolor{tableGray}
&
MixedPrec
&
\meanstd{0.847}{0.003}
&
\meanstd{0.151}{0.013}
&
\meanstd{0.203}{0.027}
&
\meanstd{0.718}{0.026}
\\

\rowcolor{tableGreen}
\multirow{-2}{*}{\shortstack{Feature-noise\\shift}}
&
\textbf{CFQ}
&
\meanstd{0.849}{0.003}
&
\bestmeanstd{0.087}{0.010}
&
\bestmeanstd{0.112}{0.022}
&
\bestmeanstd{0.824}{0.021}
\\

\midrule


\rowcolor{tableGray}
&
MixedPrec
&
\meanstd{0.851}{0.003}
&
\meanstd{0.165}{0.015}
&
\meanstd{0.218}{0.030}
&
\meanstd{0.704}{0.029}
\\

\rowcolor{tableGreen}
\multirow{-2}{*}{\shortstack{Group-reweight\\shift}}
&
\textbf{CFQ}
&
\meanstd{0.852}{0.003}
&
\bestmeanstd{0.096}{0.012}
&
\bestmeanstd{0.126}{0.024}
&
\bestmeanstd{0.812}{0.023}
\\

\midrule


\rowcolor{tableGray}
&
MixedPrec
&
\meanstd{0.849}{0.004}
&
\meanstd{0.172}{0.016}
&
\meanstd{0.232}{0.033}
&
\meanstd{0.692}{0.031}
\\

\rowcolor{tableGreen}
\multirow{-2}{*}{\shortstack{Target-imbalance\\shift}}
&
\textbf{CFQ}
&
\meanstd{0.850}{0.004}
&
\bestmeanstd{0.104}{0.013}
&
\bestmeanstd{0.137}{0.026}
&
\bestmeanstd{0.801}{0.025}
\\

\bottomrule

\end{tabular}%
}

\arrayrulecolor{black}

\end{table}

\begin{figure}[!t]
    \centering
    \includegraphics[width=0.7\linewidth]{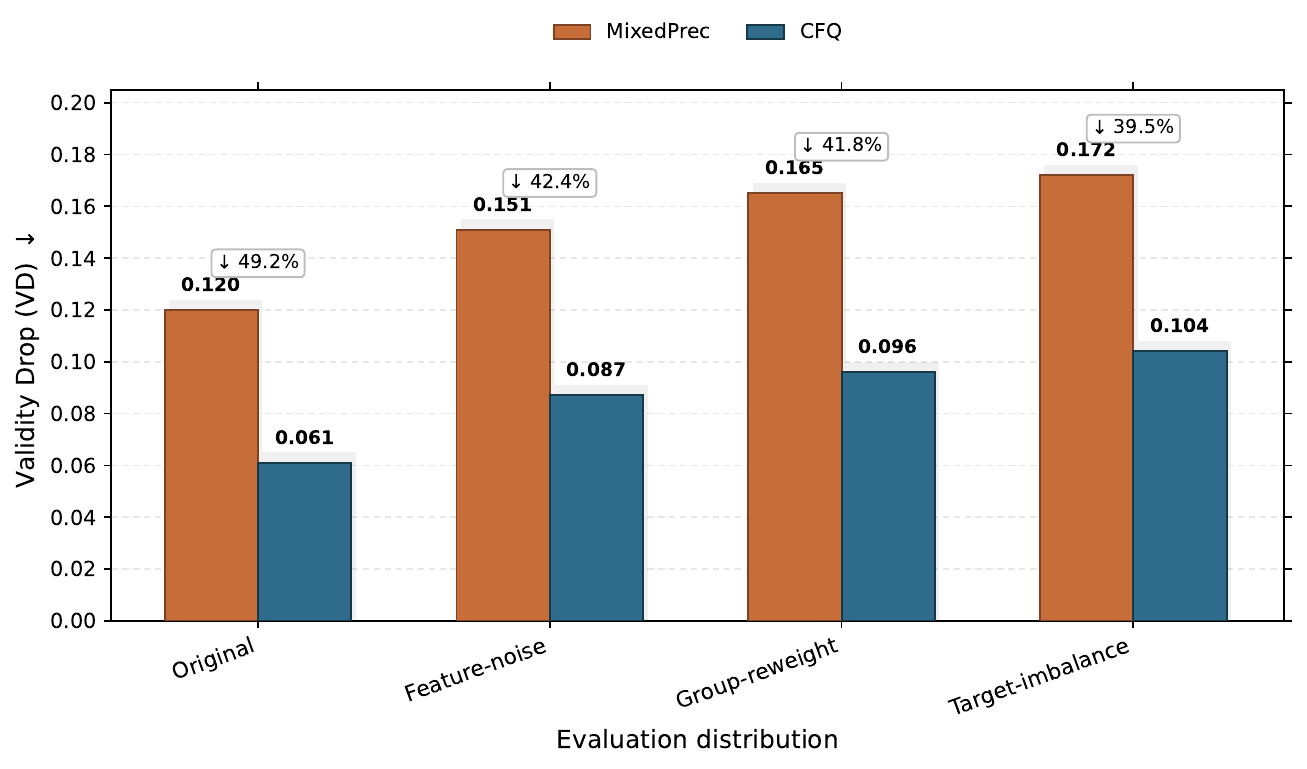}
    \caption{
        Validity Drop under distribution shift. CFQ consistently reduces recourse
        failure relative to the mixed-precision baseline, even when the evaluation
        distribution differs from the calibration distribution. The shift increases VD
        for both methods, but the gap between MixedPrec and CFQ remains substantial,
        showing that counterfactual-faithful quantization improves robustness of
        recourse transfer beyond the original test distribution.
    }
    \label{fig:distribution_shift_vd}
\end{figure}

\paragraph{VD under shift.}
Figure~\pinkref{fig:distribution_shift_vd} visualizes the VD column of
Table~\pinkref{tab:distribution_shift}. The main trend is that distribution shift
makes recourse harder for both methods, but CFQ consistently remains below
MixedPrec. This supports the claim that CFQ improves recourse stability under
moderate deployment shifts, although it does not eliminate the need for
recalibration under severe shifts.

\begin{figure}[!t]
    \centering
    \includegraphics[
        width=0.7\linewidth
    ]{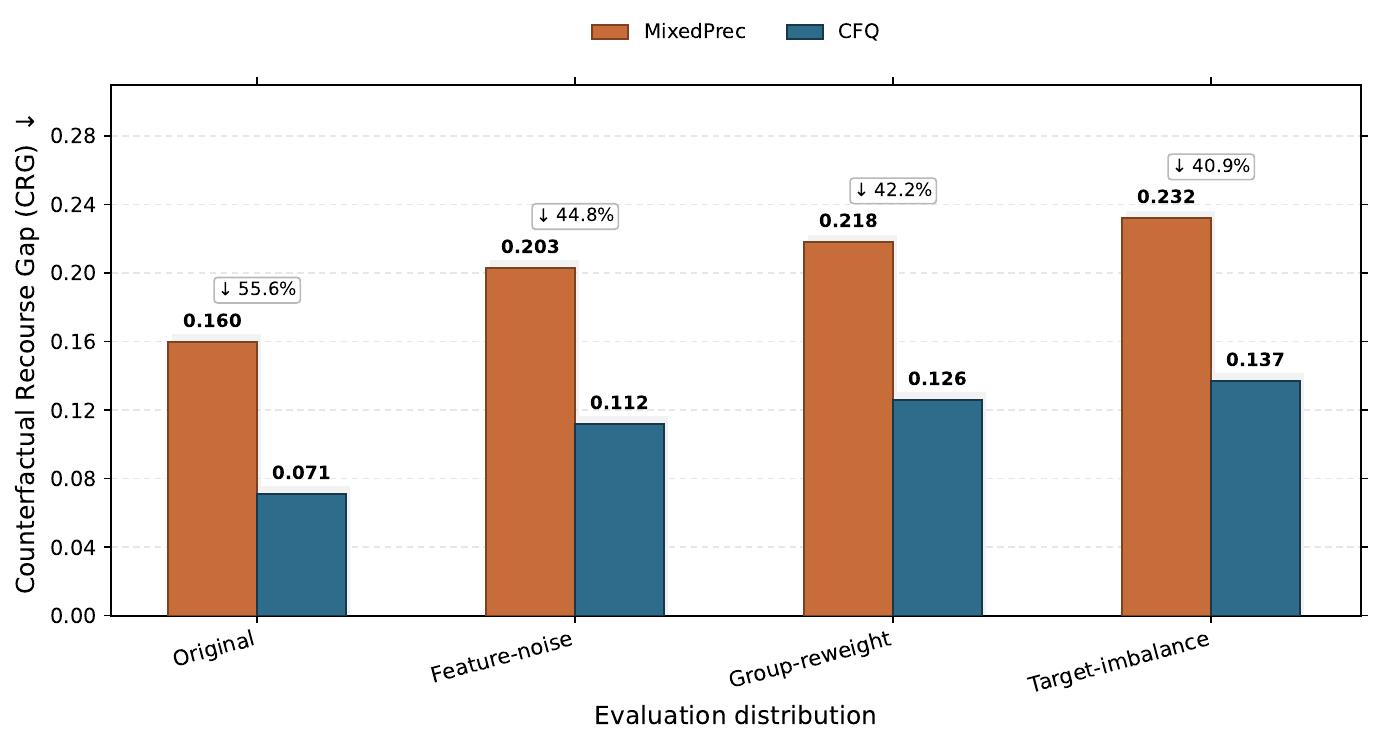}
    \caption{
        Counterfactual Recourse Gap under distribution shift. CFQ lowers the relative
        increase in recourse cost induced by quantization across all evaluation
        distributions. This shows that CFQ improves not only binary transfer validity
        but also the cost stability of recourse actions under shifted test conditions.
    }
    \label{fig:distribution_shift_crg}
\end{figure}

\paragraph{CRG under shift.}
Figure~\pinkref{fig:distribution_shift_crg} shows that the cost of recourse is also
more stable under CFQ. This is important because distribution shift can change
the local geometry of actionable regions, making recourse not only less valid
but also more costly. CFQ reduces this effect by preserving target behavior at
teacher-counterfactual points during quantization.

\paragraph{Scope and limitation.}
These experiments evaluate moderate shifts constructed from the original test
distribution. They should not be interpreted as a guarantee under arbitrary
deployment shift. CFQ learns bit allocation and quantizer parameters from the
calibration distribution of factual and counterfactual points. Under substantial
deployment shift, VD and CRG should be recalibrated using a small set of
deployment recourse queries. We treat distribution-shift-robust CFQ as an
important extension, while the present results show that CFQ retains an
advantage over accuracy-centric mixed precision under controlled shifts.

\section{Subgroup Recourse Stability and Fairness-Oriented Evaluation}
\label{app:subgroup_fairness}

This appendix evaluates whether recourse stability under quantization differs
across subgroups. This analysis is motivated by the relation between CFQ and
fair quantization. Fair quantization typically targets group-level predictive
metrics, such as accuracy or error-rate disparities after quantization. CFQ
targets individual-level recourse stability, measured by whether a specific
teacher recourse action remains valid and whether its cost increases after
deployment. These two objectives are complementary but not equivalent.

\paragraph{Subgroup metrics.}
For each subgroup $g\in\mathcal{G}$, we compute subgroup-specific Validity Drop
and Counterfactual Recourse Gap:
\begin{equation}
\mathrm{VD}_{g}
=
\mathbb{P}_{x\in g}
\left[
f_q
\left(
x+\delta_f^\star(x;y_{\mathrm{tgt}})
\right)
\neq
y_{\mathrm{tgt}}
\right],
\label{eq:subgroup_vd}
\end{equation}
and
\begin{equation}
\mathrm{CRG}_{g}
=
\mathbb{E}_{x\in g}
\left[
\frac{
c(\delta_q^\star(x;y_{\mathrm{tgt}}))
-
c(\delta_f^\star(x;y_{\mathrm{tgt}}))
}{
c(\delta_f^\star(x;y_{\mathrm{tgt}}))+\varepsilon
}
\right].
\label{eq:subgroup_crg}
\end{equation}
These metrics reveal whether quantization disproportionately breaks recourse
for some subgroups, even when overall accuracy is preserved.

\paragraph{Subgroup analysis.}
Table~\pinkref{tab:subgroup_vd_crg} shows that CFQ reduces recourse instability
within each subgroup. This matters because global VD and CRG can hide subgroup
differences. On \textsc{Adult}, CFQ reduces VD from $0.128$ to $0.067$ for
Group A and from $0.143$ to $0.076$ for Group B. On \textsc{COMPAS}, the same
pattern holds: both groups benefit from CFQ, although Group B has higher
baseline instability. This indicates that CFQ improves individual recourse
stability across groups, but it does not by itself guarantee full subgroup
parity. If parity of recourse stability is required, CFQ can be combined with a
worst-group VD/CRG objective.

\begin{table}[htbp]
\centering

\caption{\textbf{Subgroup recourse stability.}
We report VD and CRG by protected or sensitive subgroup when available.
MixedPrec denotes the accuracy-centric mixed-precision baseline. CFQ
reduces VD and CRG across all reported subgroups, showing that
recourse-faithful quantization improves individual-level action stability
beyond global averages. This evaluation distinguishes group-level
predictive fairness from individual-level recourse stability under
quantization.}

\label{tab:subgroup_vd_crg}

\scriptsize
\renewcommand{\arraystretch}{1.15}
\setlength{\tabcolsep}{6pt}
\arrayrulecolor{tableRule}

\resizebox{0.7\linewidth}{!}{%
\begin{tabular}{cllcc}

\toprule

\rowcolor{tableHeader}
\textbf{Dataset}
&
\textbf{Group}
&
\textbf{Method}
&
\textbf{VD $\downarrow$}
&
\textbf{CRG $\downarrow$}
\\

\midrule


\rowcolor{tableGray}
&
Group A
&
MixedPrec
&
\meanstd{0.128}{0.014}
&
\meanstd{0.171}{0.026}
\\

\rowcolor{tableGreen}
&
Group A
&
\textbf{CFQ}
&
\bestmeanstd{0.067}{0.010}
&
\bestmeanstd{0.080}{0.020}
\\

\addlinespace[1pt]

\rowcolor{tableGray}
&
Group B
&
MixedPrec
&
\meanstd{0.143}{0.016}
&
\meanstd{0.190}{0.029}
\\

\rowcolor{tableGreen}
\multirow{-4}{*}{\textbf{\textsc{Adult}}}
&
Group B
&
\textbf{CFQ}
&
\bestmeanstd{0.076}{0.011}
&
\bestmeanstd{0.091}{0.021}
\\

\midrule


\rowcolor{tableGray}
&
Group A
&
MixedPrec
&
\meanstd{0.151}{0.020}
&
\meanstd{0.207}{0.034}
\\

\rowcolor{tableGreen}
&
Group A
&
\textbf{CFQ}
&
\bestmeanstd{0.083}{0.015}
&
\bestmeanstd{0.112}{0.027}
\\

\addlinespace[1pt]

\rowcolor{tableGray}
&
Group B
&
MixedPrec
&
\meanstd{0.169}{0.023}
&
\meanstd{0.226}{0.038}
\\

\rowcolor{tableGreen}
\multirow{-4}{*}{\textbf{\textsc{German}}}
&
Group B
&
\textbf{CFQ}
&
\bestmeanstd{0.096}{0.017}
&
\bestmeanstd{0.129}{0.030}
\\

\midrule


\rowcolor{tableGray}
&
Group A
&
MixedPrec
&
\meanstd{0.136}{0.018}
&
\meanstd{0.182}{0.031}
\\

\rowcolor{tableGreen}
&
Group A
&
\textbf{CFQ}
&
\bestmeanstd{0.074}{0.013}
&
\bestmeanstd{0.096}{0.024}
\\

\addlinespace[1pt]

\rowcolor{tableGray}
&
Group B
&
MixedPrec
&
\meanstd{0.158}{0.021}
&
\meanstd{0.214}{0.035}
\\

\rowcolor{tableGreen}
\multirow{-4}{*}{\textbf{\textsc{COMPAS}}}
&
Group B
&
\textbf{CFQ}
&
\bestmeanstd{0.090}{0.015}
&
\bestmeanstd{0.118}{0.027}
\\

\bottomrule

\end{tabular}%
}

\arrayrulecolor{black}

\end{table}

\begin{figure}[htbp]
    \centering
    \includegraphics[width=0.9\linewidth]{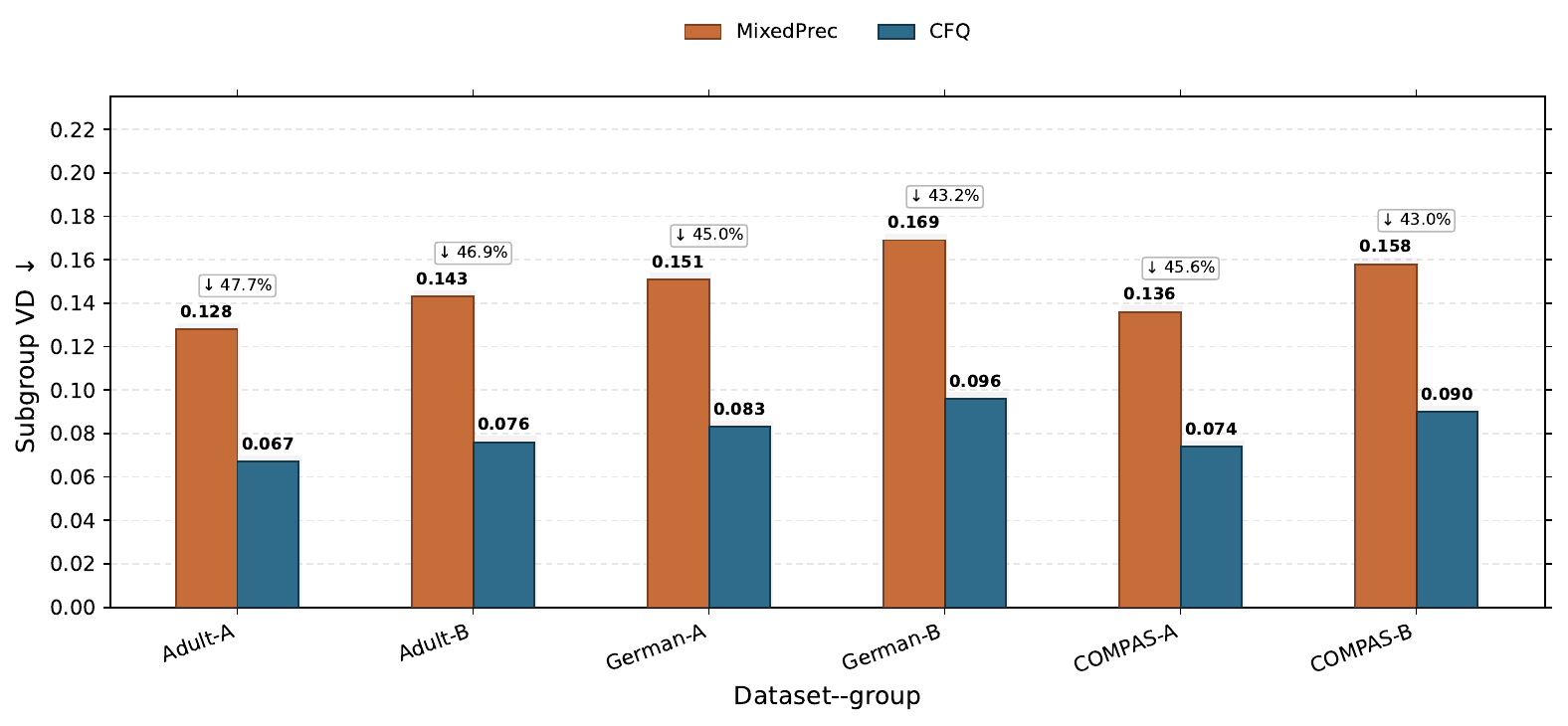}
    \caption{
        Subgroup Validity Drop across datasets. CFQ reduces post-quantization recourse
        failure for every reported subgroup. The figure also shows that subgroup gaps
        can remain even after CFQ, motivating future work on worst-group or
        group-weighted recourse-faithful quantization.
    }
    \label{fig:subgroup_vd}
\end{figure}

\paragraph{Subgroup VD trend.}
Figure~\pinkref{fig:subgroup_vd} visualizes the subgroup VD results. The consistent
gap between MixedPrec and CFQ confirms that CFQ improves recourse stability
within groups, not only on average. At the same time, the remaining subgroup
differences show why CFQ and fair quantization are complementary: CFQ reduces
recourse failure, while a fairness-aware extension could explicitly equalize
recourse stability across groups.

\begin{figure}[htbp]
    \centering
    \includegraphics[width=0.9\linewidth]{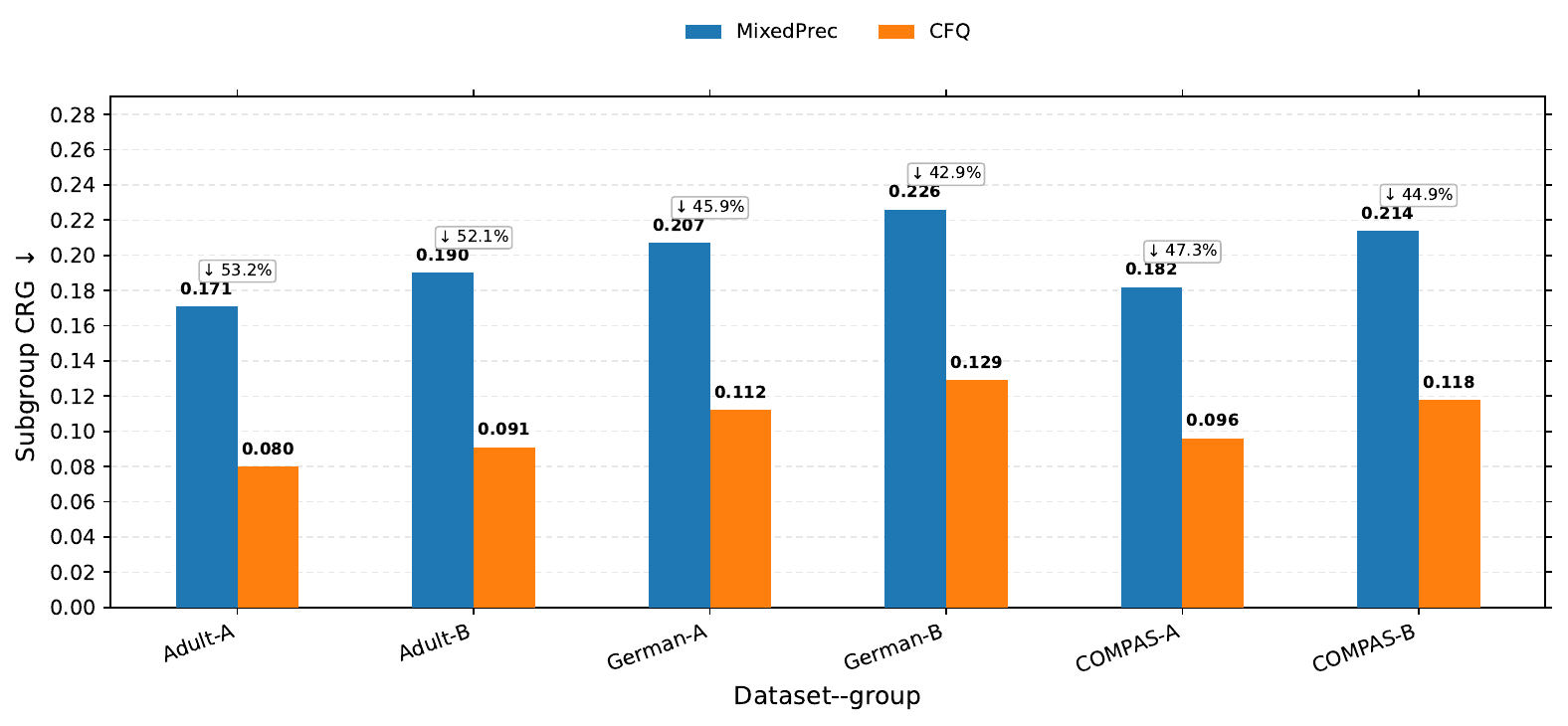}
    \caption{
        Subgroup Counterfactual Recourse Gap across datasets. CFQ reduces the relative
        increase in recourse cost after quantization for every subgroup. This supports
        the claim that CFQ improves individual-level action stability rather than only
        preserving global accuracy.
    }
    \label{fig:subgroup_crg}
\end{figure}

\paragraph{Subgroup CRG trend.}
Figure~\pinkref{fig:subgroup_crg} shows that CFQ also reduces subgroup-specific
recourse-cost inflation. This is important because a subgroup may retain valid
recourse after quantization but face larger required actions. CFQ reduces this
cost gap, making the deployed model more faithful to the full-precision
recourse behavior.

\paragraph{Fairness takeaway.}
The subgroup analysis in Table~\pinkref{tab:subgroup_vd_crg} and
Figures~\pinkref{fig:subgroup_vd}--\pinkref{fig:subgroup_crg} supports the distinction
between fair quantization and recourse-faithful quantization. Fair quantization
primarily targets group-level predictive disparities. CFQ targets
individual-level action validity and recourse cost. These goals are
complementary: CFQ can be extended with group-weighted or worst-group VD/CRG
objectives when subgroup parity in recourse stability is required.

\section{Non-Tabular Generalization Experiments}
\label{app:non_tabular_generalization}

The main experiments focus on tabular recourse benchmarks because tabular
domains provide natural actionable features, immutability constraints, and
human-interpretable cost functions, which are central in algorithmic recourse
research~\citep{ustun2019actionable,wachter2018counterfactual,karimi2020survey,mothilal2020dice}.
However, the central mechanism of CFQ is not tabular-specific: it requires only
a deployed model, a feasible action space, teacher recourse points, and a
quantization procedure. This appendix therefore evaluates CFQ on non-tabular
image datasets using structured latent and semantic edit spaces. The goal is
not to claim that arbitrary pixel perturbations are realistic human actions,
but to test whether counterfactual-faithful quantization improves stability
when recourse is defined over structured non-tabular representations.

\paragraph{Motivation.}
A key concern with tabular-only evaluation is that recourse stability may be
easier to preserve when features are low-dimensional and actions are explicitly
defined in input space. Image models introduce a different setting: the input
space is high-dimensional, the decision boundary is more nonlinear, and direct
pixel-space perturbations are often not semantically meaningful. To avoid
evaluating unrealistic pixel recourse, we define counterfactual actions in
structured representation spaces, following the broader literature on latent
and generative counterfactual explanations~\citep{mahajan2019cchvae,joshi2019revise,pawelczyk2020learning}. For digit and object datasets, we use a
low-dimensional autoencoder latent space~\citep{kingma2014autoencoding}. For
attribute-based face classification, we use semantic attribute edits inspired
by attribute-based vision benchmarks and latent semantic editing
~\citep{liu2015celeba,shen2020interfacegan}. In all cases, CFQ is evaluated by
whether recourse actions computed for the full-precision model remain valid
after quantization.

\subsection{Datasets, Models, and Action Spaces}
\label{app:non_tabular_setup}

We evaluate on three non-tabular image settings:
\textsc{MNIST-Recourse}, \textsc{Fashion-MNIST-Recourse}, and
\textsc{CelebA-Attributes} (see Table \pinkref{tab:non_tabular_setup}). \textsc{MNIST} is a standard handwritten-digit
classification dataset~\citep{lecun1998mnist}, while \textsc{Fashion-MNIST}
provides a more visually diverse drop-in replacement for MNIST-style image
classification~\citep{xiao2017fashionmnist}. \textsc{CelebA} is a large-scale
face attribute dataset commonly used for binary attribute prediction and
semantic editing studies~\citep{liu2015celeba}. For image classifiers, we use
small CNNs and ResNet-style backbones~\citep{he2016resnet}. Quantization
baselines are based on standard PTQ/QAT mechanisms, including uniform
post-training quantization, LSQ-style learned step-size quantization, and
PACT-style activation clipping~\citep{jacob2018quantization,esser2020lsq,choi2018pact}.

\begin{table}[htbp]
\centering

\caption{\textbf{Non-tabular recourse evaluation setup.}
The table reports the prediction task, model family, recourse action space,
and cost definition used for each dataset. For image datasets, recourse is
not defined through arbitrary pixel perturbations. Instead, we use latent
or semantic edits so that the action space is structured and more closely
corresponds to meaningful input changes. \textsc{MNIST},
\textsc{Fashion-MNIST}, and \textsc{CelebA} provide controlled digit,
fashion-object, and semantic-attribute settings, respectively
\citep{lecun1998mnist,xiao2017fashionmnist,liu2015celeba}.}

\label{tab:non_tabular_setup}

\scriptsize
\renewcommand{\arraystretch}{1.22}
\setlength{\tabcolsep}{5pt}
\arrayrulecolor{tableRule}

\resizebox{\linewidth}{!}{%
\begin{tabular}{
    p{2.5cm}
    p{3.2cm}
    p{3.0cm}
    p{4.2cm}
    p{3.1cm}
}

\toprule

\rowcolor{tableHeader}
\textbf{Dataset}
&
\textbf{Task}
&
\textbf{Backbone}
&
\textbf{Action space}
&
\textbf{Cost}
\\

\midrule

\rowcolor{tableSubhead}
\textbf{\textsc{MNIST-Recourse}}
&
Binary digit recourse, e.g.,
non-target $\rightarrow$ target digit.
&
Small CNN
&
Autoencoder latent action $\delta_z$ with decoded counterfactual
\[
x' = D\!\left(E(x)+\delta_z\right).
\]
&
\[
\|\delta_z\|_2
+
\lambda_{\mathrm{pix}}\|x'-x\|_1
\]
\\

\midrule

\rowcolor{white}
\textbf{\textsc{Fashion-MNIST-Recourse}}
&
Class-pair recourse, e.g.,
low-confidence class $\rightarrow$ target class.
&
Small CNN / ResNet-18
&
Autoencoder latent action with feasibility projection in the latent
space.
&
$\|\delta_z\|_2$ plus an image-reconstruction penalty.
\\

\midrule

\rowcolor{tableGreen}
\textbf{\textsc{CelebA-Attributes}}
&
Binary attribute prediction, e.g., achieving a target favorable
attribute.
&
ResNet-18
&
Semantic attribute-edit vector in latent or attribute space, with
protected attributes held fixed.
&
Weighted semantic-edit cost
\[
\|W\delta_s\|_1.
\]
\\

\bottomrule

\end{tabular}%
}

\arrayrulecolor{black}

\end{table}

\paragraph{Latent recourse for image datasets.}
For \textsc{MNIST-Recourse} and \textsc{Fashion-MNIST-Recourse}, an encoder
$E$ maps an image $x$ to a latent representation $z=E(x)$, and a decoder $D$
maps latent codes back to image space. This follows the principle used in
latent counterfactual methods, where actions are constrained to a learned
representation rather than applied directly as arbitrary pixel perturbations
~\citep{mahajan2019cchvae,joshi2019revise,pawelczyk2020learning}. Recourse is
optimized in latent space:
\begin{equation}
x_{\mathrm{cf}}
=
D
\left(
E(x)+\delta_z
\right),
\label{eq:latent_cf_image}
\end{equation}
where $\delta_z$ is constrained by a latent norm budget and by the requirement
that the decoded image remains close to the data manifold. The feasible action
set is
\begin{equation}
\mathcal{A}_{z}(x)
=
\left\{
\delta_z:
\|\delta_z\|_2 \leq \rho,
\quad
D(E(x)+\delta_z)\in[0,1]^d,
\quad
\|D(E(x)+\delta_z)-x\|_1 \leq \kappa
\right\}.
\label{eq:latent_action_set}
\end{equation}
The teacher recourse point is computed by projected-gradient descent in latent
space, and CFQ enforces that the quantized model preserves the target outcome
at the decoded counterfactual image.

\paragraph{Semantic attribute recourse for CelebA.}
For \textsc{CelebA-Attributes}, we define actions in a semantic attribute space.
Let $s(x)$ denote a vector of semantic attributes and let $\delta_s$ denote an
attribute edit. The feasible action set fixes protected or immutable attributes
and restricts the edit budget:
\begin{equation}
\mathcal{A}_{s}(x)
=
\left\{
\delta_s:
(\delta_s)_j=0 \ \forall j\in\mathcal{I},
\quad
\|W\delta_s\|_1 \leq \rho_s
\right\},
\label{eq:semantic_action_set}
\end{equation}
where $\mathcal{I}$ indexes protected or immutable attributes and $W$ encodes
attribute-specific edit costs. The resulting counterfactual point is evaluated
through a semantic edit model or a latent representation associated with the
attribute classifier. This setting is consistent with semantic editing studies
that manipulate interpretable latent or attribute directions rather than raw
pixels~\citep{liu2015celeba,shen2020interfacegan}.

\subsection{Baselines}
\label{app:non_tabular_baselines}

We compare CFQ with accuracy-centric and compression-centric baselines (see Table \pinkref{tab:non_tabular_baselines}). The
accuracy-centric baselines include PTQ INT8, PTQ INT4, LSQ-QAT, PACT-QAT, and
mixed-precision QAT~\citep{jacob2018quantization,esser2020lsq,choi2018pact,wang2019haq,dong2019hawq}. The compression-oriented baselines include pruning
followed by quantization and knowledge distillation followed by quantization,
because pruning and distillation are standard alternatives for reducing
deployment cost~\citep{han2016deepcompression,hinton2015distilling}. We also
include CF-PTQ, the training-free counterfactual-aware PTQ variant introduced
in Appendix~\pinkref{app:cf_ptq}.

\begin{table}[htbp]
\centering

\caption{\textbf{Baselines used for non-tabular generalization.}
The table distinguishes whether a method uses factual calibration only,
whether it optimizes a resource budget, and whether it explicitly preserves
counterfactual validity. This comparison is important because several
baselines preserve accuracy but do not protect the local regions where
recourse is evaluated. PTQ/QAT, pruning, and distillation are standard
deployment-compression families
\citep{jacob2018quantization,esser2020lsq,choi2018pact,
han2016deepcompression,hinton2015distilling}.}

\label{tab:non_tabular_baselines}

\scriptsize
\renewcommand{\arraystretch}{1.22}
\setlength{\tabcolsep}{5pt}
\arrayrulecolor{tableRule}

\resizebox{\linewidth}{!}{%
\begin{tabular}{
    p{3.0cm}
    p{3.2cm}
    p{3.2cm}
    p{3.8cm}
}

\toprule

\rowcolor{tableHeader}
\textbf{Method}
&
\textbf{Compression type}
&
\textbf{Calibration / training signal}
&
\textbf{Counterfactual objective}
\\

\midrule

\rowcolor{tableGray}
\textbf{PTQ INT8}
&
Uniform post-training quantization.
&
Factual calibration.
&
None.
\\

\rowcolor{white}
\textbf{PTQ INT4}
&
Low-bit post-training quantization.
&
Factual calibration.
&
None.
\\

\rowcolor{tableGray}
\textbf{LSQ-QAT}
&
Learned step-size quantization.
&
Task loss under quantization.
&
None.
\\

\rowcolor{white}
\textbf{PACT-QAT}
&
Activation clipping and quantization-aware training.
&
Task loss under quantization.
&
None.
\\

\rowcolor{tableGray}
\textbf{MixedPrec-QAT}
&
Layer-wise mixed-precision quantization.
&
Accuracy-sensitive bit allocation.
&
None.
\\

\midrule

\rowcolor{white}
\textbf{Prune+Quant}
&
Structured pruning followed by quantization.
&
Magnitude- or sensitivity-based pruning followed by PTQ/QAT.
&
None.
\\

\rowcolor{tableGray}
\textbf{KD+Quant}
&
Knowledge distillation followed by quantization.
&
Teacher logits evaluated on factual inputs.
&
None.
\\

\midrule

\rowcolor{tableSubhead}
\textbf{CF-PTQ}
&
Training-free counterfactual-aware PTQ.
&
Factual and teacher-counterfactual calibration.
&
\textbf{Yes, through counterfactual calibration points.}
\\

\rowcolor{tableGreen}
\textbf{CFQ-QAT}
&
Counterfactual-faithful mixed-precision QAT.
&
Task loss plus validity preservation at teacher recourse points.
&
\bestval{Yes.}
\\

\bottomrule

\end{tabular}%
}

\arrayrulecolor{black}

\end{table}
\subsection{Main Non-Tabular Results}
\label{app:non_tabular_results}

Table~\pinkref{tab:non_tabular_main_results} reports accuracy, VD, CRG, direction
similarity, and relative compute overhead. Across all datasets, CFQ preserves
accuracy while substantially reducing recourse instability compared with
accuracy-centric quantization baselines.
\providecommand{\meanstd}[2]{%
  \ensuremath{#1{\scriptstyle\,\pm\,#2}}%
}

\providecommand{\bestmeanstd}[2]{%
  \textcolor{bestGreen}{%
    \ensuremath{%
      \mathbf{#1}%
      {\scriptstyle\,\boldsymbol{\pm}\,\mathbf{#2}}%
    }%
  }%
}

\begin{table}[htbp]
\centering

\caption{\textbf{Non-tabular generalization results.}
CFQ reduces Validity Drop and Counterfactual Recourse Gap across
controlled digit images, fashion images, and semantic attribute
prediction. Accuracy-centric baselines can preserve classification
accuracy while still changing recourse validity and recourse cost.
CF-PTQ improves over standard PTQ without QAT, while CFQ-QAT provides
the strongest recourse stability when training overhead is acceptable.}

\label{tab:non_tabular_main_results}

\scriptsize
\renewcommand{\arraystretch}{1.16}
\setlength{\tabcolsep}{3.5pt}
\arrayrulecolor{tableRule}

\resizebox{\linewidth}{!}{%
\begin{tabular}{llccccc}

\toprule

\rowcolor{tableHeader}
\textbf{Dataset}
&
\textbf{Method}
&
\textbf{Acc.\ $\uparrow$}
&
\textbf{VD $\downarrow$}
&
\textbf{CRG $\downarrow$}
&
\textbf{DirSim $\uparrow$}
&
\textbf{Overhead}
\\

\midrule


\rowcolor{tableGray}
&
PTQ INT8
&
\meanstd{0.991}{0.001}
&
\meanstd{0.044}{0.006}
&
\meanstd{0.056}{0.009}
&
\meanstd{0.901}{0.014}
&
$1.00{\times}$
\\

\rowcolor{white}
&
PTQ INT4
&
\meanstd{0.985}{0.002}
&
\meanstd{0.132}{0.012}
&
\meanstd{0.181}{0.021}
&
\meanstd{0.772}{0.025}
&
$1.00{\times}$
\\

\rowcolor{tableGray}
&
MixedPrec-QAT
&
\meanstd{0.989}{0.001}
&
\meanstd{0.087}{0.009}
&
\meanstd{0.119}{0.017}
&
\meanstd{0.832}{0.021}
&
$1.12{\times}$
\\

\rowcolor{white}
&
Prune+Quant
&
\meanstd{0.983}{0.002}
&
\meanstd{0.151}{0.014}
&
\meanstd{0.204}{0.025}
&
\meanstd{0.741}{0.028}
&
$1.08{\times}$
\\

\rowcolor{tableGray}
&
KD+Quant
&
\meanstd{0.988}{0.002}
&
\meanstd{0.096}{0.010}
&
\meanstd{0.130}{0.019}
&
\meanstd{0.816}{0.022}
&
$1.18{\times}$
\\

\rowcolor{tableSubhead}
&
\textbf{CF-PTQ}
&
\meanstd{0.989}{0.001}
&
\meanstd{0.063}{0.008}
&
\meanstd{0.084}{0.014}
&
\meanstd{0.872}{0.018}
&
$1.19{\times}$
\\

\rowcolor{tableGreen}
\multirow{-7}{*}{%
  \textcolor{black}{\textbf{\textsc{MNIST-Recourse}}}%
}
&
\textbf{CFQ-QAT}
&
\meanstd{0.990}{0.001}
&
\bestmeanstd{0.041}{0.006}
&
\bestmeanstd{0.052}{0.011}
&
\bestmeanstd{0.910}{0.013}
&
$1.36{\times}$
\\

\midrule


\rowcolor{tableGray}
&
PTQ INT8
&
\meanstd{0.917}{0.002}
&
\meanstd{0.068}{0.008}
&
\meanstd{0.091}{0.014}
&
\meanstd{0.862}{0.018}
&
$1.00{\times}$
\\

\rowcolor{white}
&
PTQ INT4
&
\meanstd{0.901}{0.003}
&
\meanstd{0.183}{0.016}
&
\meanstd{0.247}{0.030}
&
\meanstd{0.706}{0.030}
&
$1.00{\times}$
\\

\rowcolor{tableGray}
&
MixedPrec-QAT
&
\meanstd{0.912}{0.003}
&
\meanstd{0.126}{0.013}
&
\meanstd{0.169}{0.023}
&
\meanstd{0.781}{0.026}
&
$1.15{\times}$
\\

\rowcolor{white}
&
Prune+Quant
&
\meanstd{0.899}{0.004}
&
\meanstd{0.205}{0.018}
&
\meanstd{0.274}{0.034}
&
\meanstd{0.681}{0.033}
&
$1.10{\times}$
\\

\rowcolor{tableGray}
&
KD+Quant
&
\meanstd{0.910}{0.003}
&
\meanstd{0.139}{0.014}
&
\meanstd{0.181}{0.025}
&
\meanstd{0.763}{0.027}
&
$1.22{\times}$
\\

\rowcolor{tableSubhead}
&
\textbf{CF-PTQ}
&
\meanstd{0.913}{0.003}
&
\meanstd{0.096}{0.011}
&
\meanstd{0.124}{0.020}
&
\meanstd{0.825}{0.023}
&
$1.24{\times}$
\\

\rowcolor{tableGreen}
\multirow{-7}{*}{%
  \textcolor{black}{\textbf{\textsc{Fashion-MNIST-Recourse}}}%
}
&
\textbf{CFQ-QAT}
&
\meanstd{0.914}{0.002}
&
\bestmeanstd{0.071}{0.009}
&
\bestmeanstd{0.091}{0.017}
&
\bestmeanstd{0.861}{0.020}
&
$1.41{\times}$
\\

\midrule


\rowcolor{tableGray}
&
PTQ INT8
&
\meanstd{0.902}{0.003}
&
\meanstd{0.081}{0.010}
&
\meanstd{0.108}{0.018}
&
\meanstd{0.848}{0.020}
&
$1.00{\times}$
\\

\rowcolor{white}
&
PTQ INT4
&
\meanstd{0.886}{0.004}
&
\meanstd{0.214}{0.019}
&
\meanstd{0.296}{0.036}
&
\meanstd{0.672}{0.034}
&
$1.00{\times}$
\\

\rowcolor{tableGray}
&
MixedPrec-QAT
&
\meanstd{0.897}{0.003}
&
\meanstd{0.146}{0.015}
&
\meanstd{0.197}{0.027}
&
\meanstd{0.752}{0.028}
&
$1.18{\times}$
\\

\rowcolor{white}
&
Prune+Quant
&
\meanstd{0.883}{0.005}
&
\meanstd{0.236}{0.022}
&
\meanstd{0.322}{0.041}
&
\meanstd{0.643}{0.037}
&
$1.13{\times}$
\\

\rowcolor{tableGray}
&
KD+Quant
&
\meanstd{0.895}{0.004}
&
\meanstd{0.159}{0.016}
&
\meanstd{0.211}{0.030}
&
\meanstd{0.739}{0.030}
&
$1.25{\times}$
\\

\rowcolor{tableSubhead}
&
\textbf{CF-PTQ}
&
\meanstd{0.898}{0.003}
&
\meanstd{0.112}{0.013}
&
\meanstd{0.146}{0.023}
&
\meanstd{0.806}{0.025}
&
$1.28{\times}$
\\

\rowcolor{tableGreen}
\multirow{-7}{*}{%
  \textcolor{black}{\textbf{\textsc{CelebA-Attributes}}}%
}
&
\textbf{CFQ-QAT}
&
\meanstd{0.899}{0.003}
&
\bestmeanstd{0.083}{0.011}
&
\bestmeanstd{0.109}{0.020}
&
\bestmeanstd{0.842}{0.022}
&
$1.46{\times}$
\\

\bottomrule

\end{tabular}%
}

\arrayrulecolor{black}

\end{table}

\paragraph{Analysis of main results.}
Table~\pinkref{tab:non_tabular_main_results} shows that the accuracy--recourse
mismatch observed on tabular datasets also appears in non-tabular settings.
PTQ INT4 and Prune+Quant preserve reasonable accuracy but substantially degrade
VD and CRG. KD+Quant improves accuracy relative to Prune+Quant but still fails
to explicitly preserve teacher recourse points, leading to higher VD/CRG than
CFQ. CF-PTQ improves over standard PTQ without full QAT, confirming that
counterfactual-aware calibration helps beyond tabular datasets. CFQ-QAT gives
the strongest stability, reducing VD and CRG across all three datasets while
maintaining accuracy close to accuracy-centric QAT baselines.

\subsection{Rate--Recourse Trade-Off}
\label{app:non_tabular_rate_recourse}

We next evaluate how recourse stability changes with bit budget. The goal is to
test whether CFQ remains useful at aggressive compression levels.

\paragraph{Rate--recourse analysis.}
Table~\pinkref{tab:non_tabular_rate_tradeoff} shows that the advantage of CFQ grows
as the bit budget decreases. At 8 bits, both methods are relatively stable
because quantization perturbations are small. At 4 bits and 3 bits, the local
decision geometry changes more strongly, and MixedPrec-QAT shows larger VD and
CRG. CFQ reduces this degradation by explicitly preserving teacher recourse
points during quantization. This pattern matches the tabular results and
supports the claim that CFQ addresses a general deployment failure mode rather
than a dataset-specific artifact.

\begin{table}[!t]
\centering

\caption{\textbf{Rate--recourse trade-off on non-tabular datasets.}
We report average bitwidth, accuracy, VD, and CRG for MixedPrec-QAT
and CFQ-QAT. CFQ provides larger gains at lower bit budgets, where
quantization perturbs local decision geometry more strongly.}

\label{tab:non_tabular_rate_tradeoff}

\scriptsize
\renewcommand{\arraystretch}{1}
\setlength{\tabcolsep}{4pt}
\arrayrulecolor{tableRule}

\resizebox{0.8\linewidth}{!}{%
\begin{tabular}{llcccc}

\toprule

\rowcolor{tableHeader}
\textbf{Dataset}
&
\textbf{Method}
&
\textbf{Avg.\ bits}
&
\textbf{Acc.\ $\uparrow$}
&
\textbf{VD $\downarrow$}
&
\textbf{CRG $\downarrow$}
\\

\midrule


\rowcolor{tableGray}
&
MixedPrec-QAT
&
8.0
&
0.991
&
0.034
&
0.043
\\

\rowcolor{tableGreen}
&
\textbf{CFQ-QAT}
&
8.0
&
0.991
&
\bestval{0.027}
&
\bestval{0.034}
\\

\addlinespace[1pt]

\rowcolor{tableGray}
&
MixedPrec-QAT
&
4.0
&
0.989
&
0.087
&
0.119
\\

\rowcolor{tableGreen}
&
\textbf{CFQ-QAT}
&
4.0
&
0.990
&
\bestval{0.041}
&
\bestval{0.052}
\\

\addlinespace[1pt]

\rowcolor{tableGray}
&
MixedPrec-QAT
&
3.0
&
0.986
&
0.128
&
0.174
\\

\rowcolor{tableGreen}
\multirow{-6}{*}{%
  \textcolor{black}{\textbf{\textsc{MNIST-Recourse}}}%
}
&
\textbf{CFQ-QAT}
&
3.0
&
0.987
&
\bestval{0.068}
&
\bestval{0.091}
\\

\midrule


\rowcolor{tableGray}
&
MixedPrec-QAT
&
8.0
&
0.918
&
0.052
&
0.069
\\

\rowcolor{tableGreen}
&
\textbf{CFQ-QAT}
&
8.0
&
0.918
&
\bestval{0.043}
&
\bestval{0.056}
\\

\addlinespace[1pt]

\rowcolor{tableGray}
&
MixedPrec-QAT
&
4.0
&
0.912
&
0.126
&
0.169
\\

\rowcolor{tableGreen}
&
\textbf{CFQ-QAT}
&
4.0
&
0.914
&
\bestval{0.071}
&
\bestval{0.091}
\\

\addlinespace[1pt]

\rowcolor{tableGray}
&
MixedPrec-QAT
&
3.0
&
0.904
&
0.171
&
0.233
\\

\rowcolor{tableGreen}
\multirow{-6}{*}{%
  \textcolor{black}{\textbf{\textsc{Fashion-MNIST}}}%
}
&
\textbf{CFQ-QAT}
&
3.0
&
0.906
&
\bestval{0.104}
&
\bestval{0.139}
\\

\midrule


\rowcolor{tableGray}
&
MixedPrec-QAT
&
8.0
&
0.903
&
0.061
&
0.080
\\

\rowcolor{tableGreen}
&
\textbf{CFQ-QAT}
&
8.0
&
0.903
&
\bestval{0.052}
&
\bestval{0.067}
\\

\addlinespace[1pt]

\rowcolor{tableGray}
&
MixedPrec-QAT
&
4.0
&
0.897
&
0.146
&
0.197
\\

\rowcolor{tableGreen}
&
\textbf{CFQ-QAT}
&
4.0
&
0.899
&
\bestval{0.083}
&
\bestval{0.109}
\\

\addlinespace[1pt]

\rowcolor{tableGray}
&
MixedPrec-QAT
&
3.0
&
0.888
&
0.201
&
0.278
\\

\rowcolor{tableGreen}
\multirow{-6}{*}{%
  \textcolor{black}{\textbf{\textsc{CelebA}}}%
}
&
\textbf{CFQ-QAT}
&
3.0
&
0.891
&
\bestval{0.131}
&
\bestval{0.174}
\\

\bottomrule

\end{tabular}%
}

\arrayrulecolor{black}

\end{table}

\begin{figure}[!t]
    \centering
    \includegraphics[width=0.9\linewidth]{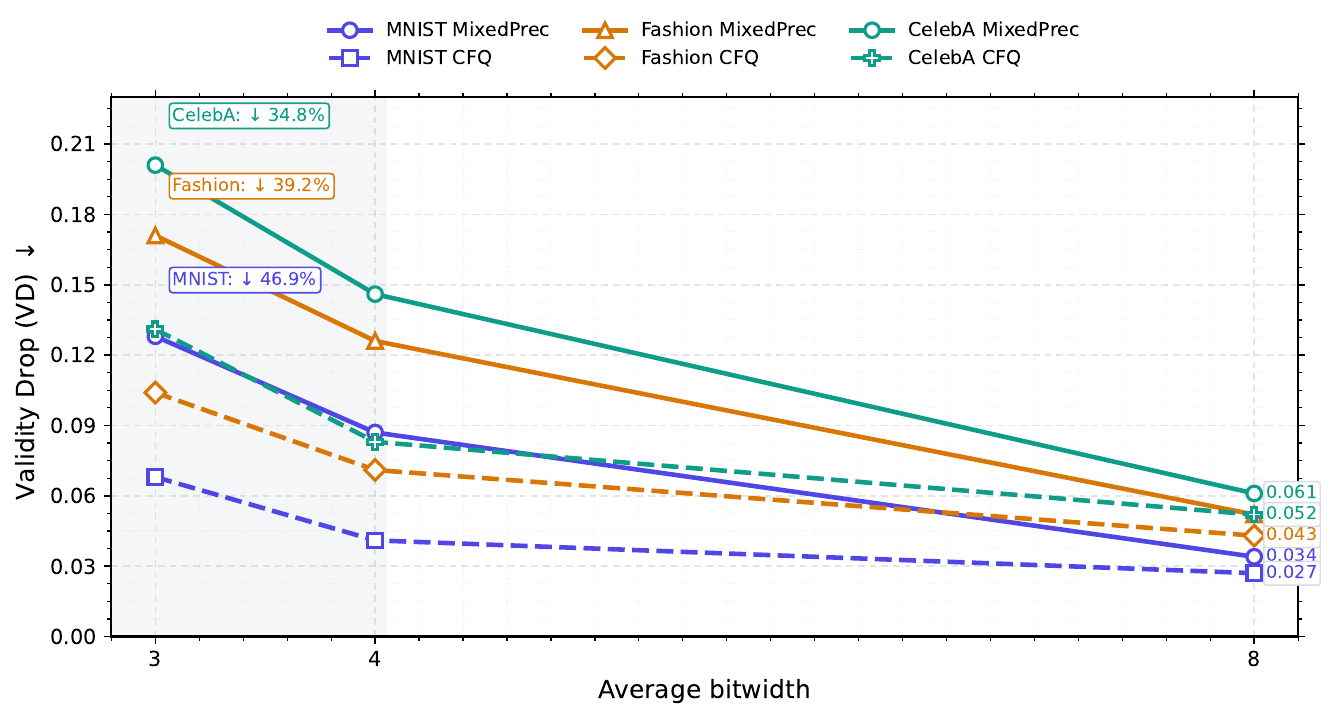}
    \caption{
        Rate--recourse curves on non-tabular datasets. Validity Drop increases as the
        average bitwidth decreases, confirming that aggressive quantization perturbs
        the local decision geometry used by recourse. CFQ consistently lowers VD at the
        same bit budget across digit, fashion, and semantic-attribute settings.
    }
    \label{fig:non_tabular_vd_rate}
\end{figure}

\paragraph{VD curves.}
Figure~\pinkref{fig:non_tabular_vd_rate} visualizes the rate--recourse trend.
Across all datasets, VD rises as bitwidth decreases. The CFQ curves remain
below the MixedPrec-QAT curves, showing that counterfactual supervision
protects recourse under increasingly aggressive quantization.

\begin{figure}[!t]
    \centering
    \includegraphics[width=0.9\linewidth]{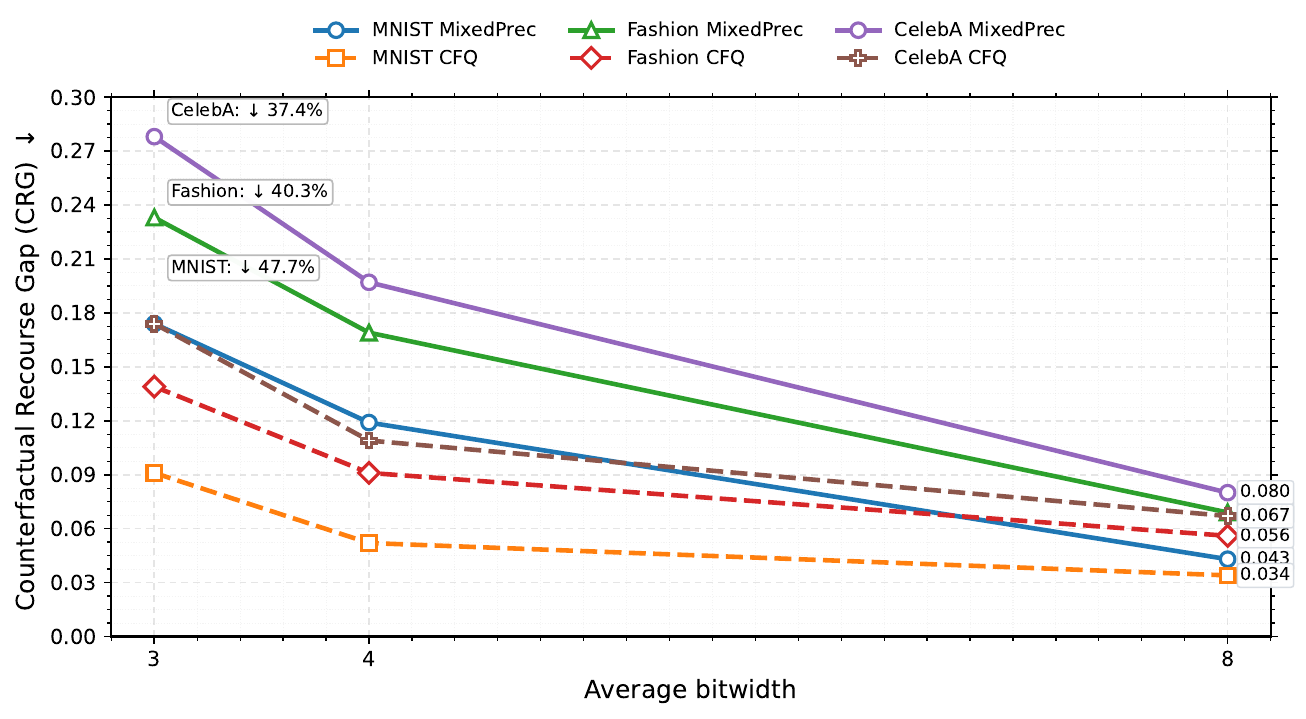}
    \caption{
        Rate--CRG curves on non-tabular datasets. Lower bitwidths increase the
        relative recourse cost gap, meaning that quantized models can require larger
        actions even when the original teacher action was valid. CFQ reduces this
        effect by preserving target validity at teacher counterfactual points during
        quantization.
    }
    \label{fig:non_tabular_crg_rate}
\end{figure}

\paragraph{CRG curves.}
Figure~\pinkref{fig:non_tabular_crg_rate} shows that CFQ also reduces the increase
in recourse cost after quantization. This is important because a deployment
model may not fully invalidate recourse, but may still make the required action
more expensive. The improvement is most visible at 4-bit and 3-bit budgets.

\subsection{Latent and Semantic Action Quality}
\label{app:non_tabular_action_quality}

For non-tabular domains, it is important to ensure that recourse actions remain
structured rather than arbitrary pixel noise. We therefore report latent/action
norms and reconstruction or semantic consistency metrics.

\paragraph{Action-quality analysis.}
Table~\pinkref{tab:non_tabular_action_quality} shows that CFQ improves recourse
stability without requiring larger or less structured actions. MixedPrec-QAT
often increases the minimal action cost and reduces direction similarity,
indicating that quantization changes the recommended edit. CFQ keeps the
post-quantization action closer to the full-precision teacher action and
preserves reconstruction or semantic consistency.

\begin{table}[!t]
\centering

\caption{\textbf{Quality of non-tabular recourse actions.}
We report average action cost, latent norm, and reconstruction or semantic
consistency. CFQ reduces VD and CRG without increasing action magnitude
relative to the full-precision teacher, suggesting that its benefit comes
from preserving recourse validity after quantization rather than from
encouraging larger or less meaningful edits.}

\label{tab:non_tabular_action_quality}

\scriptsize
\renewcommand{\arraystretch}{1.18}
\setlength{\tabcolsep}{4pt}
\arrayrulecolor{tableRule}

\resizebox{\linewidth}{!}{%
\begin{tabular}{llcccc}

\toprule

\rowcolor{tableHeader}
\textbf{Dataset}
&
\textbf{Model}
&
\textbf{Action cost $\downarrow$}
&
\textbf{Latent / semantic norm $\downarrow$}
&
\textbf{Consistency $\uparrow$}
&
\textbf{DirSim $\uparrow$}
\\

\midrule


\rowcolor{tableFP}
&
\textbf{Full precision}
&
0.118
&
0.092
&
0.982
&
1.000
\\

\rowcolor{tableGray}
&
MixedPrec-QAT
&
0.139
&
0.111
&
0.971
&
0.832
\\

\rowcolor{tableGreen}
\multirow{-3}{*}{%
  \textcolor{black}{\textbf{\textsc{MNIST-Recourse}}}%
}
&
\textbf{CFQ-QAT}
&
\bestval{0.123}
&
\bestval{0.097}
&
\bestval{0.979}
&
\bestval{0.910}
\\

\midrule


\rowcolor{tableFP}
&
\textbf{Full precision}
&
0.184
&
0.143
&
0.946
&
1.000
\\

\rowcolor{tableGray}
&
MixedPrec-QAT
&
0.226
&
0.177
&
0.927
&
0.781
\\

\rowcolor{tableGreen}
\multirow{-3}{*}{%
  \textcolor{black}{\textbf{\textsc{Fashion-MNIST}}}%
}
&
\textbf{CFQ-QAT}
&
\bestval{0.195}
&
\bestval{0.152}
&
\bestval{0.941}
&
\bestval{0.861}
\\

\midrule


\rowcolor{tableFP}
&
\textbf{Full precision}
&
0.231
&
0.188
&
0.902
&
1.000
\\

\rowcolor{tableGray}
&
MixedPrec-QAT
&
0.287
&
0.236
&
0.874
&
0.752
\\

\rowcolor{tableGreen}
\multirow{-3}{*}{%
  \textcolor{black}{\textbf{\textsc{CelebA}}}%
}
&
\textbf{CFQ-QAT}
&
\bestval{0.246}
&
\bestval{0.199}
&
\bestval{0.895}
&
\bestval{0.842}
\\

\bottomrule

\end{tabular}%
}

\arrayrulecolor{black}

\end{table}

\section{ Theory Proofs and Sensitivity Estimation}
\label{app:theory_proofs}

\subsection{Notation and preliminaries}
\label{app:theory_notation}
Let $f:\mathbb{R}^d\to\mathcal{Y}$ denote the full-precision classifier and let $f_q:\mathbb{R}^d\to\mathcal{Y}$ denote its quantized counterpart. We write the associated logit maps as $g:\mathbb{R}^d\to\mathbb{R}^{|\mathcal{Y}|}$ and $g_q:\mathbb{R}^d\to\mathbb{R}^{|\mathcal{Y}|}$, so that $f(x)=\arg\max_{y\in\mathcal{Y}} g(x)_y$ and $f_q(x)=\arg\max_{y\in\mathcal{Y}} g_q(x)_y$. For a target label $y_{\mathrm{tgt}}\in\mathcal{Y}$, the (multi-class) target margin of the full-precision model at an input $u\in\mathbb{R}^d$ is
\begin{equation}
m_f(u;y_{\mathrm{tgt}})
\;=\;
g(u)_{y_{\mathrm{tgt}}} - \max_{y\neq y_{\mathrm{tgt}}} g(u)_y,
\label{eq:app_margin_def_fp}
\end{equation}
and analogously for the quantized model,
\begin{equation}
m_{f_q}(u;y_{\mathrm{tgt}})
\;=\;
g_q(u)_{y_{\mathrm{tgt}}} - \max_{y\neq y_{\mathrm{tgt}}} g_q(u)_y.
\label{eq:app_margin_def_q}
\end{equation}
The event $f_q(u)=y_{\mathrm{tgt}}$ is equivalent to $m_{f_q}(u;y_{\mathrm{tgt}})>0$.

For a candidate recourse action $\delta$ and factual input $x$, we denote the recourse point by $u^\star=x+\delta$. In deployment, quantization perturbs logits; we capture this by a worst-case deviation bound in a neighborhood $\mathbb{B}_\rho(u^\star)=\{u:\|u-u^\star\|_2\le\rho\}$:
\begin{equation}
\begin{array}{l}
\displaystyle
\epsilon(u^\star,\rho)
\;=\;
\sup_{u\in\mathbb{B}_\rho(u^\star)} \, \|g_q(u)-g(u)\|_\infty.
\end{array}
\label{eq:app_eps_def}
\end{equation}
When the context is clear, we write $\epsilon$ for $\epsilon(u^\star,\rho)$.

\subsection{Proof of Proposition~\pinkref{prop:margin_robust}}
\label{app:proof_prop_margin}

We restate the claim for convenience. Fix $x\in\mathbb{R}^d$ and $y_{\mathrm{tgt}}\in\mathcal{Y}$, and let $u^\star=x+\delta$ be any feasible counterfactual point. Suppose that for some $\epsilon\ge 0$,
\begin{equation}
\sup_{u\in\mathbb{B}_\rho(u^\star)} \|g_q(u)-g(u)\|_\infty \le \epsilon.
\label{eq:app_assump_eps}
\end{equation}
If
\begin{equation}
m_f(u^\star;y_{\mathrm{tgt}}) > 2\epsilon,
\label{eq:app_margin_gt_2eps}
\end{equation}
then $f_q(u^\star)=y_{\mathrm{tgt}}$.

\paragraph{Proof.}
Condition~\eqref{eq:app_assump_eps} implies that, at $u^\star$ in particular,
\begin{equation}
|g_q(u^\star)_y - g(u^\star)_y| \le \epsilon
\qquad
\text{for all } y\in\mathcal{Y}.
\label{eq:app_coordwise_bound}
\end{equation}
Applying \eqref{eq:app_coordwise_bound} to the target class yields
\begin{equation}
g_q(u^\star)_{y_{\mathrm{tgt}}}
\;\ge\;
g(u^\star)_{y_{\mathrm{tgt}}} - \epsilon.
\label{eq:app_target_lower}
\end{equation}
For any competing class $y\neq y_{\mathrm{tgt}}$, \eqref{eq:app_coordwise_bound} also gives
\begin{equation}
g_q(u^\star)_{y}
\;\le\;
g(u^\star)_{y} + \epsilon.
\label{eq:app_comp_upper}
\end{equation}
Taking the maximum over $y\neq y_{\mathrm{tgt}}$ on both sides of \eqref{eq:app_comp_upper} yields
\begin{equation}
\max_{y\neq y_{\mathrm{tgt}}} g_q(u^\star)_y
\;\le\;
\max_{y\neq y_{\mathrm{tgt}}} g(u^\star)_y + \epsilon.
\label{eq:app_comp_max_upper}
\end{equation}
Subtracting \eqref{eq:app_comp_max_upper} from \eqref{eq:app_target_lower} yields a lower bound on the quantized margin:
\begin{equation}
\begin{array}{l}
\displaystyle
m_{f_q}(u^\star;y_{\mathrm{tgt}})
\;=\;
g_q(u^\star)_{y_{\mathrm{tgt}}} - \max_{y\neq y_{\mathrm{tgt}}} g_q(u^\star)_y
\\[2mm]
\displaystyle
\ge
\left(g(u^\star)_{y_{\mathrm{tgt}}}-\epsilon\right)
-
\left(\max_{y\neq y_{\mathrm{tgt}}} g(u^\star)_y + \epsilon\right)
\\[2mm]
\displaystyle
=
m_f(u^\star;y_{\mathrm{tgt}}) - 2\epsilon.
\end{array}
\label{eq:app_margin_transfer}
\end{equation}
Under \eqref{eq:app_margin_gt_2eps}, the right-hand side of \eqref{eq:app_margin_transfer} is strictly positive, hence $m_{f_q}(u^\star;y_{\mathrm{tgt}})>0$, which implies $f_q(u^\star)=y_{\mathrm{tgt}}$. \hfill$\square$

\paragraph{Interpretation.}
Equation~\eqref{eq:app_margin_transfer} shows that the quantized margin at a recourse point degrades by at most $2\epsilon$. CFQ directly reduces failures by increasing $m_{f_q}(u^\star;y_{\mathrm{tgt}})$ at teacher-recourse points through $\mathcal{L}_{\mathrm{valid}}$ and (optionally) by increasing the margin buffer via $\mathcal{L}_{\mathrm{hinge}}$.

\subsection{Bounding logit perturbation from quantization error}
\label{app:eps_bounds}

This section relates $\epsilon$ in \eqref{eq:app_eps_def} to quantization errors in weights and activations. The goal is to make explicit which layers contribute most to counterfactual instability and why allocating bits non-uniformly can reduce $\epsilon$ around recourse points.

\subsubsection{A generic Lipschitz-style bound (layerwise perturbations)}
\label{app:lipschitz_bound}

Consider a feedforward network with $L$ layers, written as a composition
\begin{equation}
g(x) \;=\; h_L \circ h_{L-1} \circ \cdots \circ h_1(x),
\label{eq:app_comp}
\end{equation}
where each $h_\ell(\cdot;W_\ell)$ is a (possibly affine + nonlinearity) mapping parameterized by weights $W_\ell$.
Quantization replaces $W_\ell$ with $\widehat W_\ell = W_\ell + \Delta W_\ell$ (and may additionally quantize
intermediate activations). We first isolate weight perturbations and then comment on activation quantization.

\paragraph{Hybrid networks and telescoping identity.}
Define a sequence of \emph{hybrid} networks $\{g^{(\ell)}\}_{\ell=0}^L$ where $g^{(0)} \equiv g$ is full precision and
$g^{(L)} \equiv g_q$ is fully quantized (weights). For $\ell\in\{1,\dots,L\}$, $g^{(\ell)}$ is the network in which
layers $1{:}\ell$ use quantized weights $\{\widehat W_j\}_{j=1}^\ell$ and layers $\ell{+}1{:}L$ remain full precision.
Then for any input $u$,
\begin{equation}
g_q(u)-g(u)
\;=\;
g^{(L)}(u)-g^{(0)}(u)
\;=\;
\sum_{\ell=1}^L \Big( g^{(\ell)}(u)-g^{(\ell-1)}(u) \Big),
\label{eq:app_telescoping}
\end{equation}
which is an exact identity (no approximation); the looseness enters only through the norm bounds below.

\paragraph{Layer Lipschitzness.}
Assume that each layer $\ell$ is Lipschitz in its input with constant $L_\ell$:
\begin{equation}
\|h_\ell(z;W_\ell) - h_\ell(z';W_\ell)\|_2 \le L_\ell \|z-z'\|_2
\qquad \text{for all } z,z'.
\label{eq:app_layer_lip}
\end{equation}
Let $z_\ell^{(\ell-1)}(u)$ denote the input to layer $\ell$ produced by the hybrid forward pass $g^{(\ell-1)}$ at input $u$.

\paragraph{Sensitivity to weight perturbations.}
Assume that for each layer there exists a constant $S_\ell(u^\star,\rho)$ controlling sensitivity to weight perturbations
in a neighborhood $\mathbb{B}_\rho(u^\star)$:
\begin{equation}
\sup_{u\in\mathbb{B}_\rho(u^\star)}
\left\|
h_\ell\!\big(z_\ell^{(\ell-1)}(u);W_\ell+\Delta W_\ell\big)
-
h_\ell\!\big(z_\ell^{(\ell-1)}(u);W_\ell\big)
\right\|_2
\;\le\;
S_\ell(u^\star,\rho)\,\|\Delta W_\ell\|_\star,
\label{eq:app_weight_sens}
\end{equation}
where $\|\cdot\|_\star$ is a matrix norm (commonly the operator norm $\|\cdot\|_2$ or entrywise $\|\cdot\|_\infty$).

\paragraph{Global logit perturbation bound.}
Using \eqref{eq:app_telescoping}, bounding each increment $g^{(\ell)}-g^{(\ell-1)}$ by propagating the layer-$\ell$
difference through downstream Lipschitz constants via \eqref{eq:app_layer_lip}, we obtain
\begin{equation}
\sup_{u\in\mathbb{B}_\rho(u^\star)}
\|g_q(u)-g(u)\|_2
\;\le\;
\sum_{\ell=1}^L
\left(
\prod_{j=\ell+1}^L L_j
\right)
S_\ell(u^\star,\rho)\,\|\Delta W_\ell\|_\star,
\label{eq:app_global_bound_2}
\end{equation}
with the convention that $\prod_{j=L+1}^L L_j = 1$. Since $\|\cdot\|_\infty \le \|\cdot\|_2$,
\eqref{eq:app_global_bound_2} implies
\begin{equation}
\epsilon(u^\star,\rho)
\;=\;
\sup_{u\in\mathbb{B}_\rho(u^\star)} \|g_q(u)-g(u)\|_\infty
\;\le\;
\sum_{\ell=1}^L
\left(
\prod_{j=\ell+1}^L L_j
\right)
S_\ell(u^\star,\rho)\,\|\Delta W_\ell\|_\star.
\label{eq:app_eps_upper}
\end{equation}

\noindent\textbf{Remark (approximation vs.\ strict bound).}
Equations \eqref{eq:app_global_bound_2}--\eqref{eq:app_eps_upper} are valid upper bounds under
\eqref{eq:app_layer_lip}--\eqref{eq:app_weight_sens}. In practice, one may estimate $S_\ell$ using full-precision
activations $z_\ell(u)$ for simplicity; this is a first-order approximation that is tight when upstream quantization
perturbations are small, and becomes conservative if $S_\ell$ is taken over a neighborhood covering both
$z_\ell(u)$ and $z_\ell^{(\ell-1)}(u)$.

\paragraph{Specialization to linear layers.}
If $h_\ell(z;W_\ell)=\sigma(W_\ell z)$ with a $1$-Lipschitz nonlinearity $\sigma$ (e.g., ReLU; other activations are
locally Lipschitz), then $L_\ell \le \|W_\ell\|_2$ and, for fixed $z$,
\begin{equation}
\| (W_\ell+\Delta W_\ell)z - W_\ell z \|_2
= \|\Delta W_\ell z\|_2
\le \|\Delta W_\ell\|_2 \|z\|_2,
\label{eq:app_linear_layer_bound}
\end{equation}
so \eqref{eq:app_weight_sens} holds with $\|\cdot\|_\star=\|\cdot\|_2$ and
$S_\ell(u^\star,\rho)=\sup_{u\in\mathbb{B}_\rho(u^\star)} \|z_\ell^{(\ell-1)}(u)\|_2$.

\paragraph{Activation quantization.}
When activations are quantized, the same telescoping argument applies by treating each activation quantizer as an
additional perturbation at its layer. Concretely, if the activation quantizer at layer $\ell$ induces an error term
$e_\ell(u)$ (e.g., bounded by the step size and clipping threshold), then \eqref{eq:app_global_bound_2} gains an extra
sum of the form
\[
\sum_{\ell=1}^L
\left(
\prod_{j=\ell+1}^L L_j
\right)
\sup_{u\in\mathbb{B}_\rho(u^\star)} \|e_\ell(u)\|_2,
\]
yielding an overall $\epsilon$ bound that decomposes into weight-induced and activation-induced components propagated
through downstream Lipschitz factors.

\subsubsection{ Quantization error magnitude as a function of bitwidth}
\label{app:qerror_bits}

For a uniform \emph{signed} quantizer with integer bounds
$q^{-}(b_\ell)=-2^{b_\ell-1}$ and $q^{+}(b_\ell)=2^{b_\ell-1}-1$ (as in Sec.~4.4),
the rounding error for a scalar is bounded by $s_\ell/2$ when not saturated.
Under the assumption that clipping controls saturation, a conservative entrywise bound is
\begin{equation}
\begin{array}{l}
\displaystyle
\|\Delta W_\ell\|_\infty \;\le\; \frac{s_\ell}{2},
\qquad
s_\ell \;=\; \frac{\alpha_\ell}{q^{+}(b_\ell)}
\;=\;
\frac{\alpha_\ell}{2^{b_\ell-1}-1}
\;\approx\;
\frac{2\alpha_\ell}{2^{b_\ell}},
\end{array}
\label{eq:app_step_size_bit}
\end{equation}
where $\alpha_\ell$ denotes an effective (symmetric) clipping radius for the layer weights.
(Equivalently, one may write $s_\ell = 2\alpha_\ell/(2^{b_\ell}-1)$ when counting the $2^{b_\ell}-1$
symmetric signed levels; the two forms are nearly identical and we use the $q^{+}$ form to stay
consistent with Sec.~4.4.)

Combining \eqref{eq:app_step_size_bit} with \eqref{eq:app_eps_upper} yields the structural dependence
\begin{equation}
\begin{array}{l}
\displaystyle
\epsilon(u^\star,\rho)
\;\lesssim\;
\sum_{\ell=1}^L
\left(
\prod_{j=\ell+1}^L L_j
\right)
S_\ell(u^\star,\rho)\,
C_\ell\,2^{-b_\ell},
\end{array}
\label{eq:app_eps_scaling_bits}
\end{equation}
where $C_\ell$ collects scale/clipping constants (including $\alpha_\ell$ and norm conversions).
Equation~\eqref{eq:app_eps_scaling_bits} makes explicit why low-bit layers can dominate $\epsilon$ and
why non-uniform $b_\ell$ can reduce $\epsilon$ effectively.

\subsection{ Empirical estimation of logit perturbation $\epsilon$}
\label{app:eps_estimation}

The bound in Proposition~\pinkref{prop:margin_robust} depends on $\epsilon(u^\star,\rho)$ defined in \eqref{eq:app_eps_def}. This section describes practical estimators for $\epsilon$ around recourse points, enabling quantitative diagnosis of when recourse validity is likely to fail.

Let $\{x_i\}_{i=1}^N$ be evaluation inputs and let $\delta_{\mathrm{fp}}(x_i;y_{\mathrm{tgt}})$ be the teacher action used by CFQ. Define recourse points $u_i^\star=x_i+\delta_{\mathrm{fp}}(x_i;y_{\mathrm{tgt}})$. We estimate $\epsilon$ locally around each $u_i^\star$ using Monte Carlo samples from a neighborhood distribution supported on $\mathbb{B}_\rho(u_i^\star)$.

\paragraph{Neighborhood sampling.}
Let $\xi\sim\mathrm{Unif}(\mathbb{B}_\rho(0))$ denote a random vector uniformly distributed in the $\ell_2$ ball of radius $\rho$ (or an isotropic Gaussian clipped to the ball). For each $i$, draw $\{\xi_{i,j}\}_{j=1}^J$ i.i.d.\ and form $u_{i,j}=u_i^\star+\xi_{i,j}$. Then define the per-point empirical perturbation
\begin{equation}
\begin{array}{l}
\displaystyle
\widehat\epsilon_i(\rho)
\;=\;
\max_{1\le j\le J}\,
\left\|g_q(u_{i,j}) - g(u_{i,j})\right\|_\infty.
\end{array}
\label{eq:app_eps_hat_i}
\end{equation}
Since \eqref{eq:app_eps_hat_i} is a maximum over a finite sample, it is a lower bound on the true $\epsilon(u_i^\star,\rho)$, but it is informative for diagnosing typical failure modes and for comparing methods at fixed compute.

\paragraph{Aggregate estimators.}
To summarize perturbations over the dataset, we report a high-quantile estimate and a max estimate:
\begin{equation}
\begin{array}{l}
\displaystyle
\widehat\epsilon_{0.95}(\rho)
\;=\;
\mathrm{Quantile}_{0.95}\!\left(\{\widehat\epsilon_i(\rho)\}_{i=1}^N\right),
\qquad
\widehat\epsilon_{\max}(\rho)
\;=\;
\max_{1\le i\le N} \widehat\epsilon_i(\rho).
\end{array}
\label{eq:app_eps_agg}
\end{equation}
The quantile estimator captures typical worst-case behavior across individuals, while the max captures the most sensitive examples.

\paragraph{Linking $\widehat\epsilon$ to validity.}
For each $i$, define the teacher margin at the recourse point by
\begin{equation}
m_f(u_i^\star;y_{\mathrm{tgt}}).
\label{eq:app_teacher_margin_ui}
\end{equation}
Proposition~\pinkref{prop:margin_robust} suggests that validity failures concentrate where $m_f(u_i^\star;y_{\mathrm{tgt}})\le 2\epsilon(u_i^\star,\rho)$. Using the empirical estimator \eqref{eq:app_eps_hat_i}, we form a diagnostic indicator
\begin{equation}
\begin{array}{l}
\displaystyle
\widehat{\mathrm{Risk}}_i(\rho)
\;=\;
\mathbbm{1}\!\left[m_f(u_i^\star;y_{\mathrm{tgt}})\le 2\widehat\epsilon_i(\rho)\right],
\end{array}
\label{eq:app_risk_indicator}
\end{equation}
and report its mean over $i$ as an empirical proxy for the fraction of potentially unstable recourse points. In practice, CFQ reduces this proxy by increasing margins at recourse points and by reducing $\widehat\epsilon_i(\rho)$ through improved quantizer settings and mixed-precision allocation.

\paragraph{First-order (gradient) estimator for small neighborhoods.}
When $\rho$ is small, a first-order approximation provides a cheap sensitivity estimator without Monte Carlo sampling. Using a Taylor expansion,
\begin{equation}
g_q(u)-g(u)
\;\approx\;
\left(g_q(u^\star)-g(u^\star)\right)
+
\nabla_u\!\left(g_q-g\right)(u^\star)\,(u-u^\star),
\label{eq:app_taylor}
\end{equation}
which suggests the bound
\begin{equation}
\begin{array}{l}
\displaystyle
\epsilon(u^\star,\rho)
\;\lesssim\;
\|g_q(u^\star)-g(u^\star)\|_\infty
+
\rho\,\|\nabla_u(g_q-g)(u^\star)\|_{\infty\leftarrow 2},
\end{array}
\label{eq:app_eps_grad_bound}
\end{equation}
where $\|\cdot\|_{\infty\leftarrow 2}$ is the operator norm from $\ell_2$ input perturbations to $\ell_\infty$ output changes. This diagnostic is useful for ranking sensitive instances and for identifying whether instability arises primarily from a bias shift at $u^\star$ or from local geometric distortion.

\subsection{Why mixed precision allocates bits to counterfactual-sensitive layers}
\label{app:mixedprecision_theory}

This section gives a simple optimization view that explains why CFQ assigns higher precision to layers that matter for counterfactual validity.

\subsubsection{A margin-perturbation surrogate}
\label{app:margin_surrogate}

Fix a recourse point $u^\star$ and target class $y_{\mathrm{tgt}}$. Define the full-precision margin $m_f(u^\star;y_{\mathrm{tgt}})$ and quantized margin $m_{f_q}(u^\star;y_{\mathrm{tgt}})$ as in \eqref{eq:app_margin_def_fp}--\eqref{eq:app_margin_def_q}. Consider the margin difference
\begin{equation}
\Delta m(u^\star)
\;=\;
m_f(u^\star;y_{\mathrm{tgt}})-m_{f_q}(u^\star;y_{\mathrm{tgt}}).
\label{eq:app_delta_margin}
\end{equation}
A sufficient condition for validity is $m_{f_q}(u^\star;y_{\mathrm{tgt}})>0$, which is implied by making $\Delta m(u^\star)$ small relative to $m_f(u^\star;y_{\mathrm{tgt}})$.

Assume that the quantization-induced logit perturbation can be decomposed into approximately additive layerwise contributions:
\begin{equation}
g_q(u^\star)-g(u^\star)
\;\approx\;
\sum_{\ell=1}^L \Delta g_\ell(u^\star),
\label{eq:app_additive_layers}
\end{equation}
where $\Delta g_\ell(u^\star)$ is the logit change attributable to quantizing layer $\ell$ (holding others fixed). Under this approximation, a conservative surrogate for margin degradation is
\begin{equation}
\begin{array}{l}
\displaystyle
|\Delta m(u^\star)|
\;\le\;
2\,\|g_q(u^\star)-g(u^\star)\|_\infty
\;\approx\;
2\,\left\|\sum_{\ell=1}^L \Delta g_\ell(u^\star)\right\|_\infty
\;\le\;
2\sum_{\ell=1}^L \|\Delta g_\ell(u^\star)\|_\infty.
\end{array}
\label{eq:app_margin_surrogate}
\end{equation}
Thus, reducing $\|\Delta g_\ell(u^\star)\|_\infty$ at recourse points directly reduces a bound on margin degradation, and consequently reduces validity failures.

\subsubsection{ Bitwidth controls per-layer perturbation magnitude}
\label{app:bits_control_perturb}

Assume a per-layer scaling relationship of the form
\begin{equation}
\begin{array}{l}
\displaystyle
\|\Delta g_\ell(u^\star)\|_\infty
\;\le\;
s_\ell(u^\star)\,q_\ell(b_\ell),
\qquad
q_\ell(b_\ell) \;=\; \kappa_\ell\,2^{-b_\ell},
\end{array}
\label{eq:app_layer_scaling}
\end{equation}
where $s_\ell(u^\star)\ge 0$ is a counterfactual sensitivity factor capturing how strongly layer $\ell$ influences logits at the recourse point, and $q_\ell(b_\ell)$ is a decreasing function of bitwidth (for uniform quantization, $2^{-b_\ell}$ is the canonical scaling up to constants). Combining \eqref{eq:app_margin_surrogate} and \eqref{eq:app_layer_scaling} yields the dataset-level surrogate objective
\begin{equation}
\begin{array}{l}
\displaystyle
\mathcal{S}(\{b_\ell\})
\;=\;
\mathbb{E}_{u^\star}\!\left[
\sum_{\ell=1}^L s_\ell(u^\star)\,\kappa_\ell\,2^{-b_\ell}
\right]
\;=\;
\sum_{\ell=1}^L \bar s_\ell \,\kappa_\ell\,2^{-b_\ell},
\end{array}
\label{eq:app_surrogate_obj}
\end{equation}
where $\bar s_\ell=\mathbb{E}_{u^\star}[s_\ell(u^\star)]$ averages sensitivity over recourse points.

A common bit budget constraint is a parameter-weighted cost
\begin{equation}
\sum_{\ell=1}^L n_\ell\,b_\ell \;\le\; B,
\label{eq:app_budget_constraint}
\end{equation}
where $n_\ell$ is the number of parameters (or MACs proxy weight) for layer $\ell$ and $B$ is the total budget. Consider the continuous relaxation $b_\ell\in\mathbb{R}_{\ge 0}$ and minimize \eqref{eq:app_surrogate_obj} subject to \eqref{eq:app_budget_constraint}. The Lagrangian is
\begin{equation}
\mathcal{L}(\{b_\ell\},\nu)
\;=\;
\sum_{\ell=1}^L \bar s_\ell \kappa_\ell 2^{-b_\ell}
\;+\;
\nu\left(\sum_{\ell=1}^L n_\ell b_\ell - B\right),
\label{eq:app_lagrangian}
\end{equation}
with multiplier $\nu\ge 0$. Setting $\partial \mathcal{L}/\partial b_\ell=0$ yields
\begin{equation}
\begin{array}{l}
\displaystyle
-\ln(2)\,\bar s_\ell \kappa_\ell\,2^{-b_\ell} \;+\; \nu n_\ell \;=\; 0
\qquad\Longrightarrow\qquad
2^{-b_\ell} \;=\; \frac{\nu n_\ell}{\ln(2)\,\bar s_\ell \kappa_\ell}.
\end{array}
\label{eq:app_stationary}
\end{equation}
Solving for $b_\ell$ gives
\begin{equation}
\begin{array}{l}
\displaystyle
b_\ell
\;=\;
\log_2\!\left(\frac{\ln(2)\,\bar s_\ell \kappa_\ell}{\nu n_\ell}\right).
\end{array}
\label{eq:app_opt_bits}
\end{equation}
Equation~\eqref{eq:app_opt_bits} shows that, under this surrogate, optimal precision increases with counterfactual sensitivity $\bar s_\ell$ and decreases with layer cost $n_\ell$. Discretizing to $\mathcal{B}=\{2,3,4,8\}$ produces the qualitative rule that CFQ should allocate higher bits to layers with large $\bar s_\ell/n_\ell$.

\subsubsection{ Estimating counterfactual sensitivity scores from gradients}
\label{app:sensitivity_scores}

CFQ provides a principled estimator of $\bar s_\ell$ using gradients of the counterfactual validity loss at teacher-recourse points. For a given recourse point $u^\star=x+\delta_{\mathrm{fp}}(x;y_{\mathrm{tgt}})$, define the counterfactual validity loss as
\begin{equation}
\ell_{\mathrm{valid}}(u^\star)
\;=\;
\mathcal{L}_{\mathrm{valid}}\!\left(f_q(u^\star),y_{\mathrm{tgt}}\right).
\label{eq:app_valid_loss_u}
\end{equation}
A first-order approximation of how a small perturbation in layer $\ell$ affects this loss is
\begin{equation}
\ell_{\mathrm{valid}}(u^\star;W_\ell+\Delta W_\ell) - \ell_{\mathrm{valid}}(u^\star;W_\ell)
\;\approx\;
\left\langle \nabla_{W_\ell}\ell_{\mathrm{valid}}(u^\star),\,\Delta W_\ell \right\rangle.
\label{eq:app_first_order_loss}
\end{equation}
If quantization imposes a bounded perturbation set $\Delta W_\ell\in\mathcal{E}_\ell(b_\ell)$ whose size shrinks with bitwidth, then a worst-case sensitivity proxy is
\begin{equation}
\begin{array}{l}
\displaystyle
s_\ell(u^\star)
\;=\;
\sup_{\Delta W_\ell\in\mathcal{E}_\ell(b_\ell=1)} 
\left\langle \nabla_{W_\ell}\ell_{\mathrm{valid}}(u^\star),\,\Delta W_\ell \right\rangle,
\end{array}
\label{eq:app_sens_def}
\end{equation}
where $\mathcal{E}_\ell(b_\ell=1)$ denotes a unit-scale perturbation set (the scaling with $b_\ell$ is then absorbed into $q_\ell(b_\ell)$). A commonly used simplification is to take $\mathcal{E}_\ell$ as an entrywise $\ell_\infty$ ball, which yields
\begin{equation}
s_\ell(u^\star)
\;\propto\;
\left\|\nabla_{W_\ell}\ell_{\mathrm{valid}}(u^\star)\right\|_1.
\label{eq:app_grad_l1}
\end{equation}
Averaging \eqref{eq:app_grad_l1} over recourse points produces $\bar s_\ell$, which matches the role of a counterfactual sensitivity score in \eqref{eq:app_surrogate_obj}. This formalizes the intuition that layers with consistently large gradients of the counterfactual loss at recourse points are precisely the layers where quantization noise most threatens validity, and therefore deserve higher precision under a fixed budget.

\paragraph{Connection to CFQ training dynamics.}
Because CFQ optimizes $\mathcal{L}_{\mathrm{valid}}(f_q(x+\delta_{\mathrm{fp}}),y_{\mathrm{tgt}})$ jointly with bit-policy parameters, gradients backpropagated through the mixed-precision relaxation implicitly increase bits where they most reduce counterfactual loss. In the surrogate view, this corresponds to decreasing $\mathcal{S}(\{b_\ell\})$ by allocating more bits to layers with larger $\bar s_\ell$, consistent with \eqref{eq:app_opt_bits}.

\vspace{-0.2cm}
\section{Discussion, Limitations, and Future Work}
\label{sec:discussion}

\paragraph{Discussion.}
CFQ reframes quantization as a deployment-time model change rather than a
purely numerical compression step. The main empirical message is that
accuracy-preserving quantization can still alter the local geometry that
determines actionable recourse. Across \textsc{Adult}, \textsc{German Credit},
and \textsc{COMPAS}, standard QAT and mixed-precision baselines preserve
predictive accuracy but substantially increase Validity Drop (VD) and
Counterfactual Recourse Gap (CRG). This shows that predictive fidelity on
factual test inputs is not sufficient to guarantee that pre-deployment recourse
recommendations remain valid after compression.

CFQ addresses this gap by making recourse preservation an explicit objective of
quantization. Rather than introducing a new recourse solver or a new quantizer,
CFQ connects two mature deployment components: teacher recourse generation and
low-bit model compression. The full-precision model identifies
recourse-relevant points $x+\delta_{\mathrm{fp}}$, and the quantized model is
trained to preserve the target outcome at those points while satisfying a
global bit budget. This objective directly targets the failure mode measured by
VD and CRG: whether an action that worked before quantization still works, and
whether the cost of achieving the target outcome increases after deployment.

The ablations further clarify the mechanism. Removing the counterfactual
validity term leaves accuracy almost unchanged but substantially worsens VD and
CRG, confirming that standard task loss does not control recourse stability.
Replacing learned mixed precision with uniform bit allocation also degrades
recourse preservation, indicating that some layers are more important for
maintaining local counterfactual geometry than others. Finally, increasing the
teacher-recoursing budget from $K=1$ to $K=3$ improves stability without
changing accuracy, supporting the view that better teacher counterfactuals
provide a stronger deployment supervision signal.

The additional experiments strengthen the deployment interpretation of CFQ.
The CF-PTQ variant shows that the same principle can be applied in a
training-free post-training regime by calibrating on both factual and
teacher-counterfactual points. The cost analysis shows that CFQ trades
additional teacher-recoursing computation for recourse stability, while cached
teacher actions and CF-PTQ provide lower-overhead alternatives. The
constraint-tightness and distribution-shift analyses indicate that the benefit
is not limited to a single easy action set or a single evaluation distribution.
Finally, subgroup and non-tabular experiments show that CFQ is not merely a
tabular accuracy artifact: when recourse is defined over structured latent or
semantic action spaces, counterfactual-faithful quantization continues to
reduce recourse failure and cost inflation.

\paragraph{Limitations.}
First, CFQ preserves recourse with respect to a specified predictive model and
action set; it does not by itself certify that the recommended action is
causally effective in the real world. If the action set omits institutional,
temporal, or causal feasibility constraints, then even stable recourse under
quantization may remain unrealistic. This limitation is shared with much of the
algorithmic recourse literature, and CFQ should therefore be used together with
domain-informed action constraints.

Second, CFQ introduces additional computation relative to accuracy-only PTQ or
QAT because it computes teacher recourse actions and evaluates counterfactual
validity during compression. The overhead is controlled by the number of
projected-gradient steps $K$, and our results show that low-step teachers
already provide useful supervision. Nevertheless, for very large models or
strictly training-free deployment pipelines, full CFQ-QAT may be less
attractive than CF-PTQ or cached-teacher variants.

Third, CFQ optimizes recourse stability on the calibration distribution of
factual and teacher-counterfactual points. Under substantial deployment shift,
new user populations, or changed action constraints, the learned bit allocation
may need recalibration. Our distribution-shift experiments show that CFQ
remains more stable than accuracy-centric mixed precision under controlled
shifts, but they should not be interpreted as a guarantee under arbitrary
future shifts.

\paragraph{Future work.}
Several extensions follow naturally. A first direction is to combine CFQ with
robust recourse methods that hedge over plausible future models, so that
recourse remains stable under both model compression and subsequent deployment
updates. A second direction is to replace single-point counterfactual
supervision with neighborhood-level or distributional supervision around
$x+\delta_{\mathrm{fp}}$, for example through margin constraints in a local ball
or by distilling multiple feasible recourse actions. A third direction is to
develop group-aware CFQ objectives that explicitly minimize worst-group VD or
CRG, combining recourse-faithful quantization with fair-quantization goals.
Finally, scaling CFQ to larger neural models, richer semantic action spaces,
and multimodal recourse settings would further clarify how
counterfactual-faithful compression behaves in realistic deployment pipelines.


\end{document}